\documentclass[twoside,11pt]{article}

\usepackage{amsmath}
\usepackage{appendix}
\usepackage{booktabs}
\usepackage{multirow}
\usepackage{colortbl}
\usepackage{float}
\usepackage{xcolor}
\usepackage{subcaption}
\usepackage{listings}
\usepackage{longtable}
\usepackage{footmisc}
\usepackage{jmlr2e}
\usepackage{longtable}
\usepackage{lscape}
\makeatletter
\newcommand{\printfnsymbol}[1]{%
  \textsuperscript{\@fnsymbol{#1}}%
}
\makeatother

\newcommand{\numframeworks}{9}
\newcommand{\numclassification}{71}
\newcommand{\numregression}{33}
\newcommand{\howtoframework}{\url{https://openml.github.io/automlbenchmark/docs/extending/framework/}}
\newcommand{\howtodataset}{\url{https://openml.github.io/automlbenchmark/docs/extending/benchmark/}}
\newcommand{\experimentdata}{\url{http://openml.github.io/automlbenchmark/data}}
\newcommand{\visapp}{\url{http://openml.github.io/automlbenchmark/visualization}}
\newcommand{\eg}[0]{e.g.,}
\newcommand{\ie}[0]{i.e.,}
\newcommand{\logl}[0]{log loss}
\newcommand{\rmse}[0]{rmse}

\newcommand{\systemcase}[1]{\texttt{\footnotesize \MakeUppercase{#1}}}

\newcommand{\autosklearn}[0]{\systemcase{auto-sklearn}}
\newcommand{\autosklearnii}[0]{\systemcase{auto-sklearn 2}}
\newcommand{\autogluon}[0]{\systemcase{autogluon}}
\newcommand{\autoweka}[0]{\systemcase{auto-weka}}
\newcommand{\autoxgboost}[0]{\systemcase{autoxgboost}}
\newcommand{\flaml}[0]{\systemcase{flaml}}
\newcommand{\gama}[0]{\systemcase{gama}}
\newcommand{\water}[0]{\systemcase{h2o automl}}
\newcommand{\lama}[0]{\systemcase{lightautoml}}
\newcommand{\mljar}[0]{\systemcase{mljar}}
\newcommand{\mlnet}[0]{\systemcase{mlnet}}
\newcommand{\mlplan}[0]{\systemcase{ml-plan}}
\newcommand{\mlrautoml}[0]{\systemcase{mlr3automl}}
\newcommand{\naml}[0]{\systemcase{naiveautoml}}
\newcommand{\tpot}[0]{\systemcase{tpot}}
\newcommand{\amlb}[0]{AMLB}

\lstdefinestyle{py-cmd}{
  language=bash,
  basicstyle=\small,
  columns=fixed, %flexible,
  morekeywords={python}
}

\jmlrheading{1}{2000}{1-48}{4/00}{10/00}{Pieter Gijsbers, Marcos L. P. Bueno, Stefan Coors, Erin LeDell, S\'{e}bastien Poirier, Janek Thomas, Bernd Bischl and Joaquin Vanschoren}

\ShortHeadings{AMLB: an AutoML Benchmark}{Gijsbers et al.}
\firstpageno{1}

\begin{document}

\title{AMLB: an AutoML Benchmark}
      \author{\name  Pieter Gijsbers\textsuperscript{1} \email p.gijsbers@tue.nl 
      \AND
      \name Marcos L. P. Bueno\textsuperscript{1,4} \email marcos.depaulabueno@donders.ru.nl 
      \AND
      \name  Stefan Coors\textsuperscript{2} \email stefan.coors@stat.uni-muenchen.de 
      \AND
      \name  Erin LeDell\textsuperscript{3} \email erin@h2o.ai 
      \AND
      \name  S\'ebastien Poirier\textsuperscript{3} \email sebastien@h2o.ai 
      \AND
      \name  Janek Thomas\textsuperscript{2} \email janek.thomas@stat.uni-muenchen.de 
      \AND 
      \name Bernd Bischl\textsuperscript{2} \email bernd.bischl@stat.uni-muenchen.de 
      \AND
      \name  Joaquin Vanschoren\textsuperscript{1} \email j.vanschoren@tue.nl \\
      \vskip 1em
      \noindent \textsuperscript{1} Eindhoven University of Technology, Eindhoven, The Netherlands\\
      \textsuperscript{2} Ludwig Maximilian University of Munich, Munich, Germany\\
      \textsuperscript{3} H2O.ai, Mountain View, CA, United States\\
      \textsuperscript{4} Radboud University, Nijmegen, The Netherlands\\
      }
      
\editor{TBD}

\maketitle

\begin{abstract}%   <- trailing '%' for backward compatibility of .sty file required in JMLR
%In recent years, an active field of research has developed around automated machine learning (AutoML).
% Unfortunately, c
Comparing different AutoML frameworks is notoriously challenging and often done incorrectly.
We introduce an open and extensible benchmark that follows best practices and avoids common mistakes when comparing AutoML frameworks. We conduct a thorough comparison of \numframeworks{} well-known AutoML frameworks across \numclassification{} classification and \numregression{} regression tasks.
The differences between the AutoML frameworks are explored with a multi-faceted analysis, evaluating model accuracy, its trade-offs with inference time, and framework failures. We also use Bradley-Terry trees to discover subsets of tasks where the relative AutoML framework rankings differ. 
The benchmark comes with an open-source tool that integrates with many AutoML frameworks and automates the empirical evaluation process end-to-end: from framework installation and resource allocation to in-depth evaluation. The benchmark uses public data sets, can be easily extended with other AutoML frameworks and tasks, and has a website with up-to-date results.

\end{abstract}

% no capitalized keywords

\section{Introduction}

To create useful machine learning (ML) models from data, data scientists must prepare the data for consumption by ML algorithms (for example, coercing it to a different format or by encoding categorical features), select an ML algorithm, and tune its hyperparameters.
This requires extensive expertise, such as knowing which hyperparameters to tune and how~\citep{Probst2019, weerts2020importance}.
Even with this knowledge, it is a time-consuming task, since the best choices are unique to each data set and can be interdependent on each other~\citep{JvR2018hyperparameter}.

The field of automated machine learning (AutoML) is focused on automating the design and optimization of ML pipelines in a data-driven way~\citep{automl_book}. Neural Architecture Search (NAS) is an important part of AutoML that automates the design decisions of deep neural networks.
AutoML aims to free up valuable time for experts to perform other tasks and allow novice users to train well-performing ML models.

Many different AutoML approaches have been proposed, including sequential model-based optimization \citep{hutter2011sequential, snoek2012practical}, hierarchical task planning~\citep{erol1994umcp}, and genetic programming~\citep{koza1992genetic}. 
Novel systems are being developed in both academia and industry, and a recent survey by \cite{van2021automl} showed that $69$\% of $307$ practitioners (at least partially) adopt automated model selection and hyperparameter configuration.

\subsection{The Need for Standardized Benchmarking}
With considerable effort being spent on developing and improving AutoML frameworks as well as increased usage by practitioners, there is a need for systematic and in-depth comparisons of the the various approaches to track progress in the field.
However, the comparison of AutoML frameworks is prone to several types of error.
First, selection bias may be introduced, even accidentally, when authors decide which data sets to use in their evaluation. For example, too few data sets may be selected to accurately evaluate the framework's strengths and weaknesses, and the chosen data sets may no longer be challenging for current AutoML frameworks. Without a standard suite of data sets to use for evaluation, the selection of data sets is often not reasonably justified and motivated.
Moreover, issues may arise from errors in the installation, configuration, or use of `competitor' frameworks.
Typical examples are misunderstanding memory management and/or using insufficient compute resources~\citep{DBLP:journals/corr/abs-1808-06492}, or failing to use comparable resource budgets~\citep{ferreira2021protein}.

Several suites of benchmark data sets have been used in the aforementioned papers, but none have become a standard in the AutoML community. The original selection of data sets by~\cite{AutoWEKA1} was used in several earlier papers \citep{feurer2015efficient, MLPlan}, but fails to highlight differences between current AutoML frameworks.
Most published AutoML papers use a self-selected suit of data sets on which their methods are evaluated.
For example, \cite{drori2018alphad3m}, \cite{rakotoarison2018automl} and \cite{gil2018p4ml} were all published at the same time but feature different suites on which they evaluate their contributions.
This inconsistency makes it impossible to directly compare results across papers and track progress over time. It also potentially allows for presenting cherry-picked results.

\subsection{Our Contributions}
We introduce a novel AutoML benchmark following best practices to avoid these common pitfalls while stimulating progress towards more standardized benchmarking.\footnote{A first look of this tool was presented at the ICML 2019 AutoML Workshop~\citep{amlb2019}. The current version is significantly more general, more systematic, and allows much more in-depth analysis.}
To ensure reproducibility\footnote{Using the ACM definition: an independent group can obtain the same result using the author’s own artifacts (\url{https://www.acm.org/publications/policies/artifact-review-and-badging-current}).}, we provide an open-source benchmarking tool\footnote{Code, results, and documentation at: \url{https://openml.github.io/automlbenchmark/}} that allows easy integration with AutoML frameworks, and performs end-to-end evaluations thereof on carefully curated sets of open data sets. Our focus is on tabular data. Unstructured data are out of scope for this paper, since they are best tackled with NAS, and benchmarking NAS frameworks imposes different practical constraints, as discussed in Section~\ref{ssec:other_bench}. We also restrict our evaluations to open-source AutoML frameworks.

Our benchmarking tool, dubbed \amlb{} (for AutoML Benchmark), can be used to perform evaluations of AutoML frameworks in a fully automated way. The AutoML frameworks are integrated with the benchmarking tool in direct agreement and jointly with the framework's developers to ensure correctness.
We carry out a large-scale evaluation of \numframeworks{} well-known AutoML frameworks, some of which with multiple configurations, across \numclassification{} classification and \numregression{} regression tasks.
To better understand how these systems perform across many tasks, we also introduce techniques for detailed comparison of AutoML frameworks, including final model accuracy, inference time trade-offs, and failure analysis. Finally, we provide an interactive visualization tool that may be used for further exploration of all our results or to reproduce the analyses performed in this paper.

In the remainder of this paper, we first discuss related work in Section~\ref{sec:literature} and cover several key open-source AutoML frameworks in Section~\ref{sec:frameworks}.
Next, we provide an overview of our proposed benchmarking tool in Section~\ref{sec:software} and motivate our benchmark design choices in Section~\ref{sec:design}. The results obtained by running this benchmark are analyzed in Section~\ref{sec:results}. Finally, Section~\ref{sec:conclusion} summarizes our main conclusions and sketches directions for future work.

\section{Related Literature \label{sec:literature}}
In this section, we motivate why we need benchmarks specifically designed for AutoML, review other work evaluating AutoML frameworks, and finally discuss the relevant ML benchmarking literature.
Several benchmark suites have been developed in ML~\citep{van2004benchmarking, olson2017pmlb, wu2018moleculenet, bischl2021openml, fischer2023openml}. 
The data sets in these suites often do not include problematic data characteristics found in real world tasks (for example, many missing values), as many ML algorithms are not natively able to handle them.
By contrast, in order to be applicable to a wide range of data, AutoML frameworks should be designed to handle these problematic data sets. As such, there is opportunity to allow more such problematic data sets in ML benchmarks and study how well AutoML frameworks handle these issues.
Moreover, runtime budgets are crucial in an AutoML benchmark, as most AutoML frameworks are designed to run until a given time budget is exhausted. These runtime budgets are often not specified beforehand in traditional ML benchmarks, since algorithms must usually run to completion. \footnote{One exception are performance studies, such as \citealp{kotthaus2015}.} Consequently, new benchmarks that have been designed specifically for AutoML frameworks are needed.

\subsection{Evaluation of Automated Machine Learning Frameworks}
To establish new best practices for AutoML benchmarking, it is beneficial to study the shortcomings encountered in prior benchmarks as well as lessons learned. \cite{DBLP:journals/corr/abs-1808-06492} conducted one of the first benchmark studies on AutoML frameworks.
They evaluated four open-source frameworks on both classification and regression tasks sourced from OpenML~\citep{vanschoren-sigkdd13a}, optimized for weighted F1 score and mean squared error, respectively.
Unfortunately, they encountered technical issues with most AutoML frameworks that led to a questionable experimental evaluation.
For example, \water{}~\citep{H2OAutoML} was configured to optimize to a different metric (log loss as opposed to weighted F1 score) and ran with a different setup (unlike the others, \water{} was not containerized), and \systemcase{auto\_ml}~\citep{automlrepo} had its hyperparameter optimization (HPO) disabled, making for incomparable results. This highlights the need for careful configuration of all AutoML frameworks involved.

A study by \cite{ferreira2021protein} evaluated the effectiveness of AutoML frameworks for protein abundance prediction. This evaluation compared \water{} with a 6 hour limit to \gama{} with a one hour limit, and \tpot{} with a limit defined by the hyperparameter configuration of its evolutionary optimization (1000 generations with a population size of 250). While the comparison was not the main contribution of the paper, the AutoML frameworks were directly compared against each other, and these very different budgets were not motivated.

A study on nearly 300 data sets across six different frameworks was conducted by~\cite{TruongWGHBF19}. Each experiment consisted of a single 80/20 holdout split on a 15-minute training time budget, which was chosen so that most tools returned a result on at least 70\% of the data sets. 
We postulate it is reasonable to assume that the data sets for which no result is returned by a framework are most often those data sets for which optimization is hard. For example, a large data set might cause one framework to conduct only few evaluations while it completely halts another.
Unfortunately, this makes the resulting comparisons uninterpretable, as a framework could seemingly demonstrate better performance simply because it failed to return models on data sets for which optimization was difficult. Hence, it is key that failures must be avoided as much as possible, and any remaining failures should be analysed and taken into account in subsequent analysis.

On the positive side, \cite{TruongWGHBF19} present their results across different subsets of the benchmark---for example, few versus many categorical features---which helps to highlight differences between different frameworks. The authors also conduct small-scale experiments to analyze performance over time by running the tools on multiple time budgets on a subset of data sets as well as the `robustness', which denotes the variance in final performance given the same input data. 
Unfortunately, both experiments were conducted on only one data set per sub-category, which does not enable generalizing the results. Still, studying the impact of data characteristics and budget sizes should ideally be part of AutoML benchmark design.

\citep{ZollerH21} present a survey on AutoML and combined algorithm selection and HPO frameworks (CASH,~\citealp{AutoWEKA1}). Six CASH and five AutoML frameworks are compared across 137 classification tasks, with the former limited to 325 iterations and the latter constrained to a one hour time budget.
% Compiled from OpenML100, CC18 and our workshop paper.
Among the CASH frameworks, hyperopt~\citep{bergstra2013making} performed best, although absolute differences were small between all optimizers. % PG: BOHB performing worse than even random search makes me question the results.
The AutoML frameworks are compared as they are, which might reflect common use. However, by not controlling their settings, it becomes impossible to draw conclusions about the effectiveness of individual parts of AutoML systems. A number of errors of AutoML frameworks are observed, including memory constraint violations, segmentation faults, and Java server crashes. The authors also find that most frameworks construct rather modest pipelines (with few preprocessing operators). As such, it is recommended that AutoML frameworks are controlled carefully, use a wide operator search space, and are evaluated on datasets that require non-trivial preprocessing.

Kaggle\footnote{\url{https://www.kaggle.com/} \label{foot:kaggle}}, a platform for data science competitions, is sometimes used to compare AutoML frameworks to human data scientists~\citep{ZollerH21, autogluontabular}.
\citealp{ZollerH21} found that the rankings of AutoML frameworks on benchmark datasets are very different from the rankings on competitions. Furthermore, humans still find better solutions than the examined AutoML frameworks.
However, it is hard to interpret such results. Submissions on Kaggle leaderboards range from serious attempts to random test runs, and a significant portion of them do not outperform a simple baseline.
Finally, most Kaggle results are several years old and possibly no longer represent state-of-the-art human-made models because of advances in the available hardware or algorithms.

A recent AutoML benchmark for multi-label classification \citep{Wever} proposed a general tool with a configurable search space and optimizer, which allows for the inclusion of new methods and ablation studies. Unfortunately, this approach requires that existing AutoML frameworks be re-implemented within this tool, which is difficult in such a rapidly developing field. As such, we need a simpler way to include existing and new AutoML frameworks into benchmarking tools while still allowing control over their configuration.

In addition to AutoML benchmarks, a series of competitions for tabular AutoML was hosted~\citep{guyon2019analysis}.
The first two competitions focused on tabular AutoML, where data is assumed to be independent and identically distributed.
In these competitions, participants submitted code that automatically builds a model on given data and produced predictions for a test set.
During the development phase, competitors could make use of a public leaderboard and several validation data sets.
After the development phase, the latest submissions of each participant would be evaluated on a set of new data sets to determine the final ranking.
These data sets consisted of a mix of both new data and data taken from public repositories, although they were reformatted to conceal their identity. 
In their analysis \cite{guyon2019analysis} reveal that most methods failed to return results on at least some data sets due to practical issues, such as running out of memory.

\subsection{Other (Benchmark) Literature}
\label{ssec:other_bench}

Instead of benchmarking whole (Auto-) ML frameworks, it is often useful to focus on various sub-parts and optimize them step-by-step.
One part of such a system is HPO, and its available algorithms are as numerous and diverse as the learning algorithms themselves.
Consequently, there exist various benchmarking suites to foster research by comparing those optimization algorithms.

Like non-black-box optimization, black box optimizers are typically evaluated on synthetic test functions or on real-world tasks.
A well-established benchmarking suite for continuous optimization is COCO \citep{hansen2021coco}, which includes a collection of various synthetic black-box benchmark functions. Nevergrad\footnote{https://facebookresearch.github.io/nevergrad/benchmarking.html} is a popular platform for gradient-free optimization with benchmarking functionality.
\textit{kurobako} \citep{kurobako} provides various general black-box optimizers and benchmark problems, while \textit{LassoBench} \citep{vsehic2021lassobench} is suitable for benchmarking high-dimensional optimization problems. Especially for Bayesian Optimization, \textit{Bayesmark} \citep{bayesmark} combines several benchmarks on real-world tasks.

One of the first benchmarks for empirically evaluating HPO algorithms was \textit{HPOlib} \citep{eggensperger_towards_2013}, which allows accessing real-world HPO tasks, tabular and surrogate benchmarks, and synthetic test functions using a common API. This benchmark was used by \cite{bergstra2014preliminary} in their empirical benchmark studies.
\textit{HPOBench} \citep{hpobench} is a similar successor of \textit{HPOlib} that concentrates on reproducible containerized benchmarks and multi-fidelity optimization problems.
Based on OpenML \citep{vanschoren-sigkdd13a}, \citet{arango2021hpob} recently introduced \textit{HPO-B}, a large-scale reproducible benchmark for transfer-HPO methods.
In contrast, \textit{PROFET} \citep{profet} uses a generative meta-model to generate synthetic but realistic benchmark instances.

A related benchmark is \textit{NAS-Bench-101} \citep{NAS-Bench-101}, which is a tabular data set that maps convolutional neural network architectures to their trained and evaluated performance on CIFAR-10. Its goal is to make NAS more accessible, despite the tremendous demand of computational resources. Additionally, there exists a series of NAS-Bench systems that includes \textit{NAS-Bench-1shot1} \citep{NAS-Bench-1Shot1}, \textit{NAS-Bench-301} \citep{NAS-Bench-301}, and others.

\section{AutoML Frameworks \label{sec:frameworks}}
Automated ML pipeline design was first explored by \cite{escalante2009particle},
but the first prominent AutoML framework was \autoweka{}~\citep{AutoWEKA1}.
\autoweka{} used Bayesian optimization to select and tune the algorithms in an ML pipeline based on \systemcase{weka}~\citep{WEKA}.
Since then, a plethora of new AutoML frameworks have been developed, either by iteratively improving on old designs or using novel approaches.
In this section, we will discuss the tools we considered for our benchmark.

Unfortunately, the cost of evaluating all frameworks is prohibitive, so we selected \numframeworks{} of them to evaluate in this work.
Only open source tools were considered. From those, we made selections to cover a variety of different optimization approaches.
We considered frameworks developed by industry as well as academia and included packages whose authors proactively integrated their AutoML framework, so that we are confident that they are integrated correctly.

As our experiments are run on machines without GPU, we focus on AutoML frameworks which are not significantly affected by the lack thereof.
For example, \systemcase{Auto-Keras}~\citep{autokeras}, and \systemcase{AutoPyTorch}~\citep{zimmer-tpami21a} gain large benefits from having a GPU available, thus reporting an evaluation with only CPU available might give a pessimistic performance estimation compared to typical use of those frameworks. However, even for some of our selected frameworks, such as \water{} and \autogluon{}, specific base models support GPU acceleration when it is available.

The most notable omission is \autoweka{}, which we decided to exclude based on the performance in our 2019 evaluation and its lack of updates since~\cite{amlb2019}.
Some frameworks integrated with the benchmark tool are not evaluated in this paper. 
The authors of \autoxgboost{}~\citep{autoxgboost} opted out of an evaluation because it is built on deprecated software with no plans for updates.
\mlplan{}~\citep{MLPlan}, \mlrautoml{}\footnote{Source and documentation of \mlrautoml{} at \url{https://github.com/a-hanf/mlr3automl/}.}, and \mlnet{}\footnote{Source and documentation of \mlnet{} at \url{https://github.com/dotnet/machinelearning/}.} are excluded because we encountered significant technical problems evaluating the systems. We hope to publish results from these frameworks at a later date. 
There are still many tools not yet integrated with the AutoML benchmark, and we hope that we can work together with AutoML developers to add additional integrations in the future. 

\subsection{Integrated Frameworks}

Table~\ref{table:automl_overview} offers an overview of the AutoML frameworks evaluated in this paper alongside a simplified description of their optimization and search space design, which are elaborated below.
We refer the interested reader to the original publications and our website, which has 
links to the frameworks' source code and documentation.
\naml{} was also evaluated but excluded from the analysis in Section~\ref{sec:results}, as explained in Appendix~\ref{app:naml}.

\begin{table}[t]
\centering
\begin{tabular}{llll}
\toprule
Framework       & Optimization and Search Space & Reference  \\
\midrule
\autogluon{}       & Stacked ensembles of preset pipelines & \cite{autogluontabular}  \\
\autosklearn{}     & BO of \systemcase{scikit-learn} pipelines & \cite{feurer2015efficient}  \\
\autosklearnii{}   & BO of iterative algorithms & \cite{feurer2020autosklearn}  \\
\flaml{}           & CFO of iterative algorithms & \cite{FLAML}  \\
\gama{}            & EO of \systemcase{scikit-learn} pipelines & \cite{gijsbers2020gama}   \\
\water{}       & Iterative mix of RS and ensembling & \cite{H2OAutoML}   \\
\lama{}     & BO of linear models and GBM & \cite{vakhrushev2021lightautoml}  \\
\mljar{} & Custom data science pipeline & \cite{mljar}  \\
\naml{}            & Custom data science pipeline & \cite{mohr2023naive}  \\
\tpot{}            & EO of \systemcase{scikit-learn} pipelines & \cite{TPOT}  \\
\bottomrule
\end{tabular}
\caption{Used AutoML frameworks in the experiments. Different optimization techniques such as Bayesian optimization (BO), evolutionary optimization (EO), random search (RS) or cost frugal optimization (CFO) are used across different search spaces.}
\label{table:automl_overview}
\end{table}

\subsubsection{AutoGluon-Tabular \label{sec:autogluon}}
\autogluon{} automates ML across a variety of tasks, including image, text, and tabular data. 
The subsystem that automates ML on tabular data is called \systemcase{AutoGluon-Tabular}~\citep{autogluontabular}, but we will refer to it as \autogluon{}.
In contrast to other AutoML systems discussed here, \autogluon{} does not perform a pipeline search or hyperparameter tuning.
Instead, it has a predetermined set of models that are combined through multi-layer stacking and ensembling.

\autogluon{}'s ensemble consists of three layers. 
The first layer consists of models from a range of model families trained directly on the data.
In the second layer, the same type of models are considered but rather as a stacking learner trained with both the input data and the predictions of the first layer.
In the final layer, the predictions of the second-layer models are combined into an ensemble, using an ensemble method~\citep{caruana2004ensemble} first used in AutoML by \autosklearn{}~\citep{feurer2015efficient}.

To adhere to time constraints, \autogluon{} may stop iterative algorithms prematurely or forgo training certain models altogether. Given more time, \autogluon{} will train additional models using the same algorithms and hyperparameter configurations on different data splits, which further improves the generalization of the stacking layer.

\autogluon{} offers many different presets that affect the trade-off between final model performance and inference time. In this paper, we evaluated three presets: best quality, high quality, and high quality with an inference time limit (denoted as B, HQ, and HQIL, respectively). These respective presets produce increasingly faster models at the cost of decreased model accuracy.

\subsubsection{auto-sklearn \label{sec:auto-sklearn}}
%Started as an iterative improvement on Auto-WEKA, 
Based on the design of \autoweka{}, \autosklearn{}~\citep{feurer2015efficient} also uses Bayesian optimization but is instead implemented in Python and optimizes pipelines built with \systemcase{scikit-learn}~\citep{pedregosa2011scikit}.
Additionally, it warm-starts optimization through meta-learning, starting pipeline search with the best pipelines for the most similar data sets~\citep{feurer2015initializing}.
After pipeline search has concluded, an ensemble is created from pipelines trained during search using the procedure described by \cite{caruana2004ensemble, caruana2006getting}.
\autosklearn{} has won two AutoML challenges~\citep{guyon2019analysis}, although for both entries, \autosklearn{} was customized for the competition, and not all changes are found in the public releases~\citep{feurer2018posh}.

Based on experience from the challenges, \autosklearnii{} was subsequently developed \citep{feurer2020autosklearn}.
The most notable changes include reducing the search space to only iterative learning algorithms and excluding most preprocessing, use of successive halving~\citep{successivehalving}, adaptive evaluation strategies, and replacing the data-specific warm-start strategy with a data-agnostic portfolio of pipelines.
Because these changes make version 2.0 almost entirely different from 1.0, and 1.0 has been updated since our last evaluation, we evaluate both auto-sklearn versions in this paper.
However, \autosklearnii{} does not yet support regression, and its heavy use of meta-learning made it impossible for us to perform a `clean' evaluation at this time (see Section~\ref{sec:meta-learning}).

\subsubsection{FLAML} 
The Fast and Lightweight AutoML Library (\flaml{},~\citealp{FLAML}) optimizes boosting frameworks (\systemcase{xgboost},~\citealp{xgboost}, \systemcase{catboost},~\citealp{catboost2018}, and \systemcase{lightgbm},~\citealp{ke2017lightgbm}) and a small selection of \systemcase{scikit-learn} algorithms through a multi-fidelity randomized directed search called Cost-Frugal Optimization \citep{wu2021frugal}.
This search is based on an expected cost for improvement, which tracks the expected computational cost of improving over the best model found so far for each learner.
Only after choosing which learner to tune, hyperparameter optimization proceeds by a randomized directed search, sampling a new configuration from a unit sphere around the previous sample point.
After evaluating its validation performance, the next sample point is moved to that direction (if better) or the opposite direction (if worse).
\flaml{} positions itself as a fast AutoML framework that is designed to work for small time budgets~\citep{FLAML}.

\subsubsection{GAMA} 
Designed as a modular AutoML tool for researchers, \gama{}'s search method and post processing are easily configurable and extensible~\citep{gijsbers2020gama}. 
By default, \gama{} uses genetic programming to optimize linear ML pipelines with an arbitrary amount of preprocessing algorithms.
Similar to \tpot{}, \gama{}'s evolutionary algorithm uses NSGA-II to perform multi-objective optimization~\citep{deb2002fast}, maximizing performance while minimizing the number of components in the pipeline.
By contrast, \gama{}'s evolutionary algorithm is asynchronous and does not work with distinct generations, which allows for higher resource utilization.
The final model is constructed through ensemble selection~\citep{caruana2004ensemble, caruana2006getting}, similar to \autosklearn{}.

\subsubsection{H2O AutoML} 
Built on the \systemcase{H2O} distributed machine learning platform~\citep{h2o_platform}, \water{}~\citep{H2OAutoML} evaluates a portfolio of bespoke algorithm configurations and also performs random searches over the majority of the supervised learning algorithms offered in \systemcase{H2O}.  Early stopping is applied to the searches for efficiency.  The amount of time the \water{} spends searching each algorithm is pre-specified by the authors, though this can be customized.  This allocates the amount of optimization work done at each step of the algorithm, and some stronger algorithms (e.g. XGBoost) are favored, or given more time, over others (e.g. GLM).  To further increase model performance, \water{} also trains two types of stacked ensemble models at various stages during the run: an ensemble using all available models at step $s$, and an ensemble with only the best models of each algorithm type at step $s$.  

\water{} aims to cover a large search space quickly and relies on stacking to boost model performance. 
Additionally, \water{} uses a predefined strategy for imputation, normalization, and categorical encoding for each algorithm and does not currently optimize over preprocessing pipelines by default, but preprocessing can be turned on as an option. 
The \water{} algorithm is designed to generate models that are very fast at inference time, rather than strictly focusing on maximizing model accuracy, with the goal of balancing these two competing factors to produce practical models suited for production environments.

\subsubsection{LightAutoML}
\lama{} is specifically designed with applications in the financial services industry in mind~\citep{vakhrushev2021lightautoml}.
In this framework, pipelines are designed for quick inference and interpretability. 
Only linear models and boosting frameworks (\systemcase{catboost} and \systemcase{lightgbm}) 
are considered.
Expert rules are used to evaluate likely good hyperparameter configurations and design the search space.
Tree-structured Parzen Estimators~\citep{bergstra2011implementations} are used to optimize hyperparameters of the boosting frameworks, while warm-starting and early stopping are used to optimize linear models with grid search.
Different models are combined in either a weighted voting ensemble (binary classification and regression) or with two levels of stacking (multi-class classification).
In a special `compete' mode for larger time budgets, the AutoML pipeline is run multiple times with different configurations and their resulting models are ensembled with weighted voting, which allows for a more robust model.

\subsubsection{MLJAR}
\mljar{}~\citep{mljar} starts its search with a set of predetermined models and a limited random search---similar to \water{}. 
This is followed by a feature creation and selection step, after which a hill climbing algorithm is used to further tune the best pipelines. 
After search, the models can be stacked, used in a voting ensemble, or both.
Learners from \systemcase{scikit-learn} are considered, as well as boosting packages \systemcase{xgboost}, \systemcase{catboost}, and \systemcase{lightgbm}.
\mljar{} features multiple modes for different use cases, including exploratory data analysis or finding a fast model, a `perform' mode aimed at finding a good model with good inference time, and a `compete' mode which aims to find the best possible model. We evaluate both the `compete' and `perform' presets in this work.

\subsubsection{Naive AutoML}
Proposed as a baseline by \citet{mohr2023naive}, it mimics a simple workflow a data scientist might execute. First, a base learner is chosen by evaluating their performance using their default hyperparameter configuration. Then several steps are performed in order, including feature scaling, feature selection, and hyperparameter tuning. Unfortunately, we encountered issues evaluating \naml{} which prohibited us to include its results in Section~\ref{sec:results}. Appendix~\ref{app:naml} motivates this choice, and shows the results we obtained for \naml{}.

\subsubsection{TPOT}
Tree-based Pipeline Optimization Tool~\citep{TPOT}, or \tpot{}, optimizes pipelines using genetic programming (\eg{}~\cite{mckay2010grammar}): ML pipelines can be expressed as trees where different branches represent distinct preprocessing pipelines.
These pipelines are then optimized through evolutionary optimization.
To reduce overfitting that may arise from the large search space, multi-objective optimization is used to minimize for pipeline complexity while optimizing for performance~\citep{TPOT-pareto}. 
It is also possible to reduce the search space by specifying a pipeline template~\citep{tpot-ds}, which dictates the high-level steps in the pipeline (for example, a ``Selector-Transformer-Classifier" template will result in pipelines with only those three steps, in that order).
Development has focused around genomic studies, providing specific options for dealing with this type of high dimensional data for which prior knowledge may be present~\citep{tpot-mdr}.
While \tpot{} supports neural networks in its search~\citep{tpot-nn}, the default search space uses \systemcase{scikit-learn} components and \systemcase{XGBoost} only.

\subsection{Baselines}

In addition to the integrated frameworks, the benchmark tool allows for running several baselines.
The \texttt{constant predictor} always predicts the class prior or mean target value, regardless of the values of the independent variables. 
The \texttt{Random Forest} baseline builds a forest $10$ trees at a time, until one of two criteria is met: we expect to exceed $90$\% of the memory limit or time limit by building 10 more trees, or $2000$ trees have been built.

The \texttt{Tuned Random Forest} baseline improves on the \texttt{Random Forest} baseline by using an optimized \texttt{max\_features} value. 
The \texttt{max\_features} hyperparameter defines how many random features are considered when determining each split and is found to be the most important hyperparameter~\citep{JvR2018hyperparameter}.\footnote{The hyperparameter \texttt{min\_samples\_leaf} is statistically equally important.}
The value is optimized by evaluating up to $11$ unique values for the hyperparameter with $5$-fold cross-validation before training a final model with the best found value. The \texttt{Tuned Random Forest} is our strongest baseline and aims to mimic a typical first approach for modeling by a human.

\section{Software \label{sec:software}}

We developed an open source benchmark tool that may be used for reproducible AutoML benchmarking, which we will refer to as \amlb{}.\footnote{Source code and documentation under MIT license at \url{https://github.com/openml/automlbenchmark}.}
This tool features robust automated experiment execution and supports multiple AutoML frameworks, many of which are evaluated in this paper.
\amlb{} is implemented as a Python application consisting mainly of an \emph{amlb} module and a \emph{framework} folder hosting all the officially supported extensions, which have been developed together with AutoML framework developers.
The main consideration for the design of \amlb{} is to produce correct and reproducible evaluations;
the AutoML frameworks are used as intended by their authors with little to no room for user error, and the same evaluation conditions (including framework version, data set, and resampling splits) and controlled computational environments can easily be recreated by anyone.
The \emph{amlb} module provides the following features:
\begin{itemize}
    \item a data loader to retrieve and prepare data from \emph{OpenML} or local data sets. 
    \item various benchmark runner implementations:
    \begin{itemize}
        \item a \emph{local} runner that runs the experiments directly on the local machine. This is also the runner to which each runner below delegates the final execution.
        \item \emph{container} runners (\emph{docker} and \emph{singularity} are currently supported), which allow preinstalling the \emph{amlb} application together with a full setup of one framework and consistently run all benchmark tasks against the same setup. This implementation also makes it possible to run multiple container instances in parallel.
        \item an \emph{aws} runner that allows the user to safely run the benchmark on several EC2 instances in parallel. Each EC2 instance can itself use a pre-built docker image, as used for this paper, or can configure the target framework on the fly, which is useful for experiments in a development environment.
    \end{itemize}
    \item a job executor responsible for running and orchestrating all the tasks. When used with the \emph{aws} runner, this allows distribution of the benchmark tasks across hundreds of EC2 instances in parallel, with each one being monitored remotely by the host.
    \item a post-processor responsible for collecting and formatting the predictions returned by the frameworks, handling errors, and computing the scoring metrics before writing the information needed for post-analysis to a file. % \emph{results.csv}
\end{itemize}

Figure~\ref{fig:amlb-aws-archi} in Appendix~\ref{app:software} provides an architecture overview and description of the flow of the benchmarking tool, as used in the experiments for this paper.

\subsection{Extensible Framework Structure}
\label{sec:framework_struct}
To ensure that \amlb{} is easily extensible to new AutoML frameworks, we integrate each tool through a minimal interface.
Each of the current tools requires less than 250 lines of code across at most four files (most of which is boilerplate).
The integration code handles installation of the AutoML framework as well as its software stack and provides the framework with data and recording predictions.
The integration requirements are minimal, as both input data and predictions can be exchanged both in Python objects and common file formats, which makes integration across programming languages possible (currently integrated frameworks are written in C$^\#$, Java, Python and R).
By keeping the integration requirements minimal, we hope that AutoML framework authors are encouraged to contribute integration scripts for their framework and, at the same time, avoid influencing the methods or software used to design and develop new AutoML frameworks, as opposed to providing a generic starter kit which may bias the developed AutoML frameworks~\citep{guyon2019analysis}.
Frameworks may also be integrated completely locally to allow for private benchmarking.\footnote{For information on how to add a framework, see \howtoframework{}.}

\subsection{Extensible Benchmarks}
\label{sec:ext_data}
Benchmark suites define the data sets and one or more train/test splits, which should be used to evaluate the AutoML frameworks.
\amlb{} can work directly with OpenML tasks and suites~\citep{bischl2021openml}, allowing for new evaluations without further changes to the tool or its configuration.
This is the preferred way to use \amlb{} for scientific experiments, as it guarantees that the exact evaluation procedure can be reproduced easily by others.
However, it is also possible to use data sets stored in local files with manually defined splits, e.g., to benchmark private use cases.\footnote{For information on how to add a new benchmark task or suite, see \howtodataset{}.}

\subsection{Using the Software}
\lstset{style=py-cmd}

To benchmark an AutoML framework, the user must first identify and define:
\begin{itemize}
    \item the \emph{framework} against which the benchmark is executed,
    \item the \emph{benchmark} suite listing the tasks to use in the evaluation, and 
    \item the \emph{constraints} that must be imposed on each task. This includes:
    \begin{itemize}
        \item the maximum training time. 
        \item the number of CPU cores that can be used by the framework; not all frameworks respect this constraint, but when run in \emph{aws} mode, this constraint translates to specific EC2 instances, therefore limiting the total number of CPUs available to the framework.
        \item the amount of memory that can be used by the framework; not all frameworks respect this constraint, but when run in \emph{aws} mode, this constraint translates to specific EC2 instances, therefore limiting the total amount of memory available to the framework.
        \item the amount of disk volume that can be used by the framework (only respected in \emph{aws} mode).
    \end{itemize}
    Those constraints must then be declared explicitly in a \emph{constraints.yaml} file (also in the \emph{resources} folder or as an external extension).
\end{itemize}       

\subsubsection{Commands}

Once the previous parameters have been defined, the user can run a benchmark on the command line using the basic syntax:
\begin{lstlisting}
  $ python runbenchmark.py framework_id benchmark_id constraint_id
\end{lstlisting}

\noindent For example, to evaluate the tuned random forest baseline on the classification suite for 1 hour on 8 cores, run:
\begin{lstlisting}
  $ python runbenchmark.py tunedrandomforest  openml/s/271 1h8c
\end{lstlisting}

\noindent Additional options may be used to specify the \emph{mode}, the \emph{parallelization}, and other details of the experimental setup.
For example, the following command may be used to evaluate the random forest baseline on the regression benchmark suite across 100 8-core \emph{aws} instances in parallel with a time budget of one hour.

\begin{lstlisting}
  $ python runbenchmark.py randomforest  openml/s/269 1h8c -m aws -p 100
\end{lstlisting}

\section{Benchmark Design \label{sec:design}}

In this section, we discuss both the design of the \textit{benchmark suite} (that is, the chosen data sets and evaluation procedures,~\citealp{bischl2021openml}) and the experimental setup, as well as their limitations.

\subsection{Benchmark Suites}

To facilitate a reproducible experimental evaluation, we make use of OpenML Benchmark suites~\citep{bischl2021openml}.
An OpenML benchmark suite is a collection of OpenML tasks, which each reference a data set, an evaluation procedure (such as k-fold cross-validation) and its splits, the target feature, and the type of task (regression or classification).
The benchmark suites are designed to reflect a wide range of realistic use cases for which the AutoML tools are designed.
Resource constraints are not part of the task definition. Instead, we define them separately in a local file so that each task can be evaluated with multiple resource constraints.
Both the OpenML benchmark suite (and tasks) and the resource constraints are machine-readable to ensure automated and reproducible experiments.

\subsubsection{Data Sets}
\label{sec:data sets}

We created two benchmarking suites, one with \numclassification{} classification tasks, and one with \numregression{} regression tasks.
The data sets used in these tasks are selected from previous AutoML papers \citep{AutoWEKA1}, competitions \citep{guyon2019analysis}, and ML benchmarks \citep{bischl2021openml} according to the following predefined list of criteria\footnote{The airlines datasets violate these criteria, but are included for historical reasons.}: 

\begin{itemize}
    \item \textbf{Difficulty} of the data set must be sufficiently high. 
    If a problem is easily solved by almost any algorithm, it will not be able to differentiate the various AutoML frameworks.
    This can mean that simple models (such as random forests, decision trees or logistic regression) achieve a generalization error of zero, or that the performance of these models and all evaluated AutoML tools is identical. 
    
    \item \textbf{Representative of real-world} data science problems to be solved with the frameworks. 
    In particular, we limit artificial problems. 
    We included a small selection of such problems, either based on their widespread use (\eg{} kr-vs-kp) or because they pose difficult problems, 
    but we do not want them to constitute a large part of the benchmark.
    We also limit computer vision problems on raw pixel data because those problems are better solved with dedicated deep learning solutions. 
    However, since they still make for real-world, interesting, and hard problems, we did not exclude them altogether.

    \item \textbf{No free form text features} that cannot reasonably be interpreted as a categorical feature.
    Most AutoML frameworks do not yet support feature engineering on text features and will process them as categorical features. 
    For this reason, we exclude text features, even though we admit their prevalence in many interesting real-world problems. 
    A first investigation and benchmark of multimodal AutoML with text features has been carried out by~\cite{shi2021benchmarking}.
    
    \item \textbf{Diversity} in the problem domains. 
    We do not want the benchmark to skew towards any application domain in particular. 
    There are various software quality problems in the OpenML-CC18 benchmark (jm1, kc1, kc2, pc1, pc3, pc4), but adopting them all would lead to a bias in the benchmark to this domain.
    
    \item \textbf{Independent and identically distributed} (i.i.d.) data are required for each task. 
    If the data are of temporal nature or repeated measurements have been conducted, the task is discarded.
    Both types of data are generally very interesting but are currently not supported for most AutoML systems, and we plan to extend the benchmark in the future in this direction.
    
    \item \textbf{Freely available} and hosted on OpenML.
    Data sets that can only be used on specific platforms or are not shared freely for any reasons are not included in the benchmark.

\end{itemize}
    
Reasons to exclude a data set included label-leakage and near-duplicates of other tasks in independent variables (for example, different only in categorical encoding or imputation) or dependent variable (most commonly the binarization of a regression or multi-class task).

\begin{figure}[bt]
    \centering
    \includegraphics[scale=0.33]{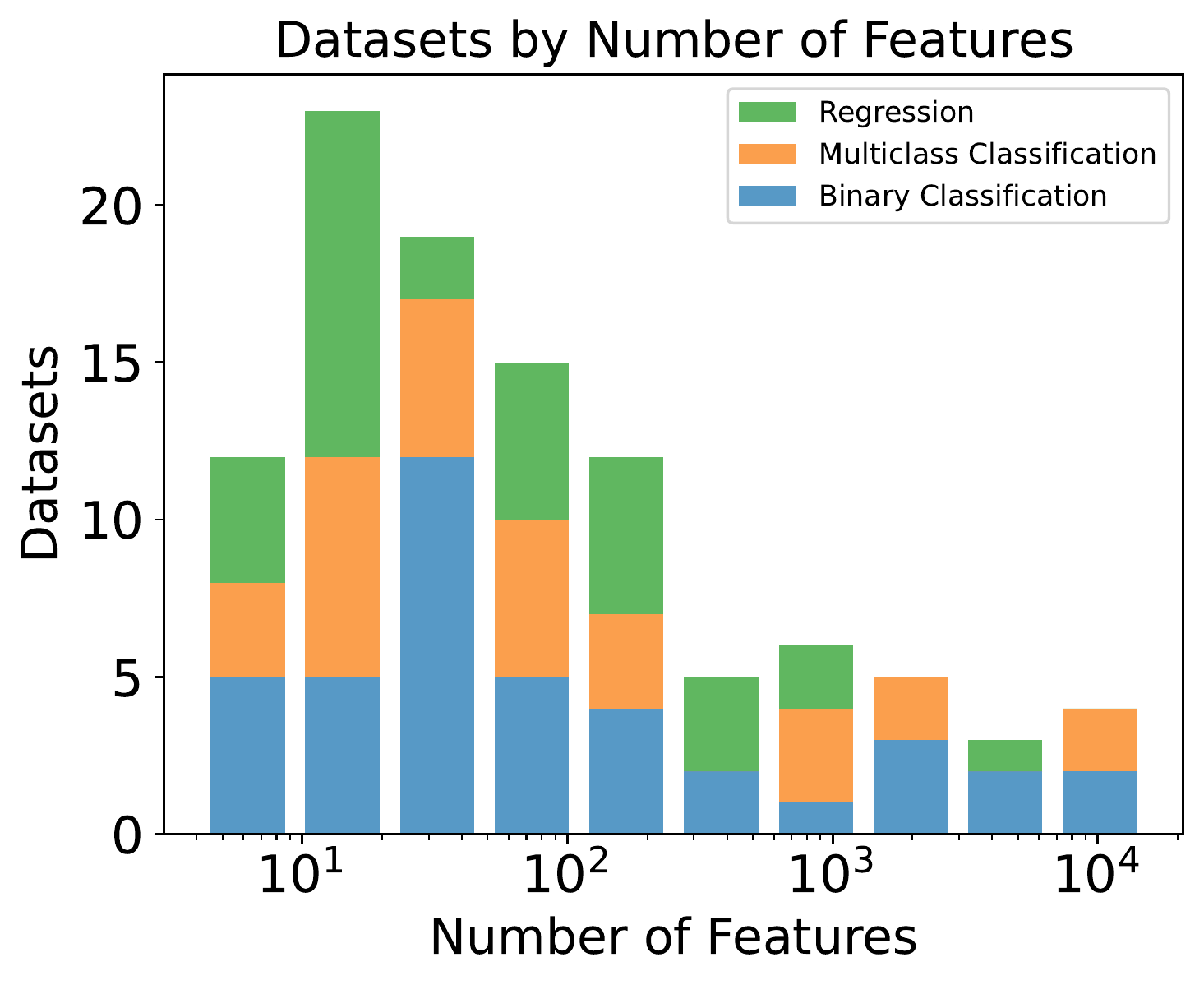}
    \includegraphics[scale=0.33]{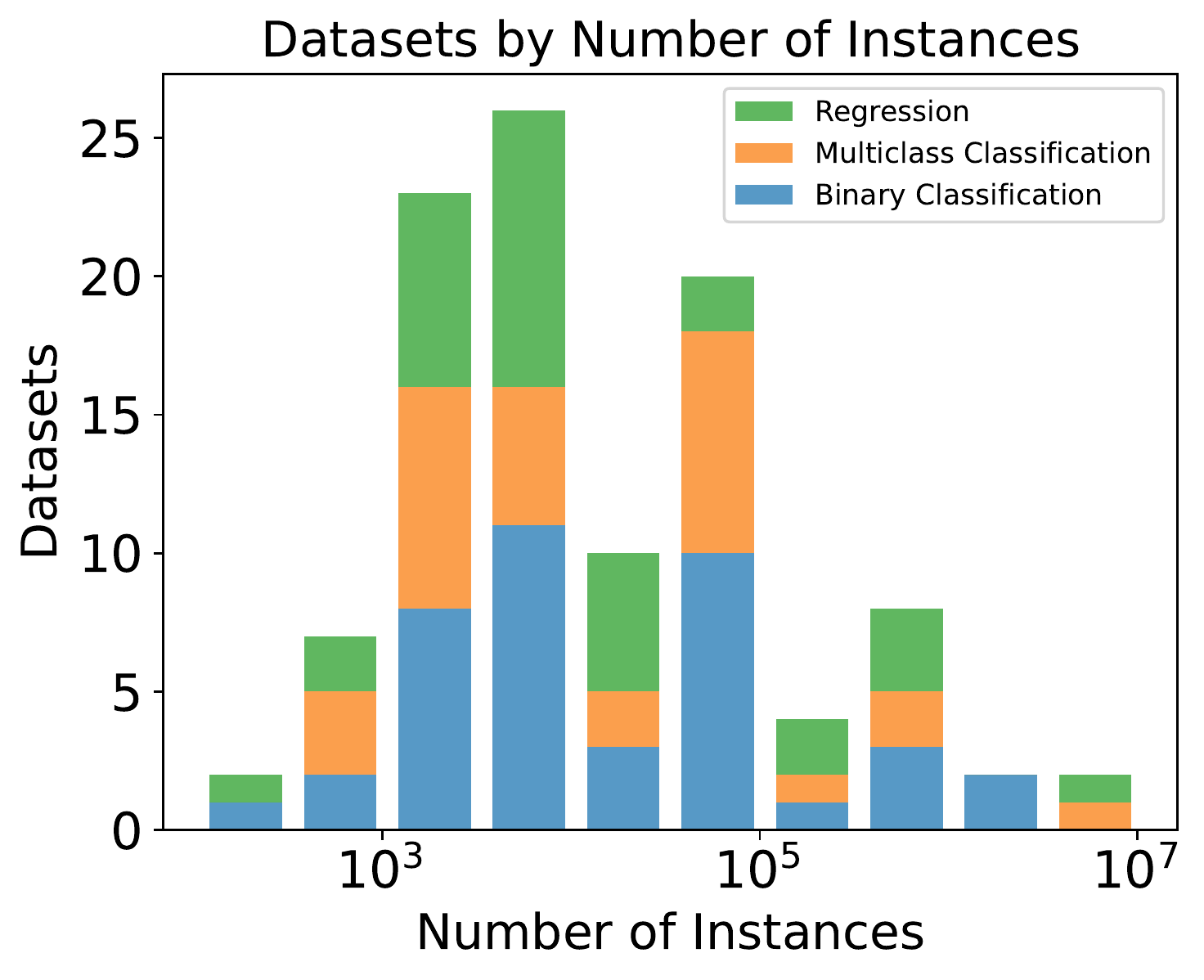}
    \includegraphics[scale=0.33]{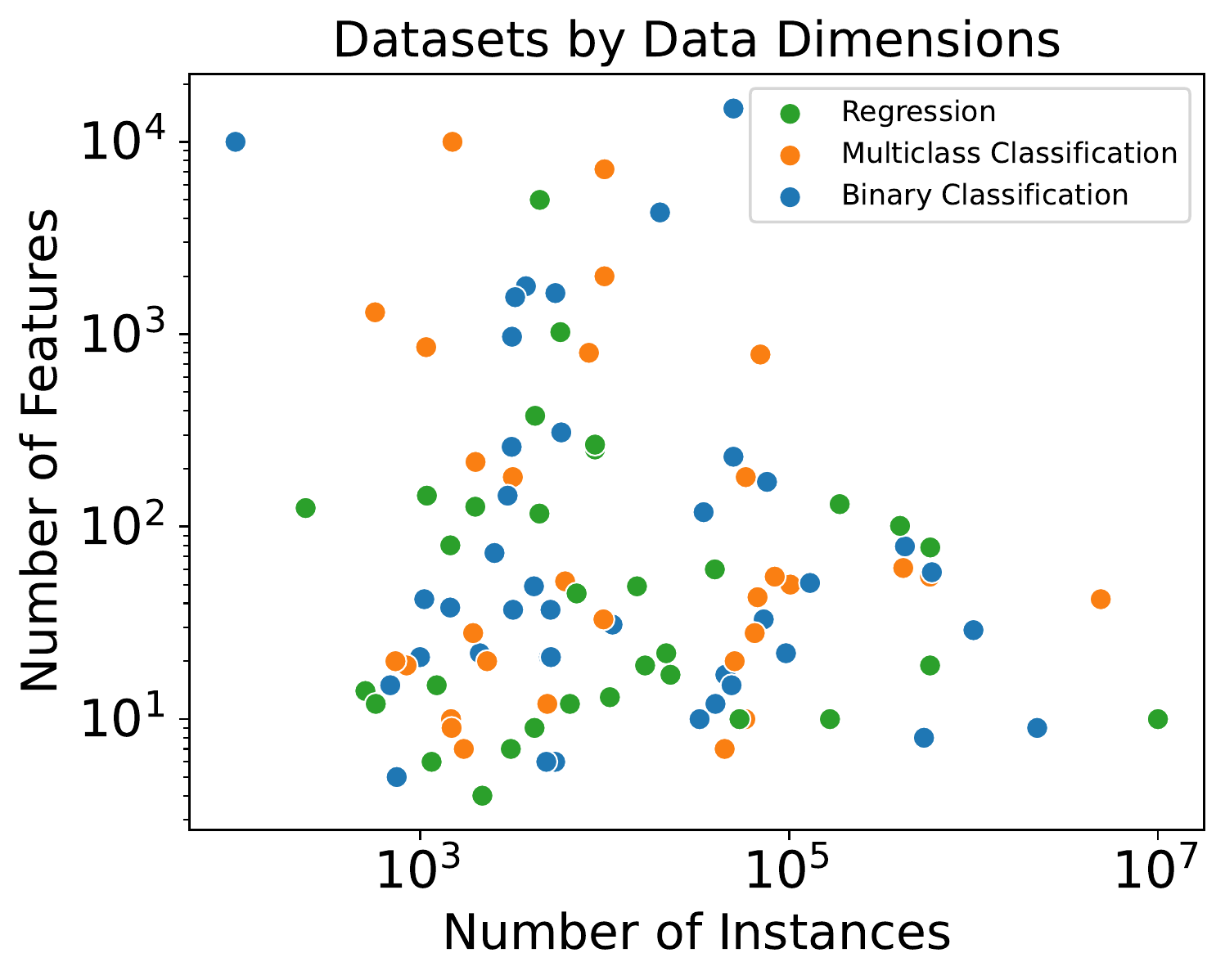}
    
    \includegraphics[scale=0.33]{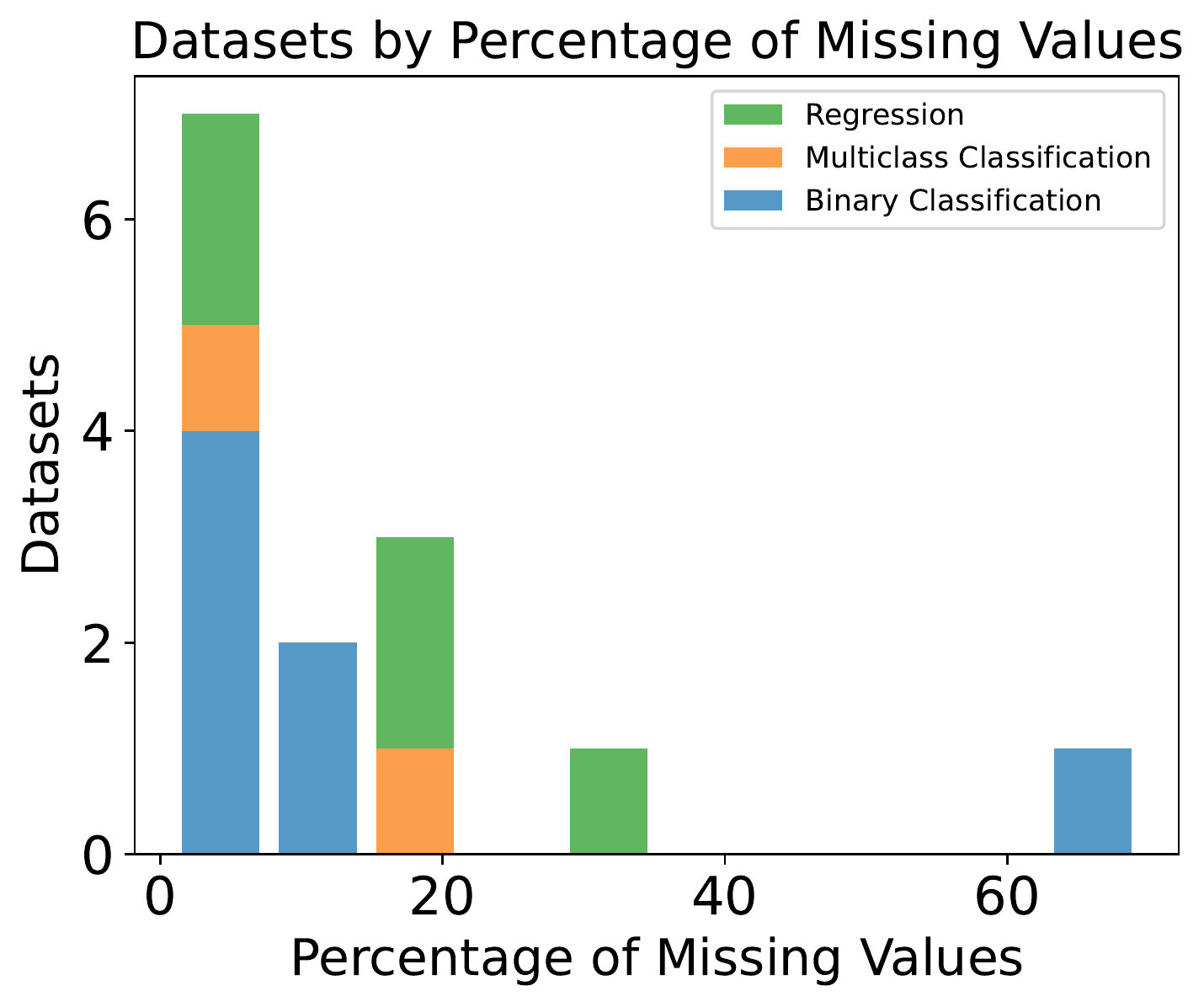}
    \includegraphics[scale=0.33]{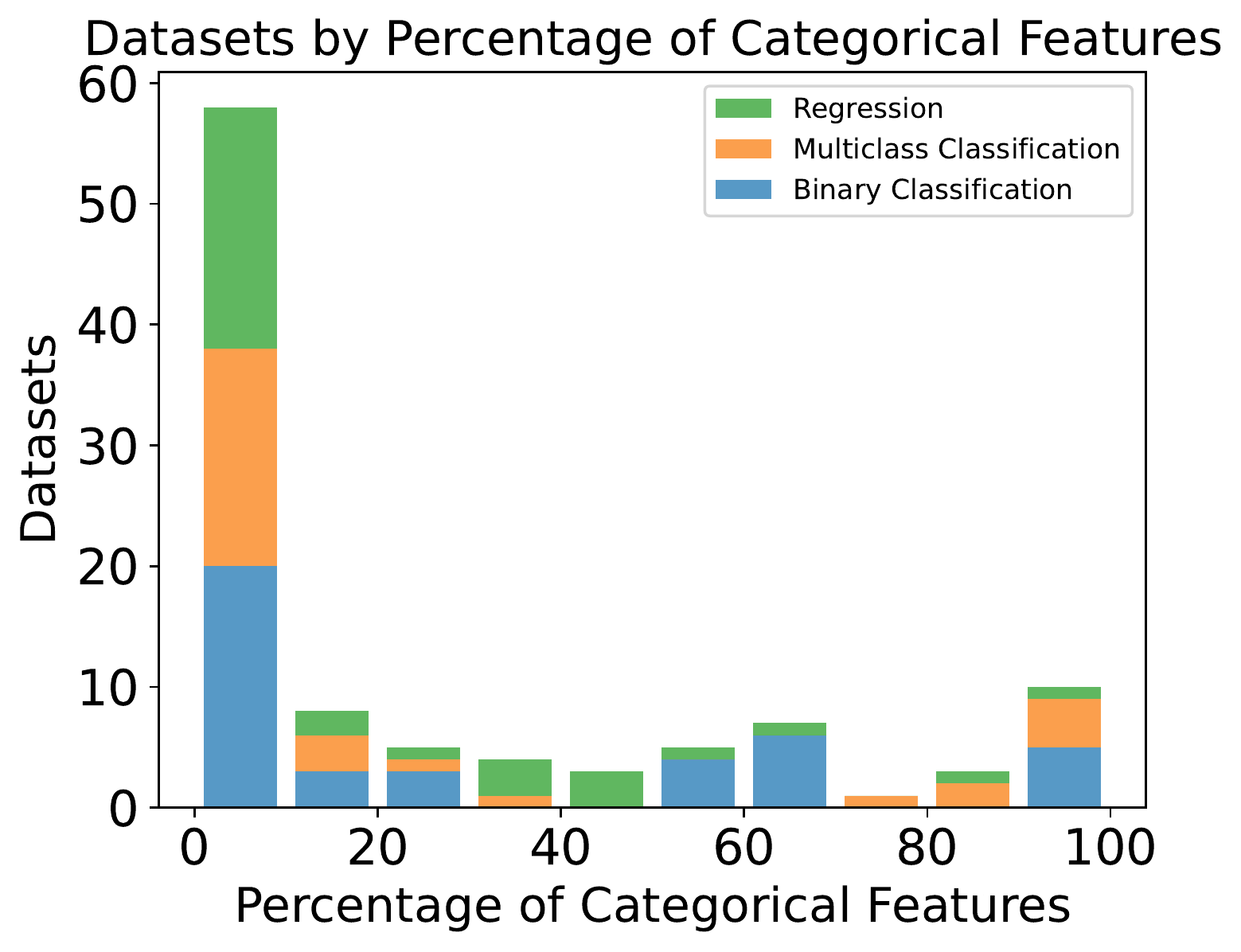}
    \includegraphics[scale=0.33]{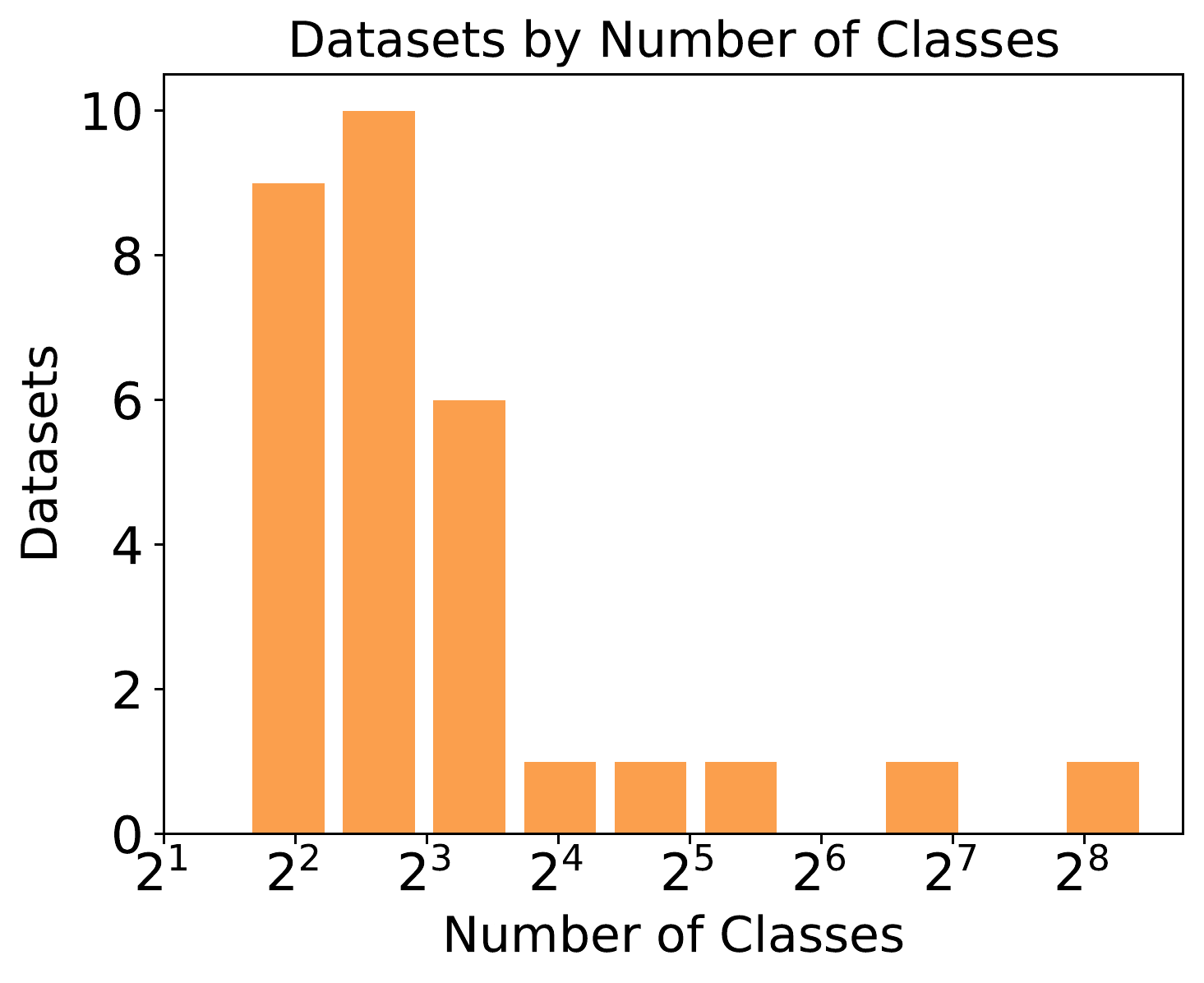}
    \caption{Properties of the tasks in both benchmarking suites.}
    \label{fig:stats_class}
\end{figure}

To study the differences between AutoML systems, the data sets vary in the number of samples and features by orders of magnitude and vary in the occurrence of numeric features, categorical features, and missing values. 
Figure~\ref{fig:stats_class} shows basic properties of the classification and regression tasks, including the distributions of the number of instances and features, the frequency of missing values and categorical features, and the number of target classes (for classification tasks).
Properties of the tasks are shown in Appendix~\ref{app:suites} and can be explored interactively on OpenML.\footnote{Visit \url{www.openml.org/s/269} for regression and \url{www.openml.org/s/271} for classification.}
While the selection spans a wide range of data types and problem domains, we recognize that there is room for improvement.
Restricting ourselves to open data sets without text features severely limits options, especially for big data sets. We hope to address this in future work. 

All data sets are available in multiple formats for the AutoML frameworks, either as files (parquet, arff, or csv) or as Python objects (pandas dataframe, numpy array), and the used format depends on the AutoML framework. 
All frameworks have access to meta-data, such as the data type of the columns, either directly from the chosen data format or as separate input, so that each AutoML framework has the same information available regardless of the chosen data format.

\subsubsection{Performance Metrics}
In our evaluation, we use area under the receiver operating characteristic curve (AUC) for binary classification, \logl{} for multi-class classification and root mean-squared error (\rmse) for regression to evaluate model performance.\footnote{We use the implementations provided by \systemcase{scikit-learn} 1.2.2.}
We chose to use these metrics because they are generally reasonable, commonly used in practice, and supported by most AutoML tools.
The latter is especially important because it is imperative that AutoML systems optimize for the same metric on which they are evaluated.
However, our tool is not limited to only these three metrics, and a wide range of performance metrics can be specified by the user.

Models are not calibrated, unless the AutoML framework does so by default.
Whether or not calibrated models are important is highly dependent on the use case, and most AutoML frameworks do not support model calibration out of the box.
This is especially important to keep in mind when we discuss results on our binary classification problems, as AUC is not sensitive to calibration.

\subsubsection{Missing Values In Experimental Results}
As will be discussed in more detail in Section~\ref{sec:errors}, not all frameworks are equally well-behaved.
There are situations where search time budgets are exceeded or the AutoML frameworks crash outright, which results in missing performance estimates.
There are multiple strategies to consider on how to deal with these missing data.

One naive approach may be to ignore missing values and aggregate over the obtained results.
However, we observe that failures do not occur at random. 
Failures correlate with data set properties, such as data set size and class imbalance, which may be correlated with ``problem difficulty" and thus performance.
Ignoring missing values thus means that AutoML frameworks may fail on harder tasks or folds and consequently obtain higher average performance estimates.
Imputing missing values with performance obtained by the same AutoML framework on other folds is subject to the same drawback.
Moreover, in case a framework fails to produce predictions on all folds of a task, neither method specifies how to deal with missing values.

Instead, we propose to impute the missing values with an interpretable and reliable baseline.
An argument may be made for using the random forest baseline, since this may be a strong fallback that AutoML frameworks could realistically implement.
However, we observe that training a random forest (of the size used in the baseline) requires a significant amount of time on larger data sets.
Automatically providing this fallback by means of imputation would provide an unfair advantage to the AutoML frameworks that are not well-behaved.
Moreover, many failures would not be remedied by having a random forest to fall back on, since the AutoML frameworks crash irrecoverably, for example, due to segmentation faults.

Therefore, we impute missing values with the constant predictor, or prior.\footnote{We except one specific error for \autogluon{}(HQIL) which is already fixed (c.f. Appendix~\ref{sec:app-errors}).}
This baseline returns the empirical class distribution for classification and the empirical mean for regression.
This is a very penalizing imputation strategy, as the constant predictor is often much worse than results obtained by the AutoML frameworks that produce predictions for the task or fold.
However, we feel this penalization for ill-behaved systems is appropriate and fairer towards the well-behaved frameworks and hope that it encourages a standard of robust, well-behaved AutoML frameworks.

\subsection{Experimental Setup}
We execute the experiments on commodity-level hardware with AutoML frameworks generally in their default configurations.

\subsubsection{Hardware}
As discussed in Section~\ref{sec:software}, \amlb{} can be run on any machine. However, for comparable hardware and easy expandability, we opt to conduct the benchmark on standard \texttt{m5.2xlarge}\footnote{More information is available at \url{https://aws.amazon.com/ec2/instance-types/m5/}.} instances available on Amazon Web Services (AWS).
These represent current commodity-level hardware with $32$~GB memory, $8$~vCPUs (Intel Xeon Platinum $8000$ series Skylake-SP processor with a sustained all core Turbo CPU clock speed of up to 3.1 GHz).
$100$~GB of gp3-SSD storage is available for storage, which can be necessary for storing a larger number of evaluated pipelines.
The use of AWS also enables others to fully reproduce and extend our results, as funding permits, since the results do not depend on private computing infrastructure.

\subsubsection{Framework Configuration}
AutoML frameworks are instantiated with their default configuration, except that we control the following settings:

\begin{itemize}
    \item \textbf{`Mode'} to declare the user intent. For example, obtaining the best possible model versus finding an interpretable (less complex) model. The modes used to evaluate each AutoML framework is chosen by their developers.
    \item \textbf{Runtime} for the search. Additionally, there is one hour leeway for data loading, making predictions, and cleanup operations, but this is not communicated to the AutoML frameworks.
    \item \textbf{Resource constraints} that specify the number of CPU cores and amount of memory available.
    \item \textbf{Target metric} to use for optimization. This is the same metric that is used for evaluation in the benchmark.
\end{itemize}

The experiment design intentionally prohibits further customization of other AutoML system configuration parameters to reflect how these systems are usually applied in practice as closely as possible (note that \amlb{} does allow for this type of customization). 
An overview of used framework versions, their `mode' configurations, and important integration details are provided in Appendix~\ref{app:framework-versions}.

\subsection{Limitations}
\label{sec:limits}
Both the design of the benchmark and the setup for the experiments described in this paper have some limitations with regard to the interpretation of their results.
Limitations in the design stem from the desire to keep the use of the frameworks as close as possible to the original vision and usage intended by developers, whereas the limitations in the experiments are caused by resource constraints and may be alleviated by running additional experiments with the benchmark software.
In this section, we highlight some important limitations and stress that this paper and the results within do \emph{not} state which AutoML framework is ultimately the best.

\subsubsection{Limitations of the Design}
Perhaps the biggest limitation of the design is the inability to attribute the performance of an AutoML framework to any one aspect of its design, as is often done with ablation studies.
The evaluated AutoML frameworks differ among multiple design choices, such as the underlying ML library, search space, preprocessing, and search algorithm.
Concretely, a performance difference between, for instance, \autosklearn{} and \tpot{} could be caused by \tpot{}'s built-in stacking, \autosklearn{}'s ensembles, the difference in Bayesian optimization versus genetic programming, the difference in how multiprocessing is employed, or a combination of these or any other difference between them. 
Software that would allow for such conclusions essentially requires each AutoML framework to be reimplemented on a shared set of algorithms for building models, search, and evaluation.
We acknowledge that this would be incredibly valuable for the research community. However, it would also no longer resemble the software as used in practice and thus would be different work altogether.
Note that it \emph{is} possible to perform ablation studies with \amlb{} for a specific AutoML framework, for example by comparing different framework configurations as done by~\cite{autogluontabular}.

Another limitation stems from only recording results produced by the final model.
Anytime performance, where information about the performance during optimization is captured as if they were final models, can be very insightful.
This method allows for the distinction between a framework that converges quickly from one that does not. This may be especially important for users who are interested to use the systems with a human-in-the-loop, such as when designing a search space or data features.
Unfortunately, many frameworks do not support collection of anytime performance and--- depending on how they are recorded---might interfere with resources used during search.
We hope to be able to record anytime performance in the future, but in this work, we only approximate it by evaluating the tools under two different time constraints ($1$ and $4$ hours).

Finally, the qualitative comparison of the frameworks is also limited.
Certain "quality of life features" like analysis of the pipeline via interpretable ML methods, reports, usability, or support are not evaluated here but are important to many users.
For a qualitative analysis of those characteristics, we refer the reader to one of the many existing overview papers on AutoML~\citep{ZollerH21,TruongWGHBF19}. We also provide an overview of various AutoML frameworks with links to their documentation on our website.\footnote{\url{https://openml.github.io/automlbenchmark/frameworks.html}}

\subsubsection{Limitations of the Experiments}
Most frameworks are highly configurable and allow the user to configure the search algorithm or its hyperparameters, among other aspects that affect the AutoML performance.
Some frameworks even provide different configuration presets for different use cases, such as a performance-oriented competition mode and a mode that produces fast or interpretable models at the cost of some performance.
However, comparing the effect on model performance of tuning AutoML hyper-hyperparameters quickly carries prohibitive costs.
We evaluated only a few modes for each framework, if any, based on the developers' recommendations.
It is likely that better results may be achieved by carefully meta-tuning the AutoML framework or evaluating additional modes.
While it is cost prohibitive for us to evaluate many different scenarios, it is easy to run the benchmark with custom configurations for the various AutoML frameworks.
This allows users to evaluate AutoML frameworks in a setting that reflects their interest.

\subsubsection{Meta-learning}\label{sec:meta-learning}
Many AutoML frameworks make use of meta-learning to better initialize and speed up the search~\citep{oboe, feurer2015efficient, feurer2020autosklearn}. %meta-tpot
Since all data in the benchmark is publicly available and many of them are well known in the AutoML community, it is likely that there is a substantial overlap between data used by the developers for meta-learning and the data used in the benchmark.
This is a very intricate problem, as we consider AutoML frameworks as black boxes.
Removing the effect of the data set that is to be evaluated from the meta-learning procedure is not solvable in general.    

In this paper specifically, both \autosklearn{} and \autosklearnii{} use meta-learning.\\*
\autosklearn{}'s meta-learning uses $208$ data sets from OpenML, each associated with well-performing ML pipelines.
The search is initialized with $k$-nearest datasets \citep{reif2012meta} using the $25$-nearest datasets based on $38$ meta-features and looking up the best known pipeline for those datasets.
\autosklearn{} can exclude data sets by name from the lookup, which we make use of in \amlb{} to ensure that there is no overlap to the $39$ data sets from \cite{amlb2019}.
Even so, it cannot be guaranteed that identical data sets with a different name might be used for meta learning out-of-the-box. 

\autosklearnii{}'s meta-learning model is more complicated and consists of:
a) a static pipeline portfolio for warm-starting the search, which is computed across hundreds of data sets using a greedy forward selection,
and b) a meta-model to predict the internal model selection strategy and budget allocation strategy.
Single data sets cannot be excluded from these meta-learning procedures, and it is not feasible to retrain the meta-models and pipeline portfolio for each data set in our benchmark.
This ultimately means that the result of \autosklearnii{} must be considered very carefully and are likely optimistic, as their meta-model is built with information about many datasets in this benchmark.
More thought and collaboration is required to address these issues and allow for the correct evaluation of AutoML systems that use meta-learning in common benchmarks.

\subsection{Overfitting the Benchmark}

One last issue that plagues any widely-adopted benchmark is the potential of algorithms overfitting on the data sets used in the benchmark.
Since freely available, interesting, and usable (refer to Section~\ref{sec:data sets} for our selection criteria) data sets are scarce, many AutoML developers use these data sets to benchmark and then improve their systems iteratively.
While this is not as direct of an issue as with meta-learning, these data sets can in general not be assumed to be truly unseen.
The only practical way to avoid this is to collect a novel set of data sets for the benchmark, which would entail a prohibitive effort.
Moreover, after publishing such a benchmark, the new data sets are published, which again gives developers the possibility to use them to improve their systems.
On the other hand, should the benchmarking data sets be kept private to avoid this issue, the benchmark is no longer entirely reproducible by independent researchers.
We hope that the size of our benchmarking suites is large enough and their design general enough that overfitting is less of an issue, but this is difficult to guarantee, and a study as outlined above may be useful to evaluate this phenomenon in the future.

\section{Results \label{sec:results}}

In this section, we provide an overview and analysis of the results obtained.
This section is accompanied by an interactive visualization tool\footnote{Results can be interactively explored at \visapp{}.}, additional information in Appendix~\ref{app:results}, and all data artifacts generated from these experiments.\footnote{Experiment artifacts can be found at: \experimentdata{}.}
For a more comprehensive comparison than what we can provide here, we strongly encourage the reader to explore the data with the interactive visualization tool.
This tool enables users to select any subset of frameworks, task types, performance measures, or data characteristics iteratively and interactively.
The tool includes overview plots for different task types as well as detailed visualizations for individual data sets.
Moreover, a statistical analysis of the results by critical difference plots and Bradley-Terry trees is implemented.

In addition to the limitations outlined in section~\ref{sec:limits}, we also want to stress that these experimental results are obtained by running the frameworks as they were in June of 2023.
Some of these frameworks are still under active development, and results from experiments run on later versions will almost certainly differ.

We used recent results as much as possible, but we also include some results of experiments ran of frameworks in September of 2021. In these cases, the benchmarked framework versions were not substantially different.
We motivate this decision in Appendix~\ref{app:framework-versions} and denote the older results with an asterisk (*) by the framework name. For example, \tpot{} denotes the June 2023 version and \tpot{}* denotes the September 2021 version.
Some results are missing: \autosklearnii{} does not support regression, and we could not complete our 4 hour classification evaluation of \autogluon{}\systemcase{(HQIL)} due to budget constraints.

We strongly encourage people to run additional experiments that match their use case with up-to-date frameworks, and use the results in this section as a reference on how to analyze the results. Results may differ substantially for different time constraints, amount of parallel processing available, availability of GPU's, and more.
Table~\ref{table:framework-versions} in Appendix~\ref{app:framework-versions} shows an overview of the versions benchmarked for each framework as well as the most current version.

\subsection{Predictive Performance}
\label{sec:performance}
\begin{figure}
     \centering
     \begin{subfigure}[b]{0.48\textwidth}
         \centering
         \includegraphics[width=\textwidth]{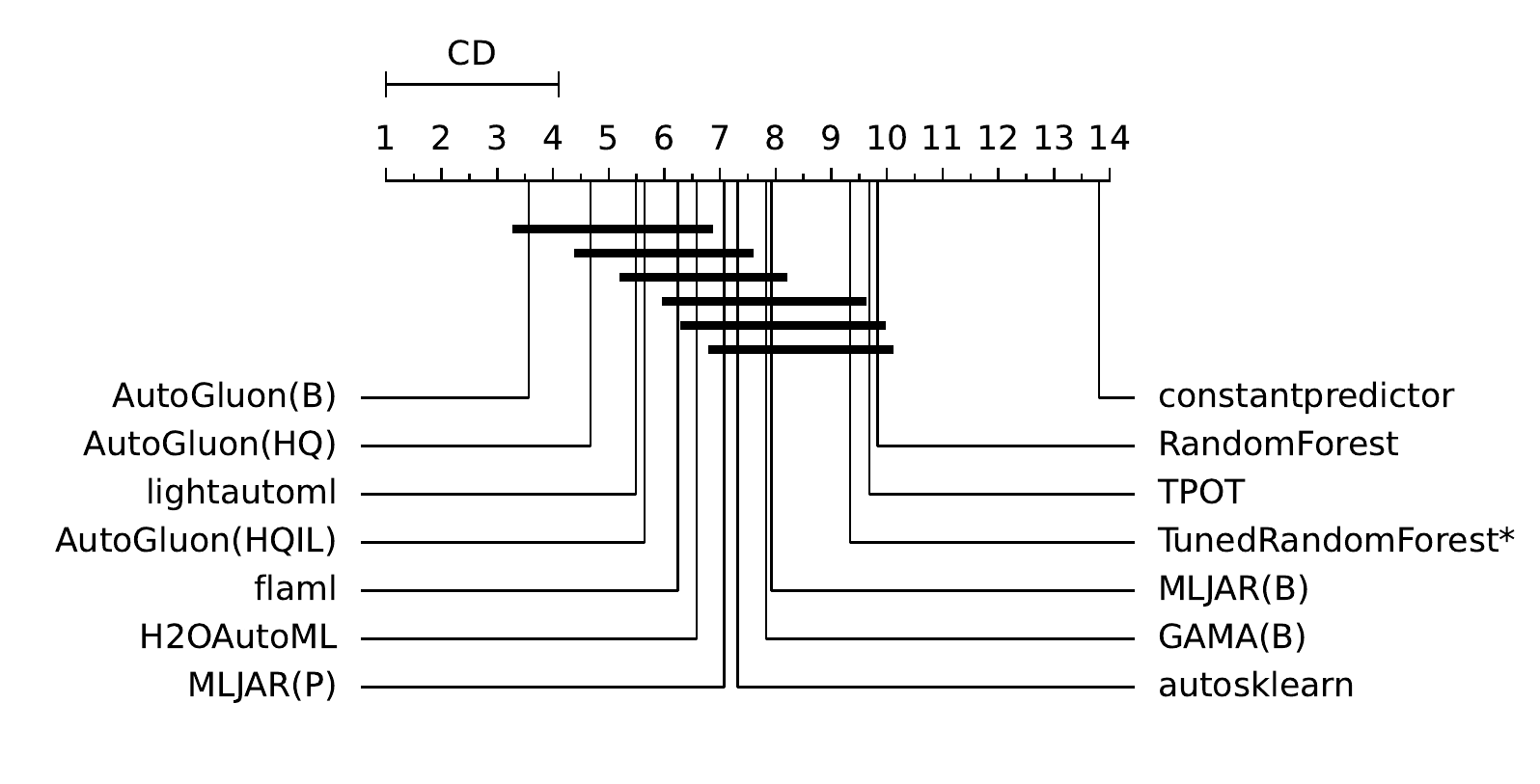}
         \caption{Binary Classification, 1 hour}
     \end{subfigure}
     \hfill
     \begin{subfigure}[b]{0.48\textwidth}
         \centering
         \includegraphics[width=\textwidth]{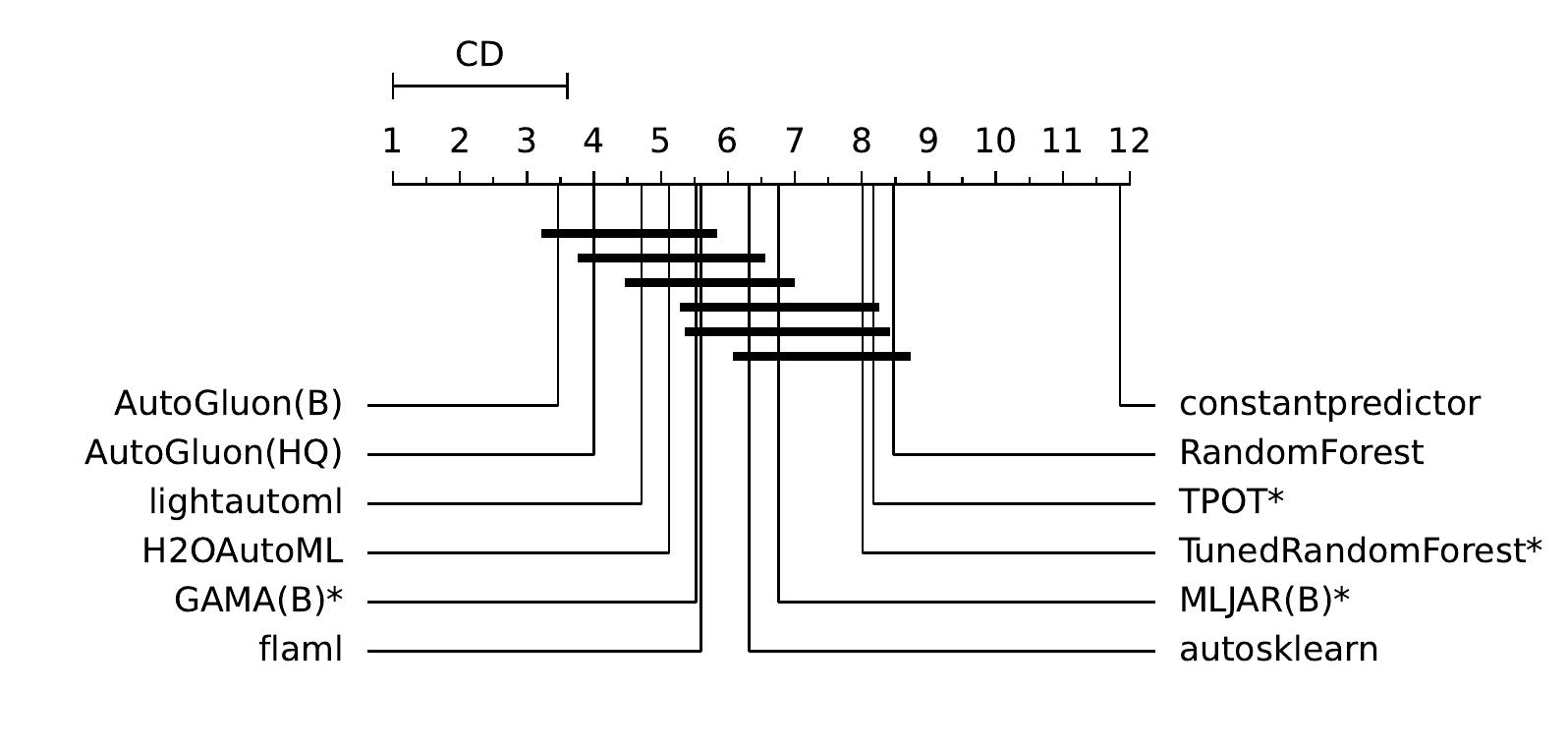}
         \caption{Binary Classification, 4 hours}
     \end{subfigure}
     \hfill
     \begin{subfigure}[b]{0.48\textwidth}
         \centering
         \includegraphics[width=\textwidth]{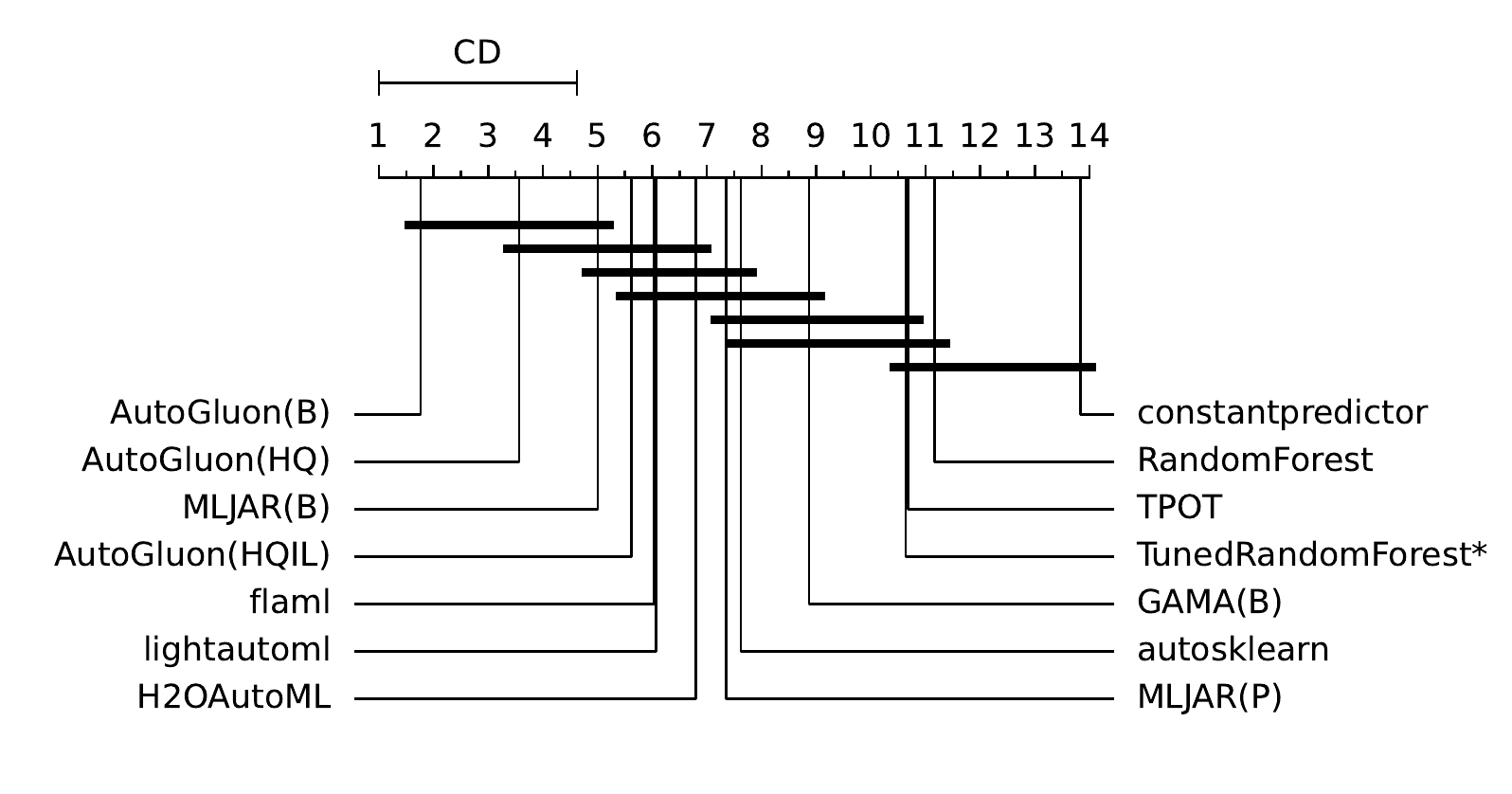}
         \caption{Multi-class Classification, 1 hour}
     \end{subfigure}
     \hfill
     \begin{subfigure}[b]{0.48\textwidth}
         \centering
         \includegraphics[width=\textwidth]{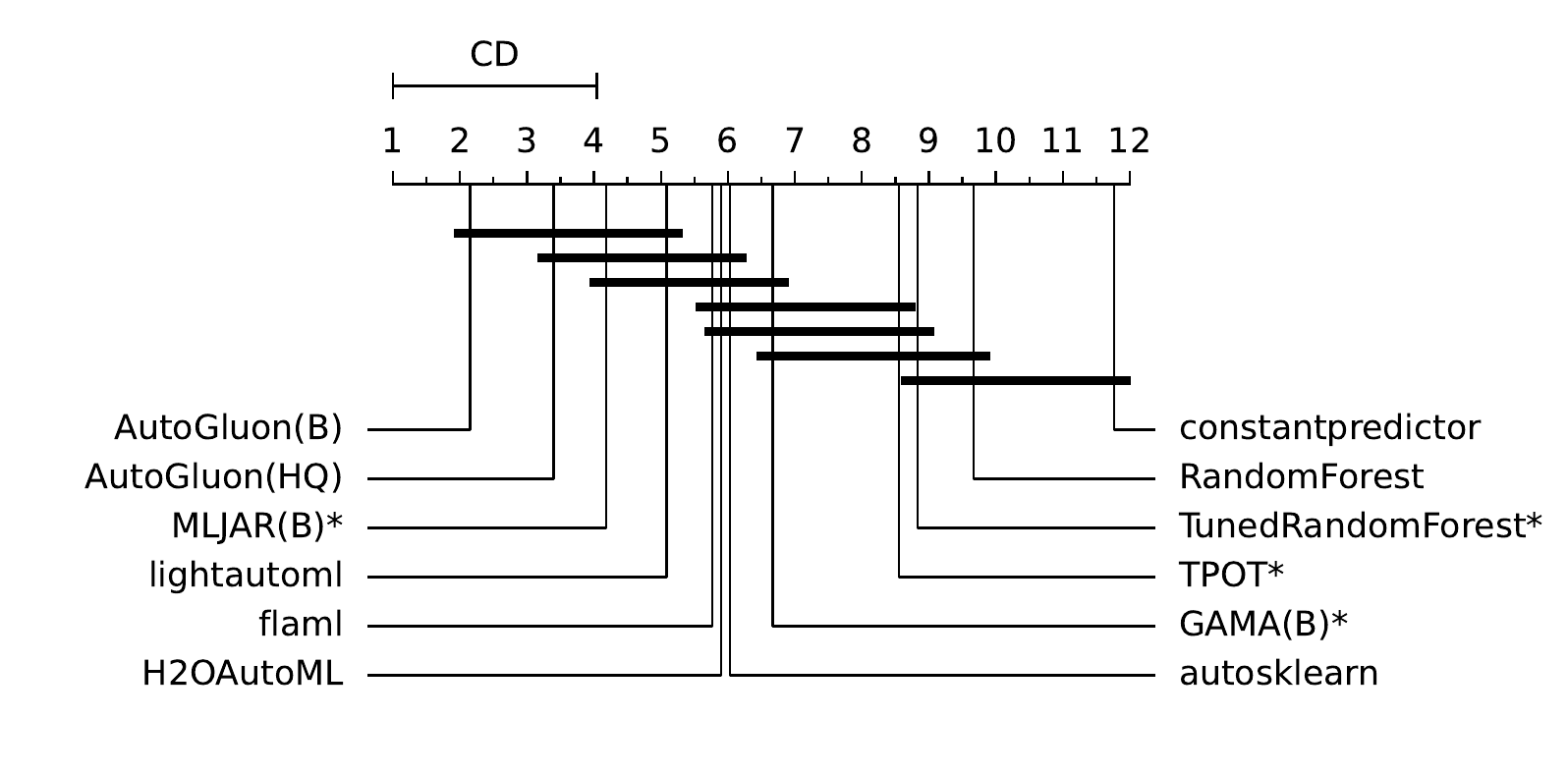}
         \caption{Multi-class Classification, 4 hours}
     \end{subfigure}
     \hfill
          \begin{subfigure}[b]{0.48\textwidth}
         \centering
         \includegraphics[width=\textwidth]{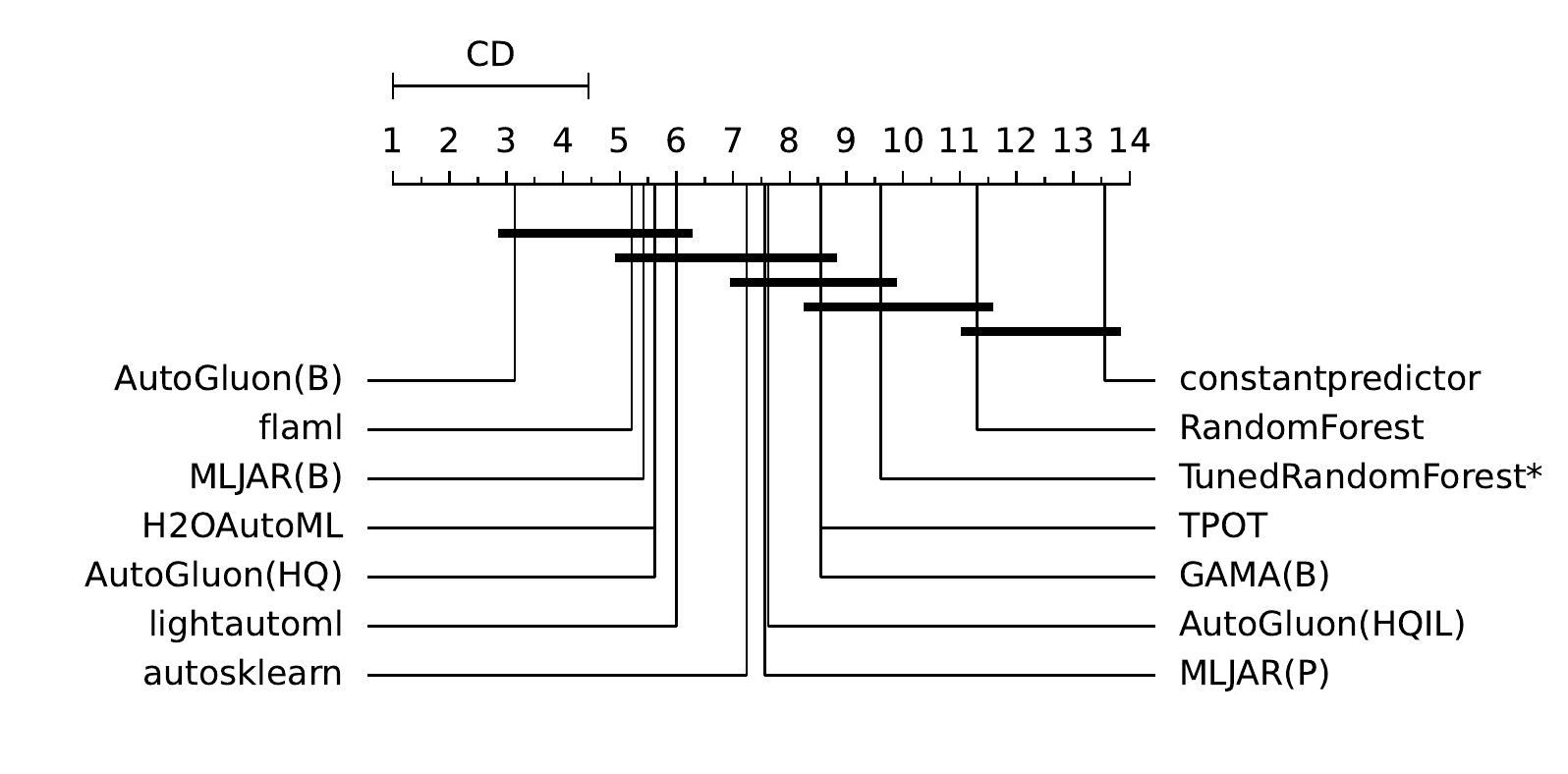}
         \caption{Regression, 1 hour}
     \end{subfigure}
     \hfill
     \begin{subfigure}[b]{0.48\textwidth}
         \centering
         \includegraphics[width=\textwidth]{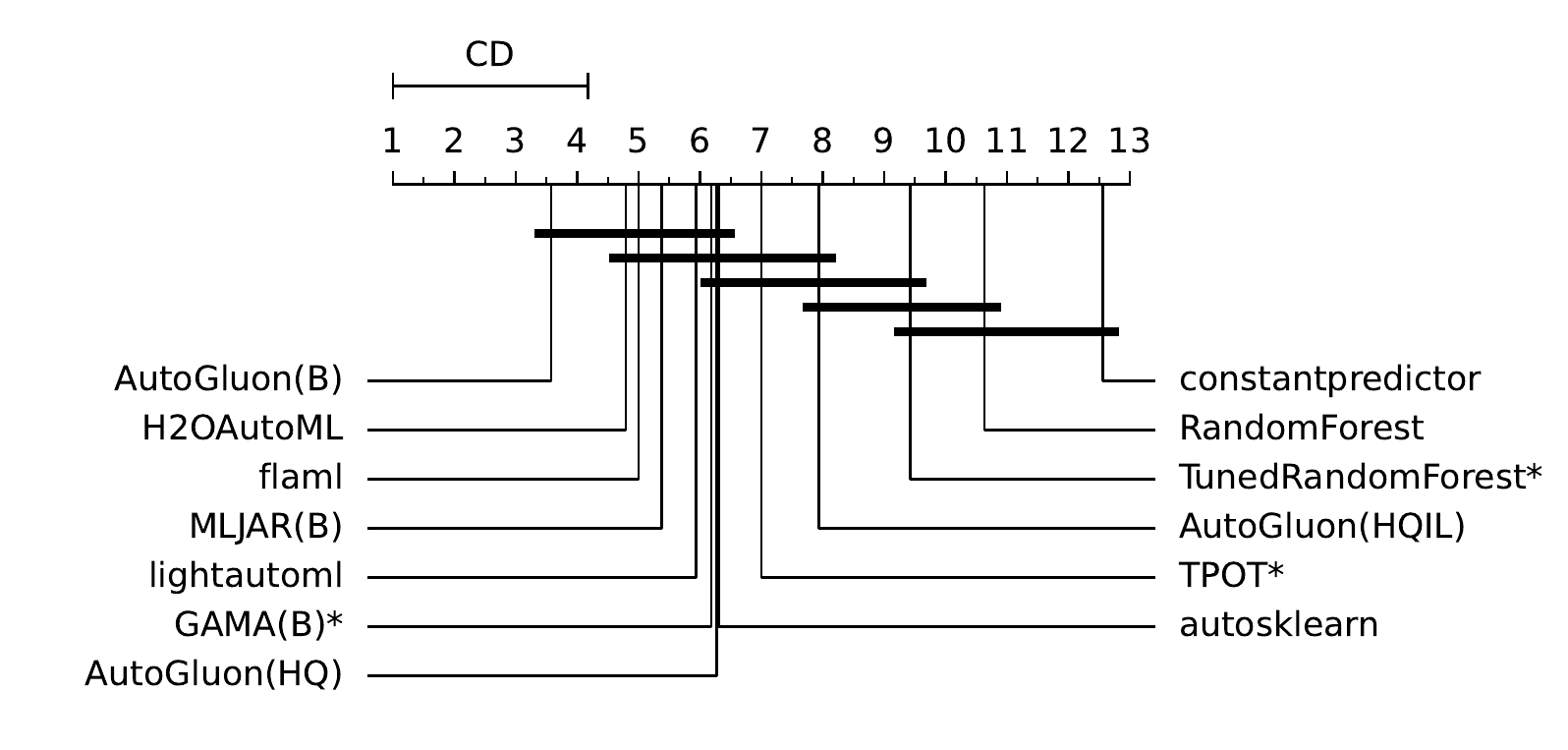}
         \caption{Regression, 4 hours}
     \end{subfigure}
          \hfill
          \begin{subfigure}[b]{0.48\textwidth}
         \centering
         \includegraphics[width=\textwidth]{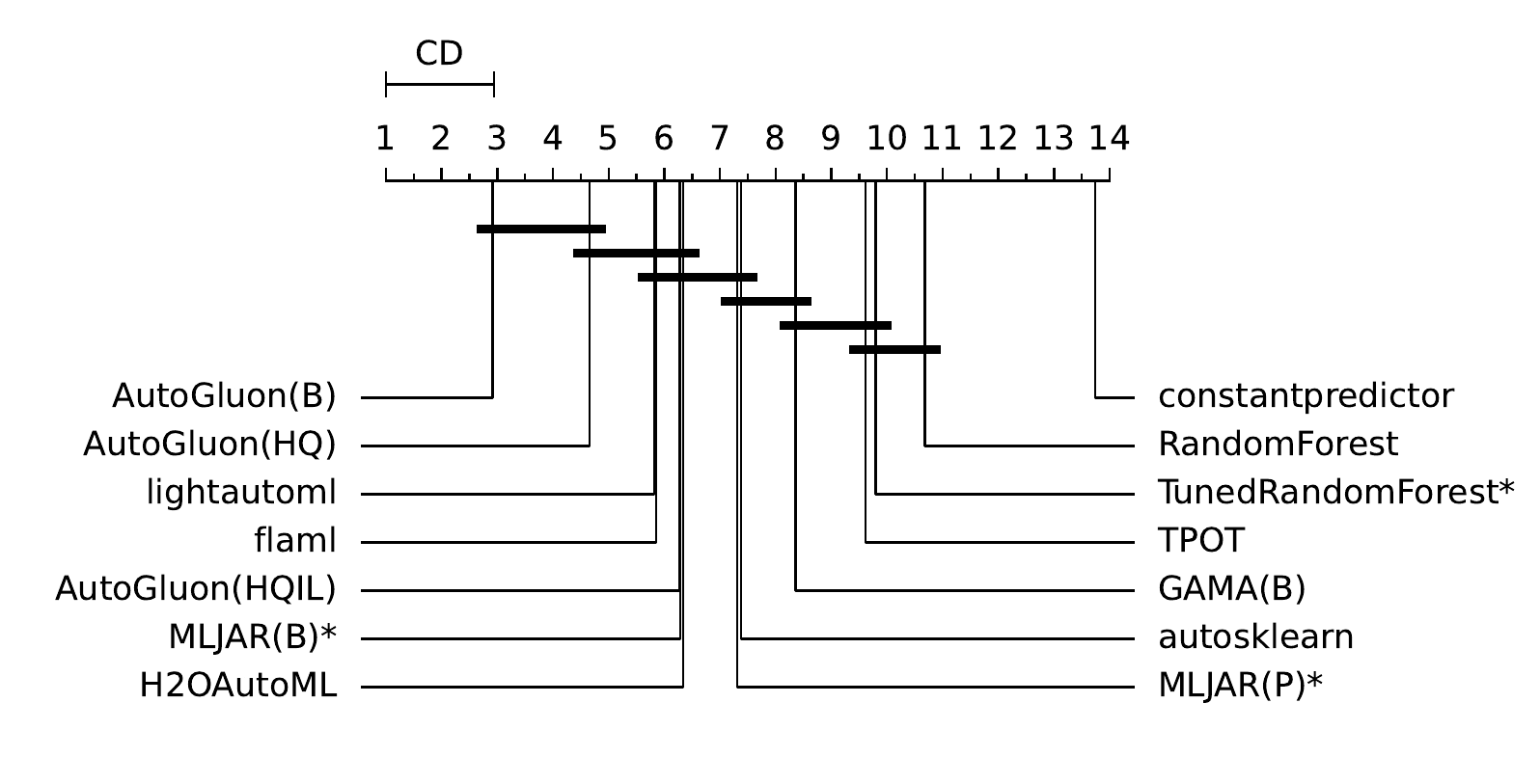}
         \caption{All tasks, 1 hour}
         \label{fig:cd-all-1h}
     \end{subfigure}
     \hfill
     \begin{subfigure}[b]{0.48\textwidth}
         \centering
         \includegraphics[width=\textwidth]{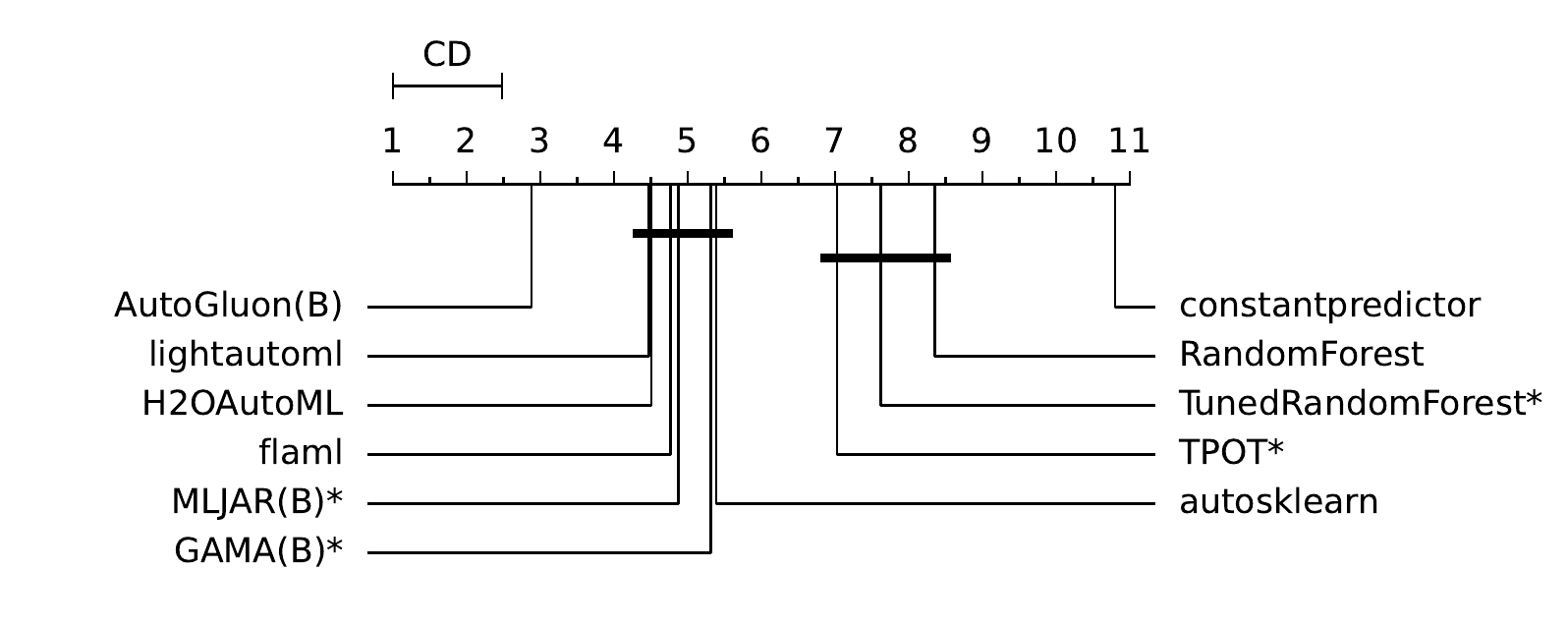}
         \caption{All tasks, 4 hours}
         \label{fig:cd-all-4h}
     \end{subfigure}
     \caption{CD plots with Nimenyi post-hoc test after imputing missing values with the constant predictor baseline. An asterisk (*) next to a framework name means the results were obtained in 2021.}
     \label{fig:main-cd-plot}
\end{figure}

To report on the results for many AutoML frameworks across whole benchmarking suites, we propose using critical difference (CD) diagrams~\citep{Demsar2006} illustrated in Figure~\ref{fig:main-cd-plot}.
In a CD diagram, the average rank of each framework as well as which ranks are statistically significantly different from each other are shown.
To calculate the average rank per task, we first impute any missing values with the constant predictor and then average the performance over all folds.
We may then test for the presence of statistically significant differences in the average rank distributions using a non-parametric Friedman test at $p < 0.05$ (here, $p \approx 0$ for every diagram) and use a Nemenyi post-hoc test to find which pairs differ. 
For each benchmarking suite and time budget, the CD diagrams are shown in Figure~\ref{fig:main-cd-plot}, which displays the rank of each framework (lower is better) averaged over all results from the given benchmarking suite and budget.
\autosklearnii{} is excluded in this comparison due to the meta-learning issues discussed in Section~\ref{sec:meta-learning}.

Overall, we observe that \autogluon{}\systemcase{(B)} and \tpot{} respectively achieve the best and worst rank among AutoML frameworks in almost every setting with respect to model accuracy. However, while \autogluon{}\systemcase{(B)} is generally not significantly better than the second best framework, it does rank statistically significantly better than the worst AutoML frameworks in any given scenario. Similarly, \tpot{} is never significantly worse than the second worst framework, but its rank is always statistically significantly worse than \autogluon{}\systemcase{(B)}.  
In all cases, the baselines obtain worse ranks than any AutoML framework except \tpot{}, although the tuned random forest is a strong baseline. Only \autogluon{}\systemcase{(B)} and \lama{} achieve significantly better ranks across all settings.
All AutoML frameworks except \autogluon{}\systemcase{(B)} and \tpot{} are generally ranked close to each other, with small differences across the various suites and budgets.

\begin{figure}[t!]
     \centering
     \begin{subfigure}[b]{0.48\textwidth}
         \centering
         \includegraphics[width=\textwidth]{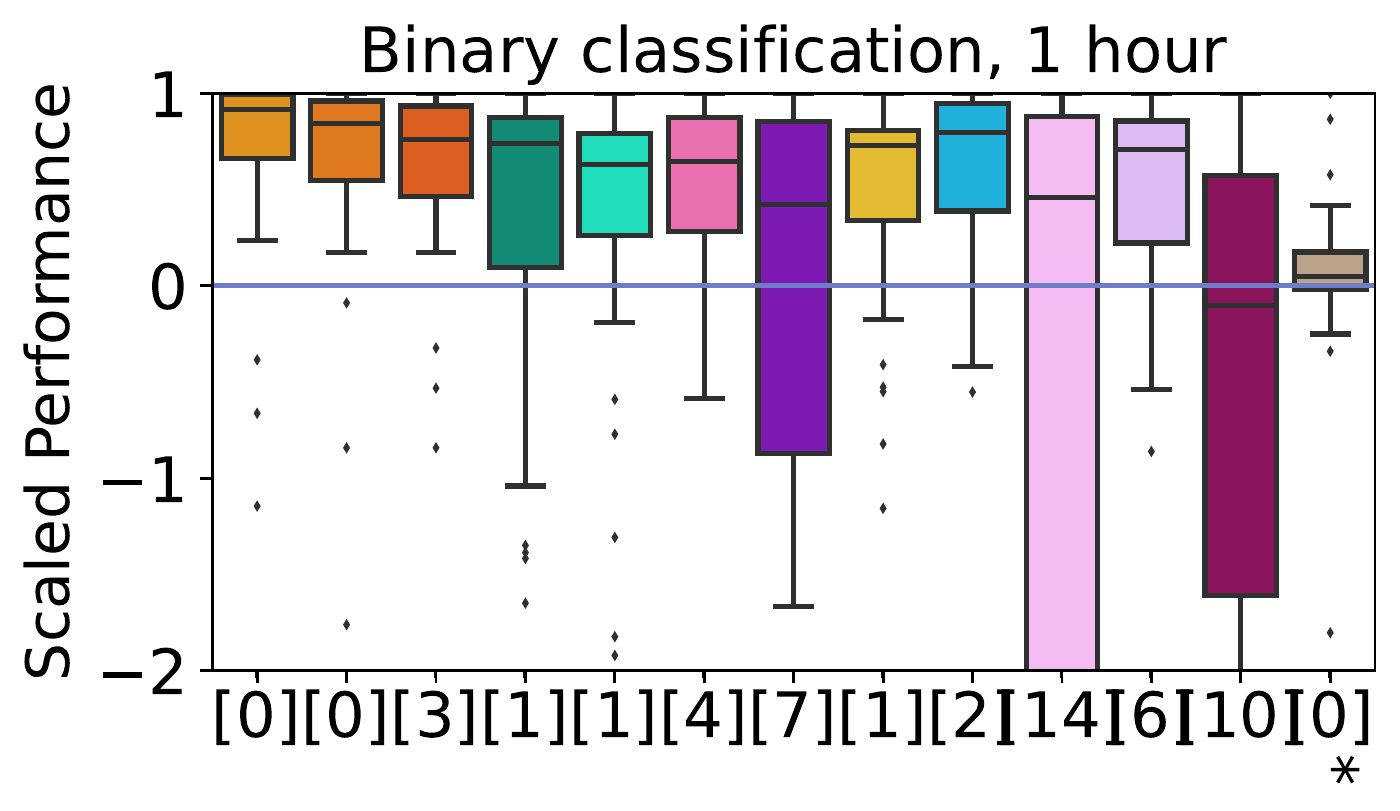}
     \end{subfigure}
     \hfill
     \begin{subfigure}[b]{0.48\textwidth}
         \centering
         \includegraphics[width=\textwidth]{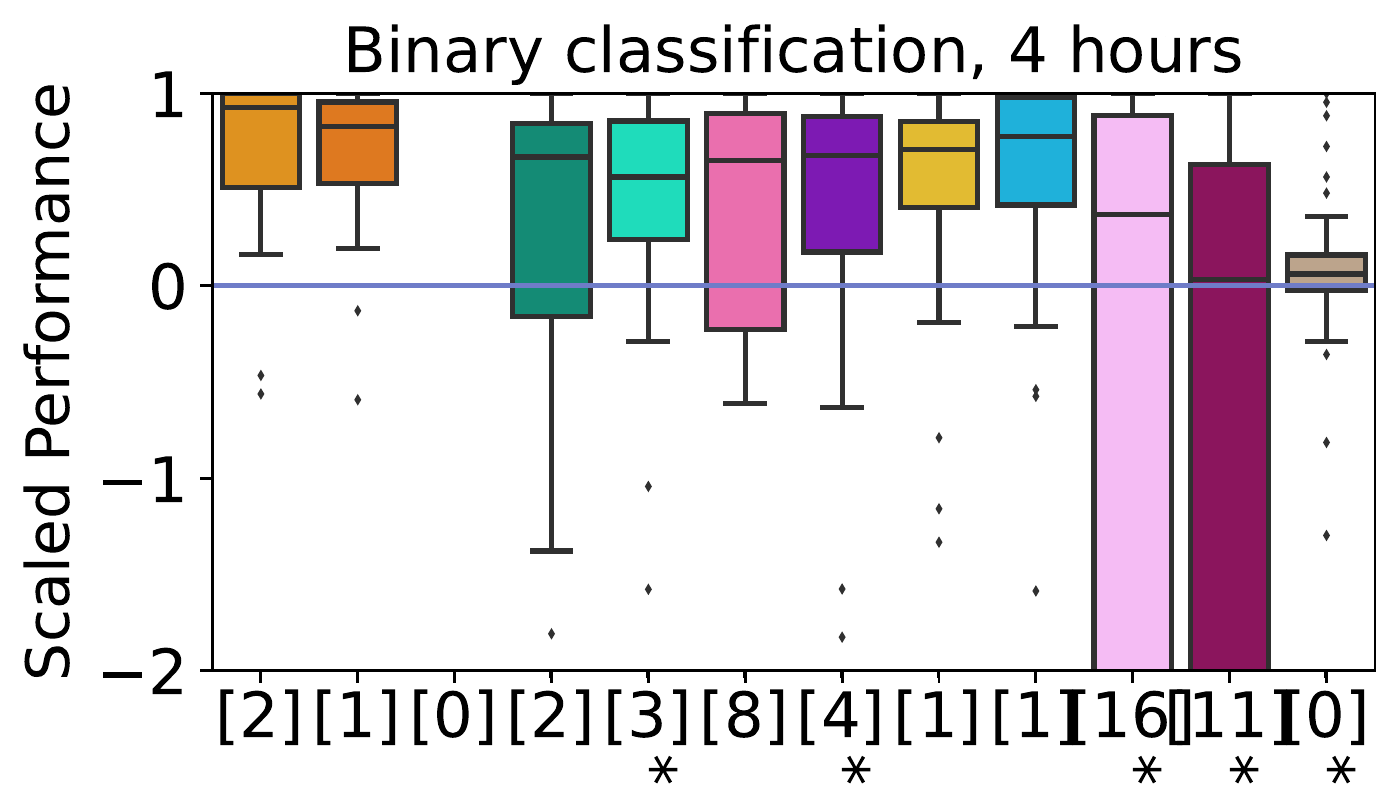}
     \end{subfigure}
     \hfill
     \begin{subfigure}[b]{0.48\textwidth}
         \centering
         \includegraphics[width=\textwidth]{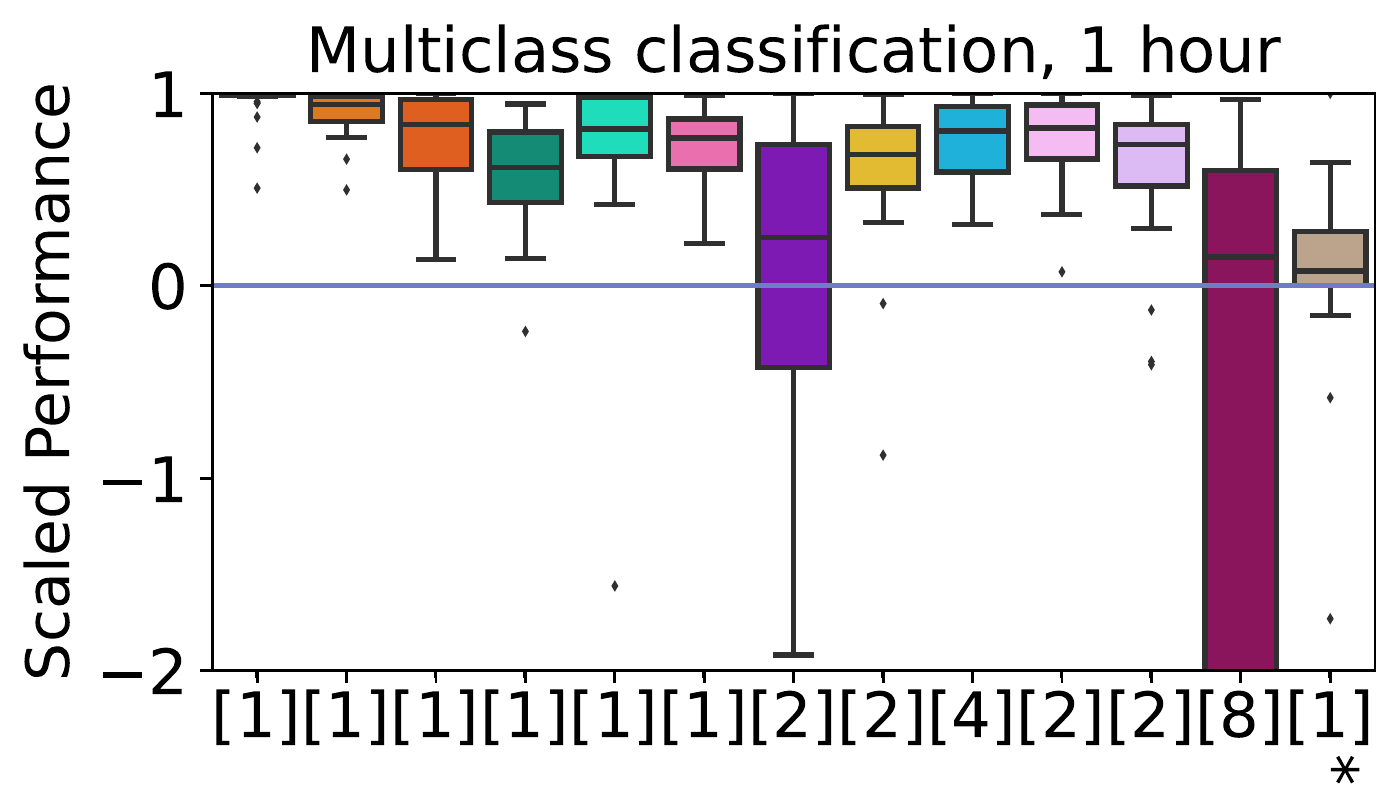}
     \end{subfigure}
     \hfill
     \begin{subfigure}[b]{0.48\textwidth}
         \centering
         \includegraphics[width=\textwidth]{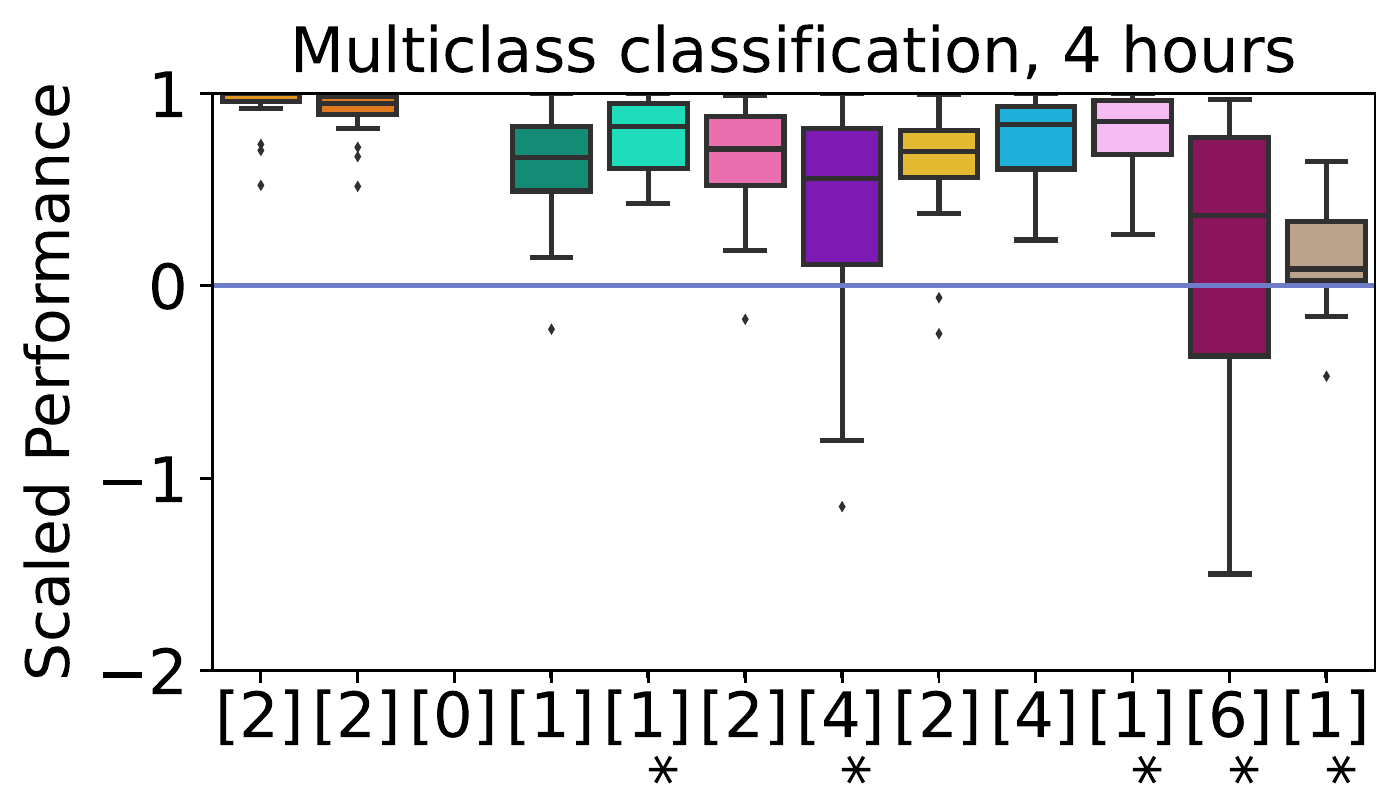}
     \end{subfigure}
     \hfill
          \begin{subfigure}[b]{0.48\textwidth}
         \centering
         \includegraphics[width=\textwidth]{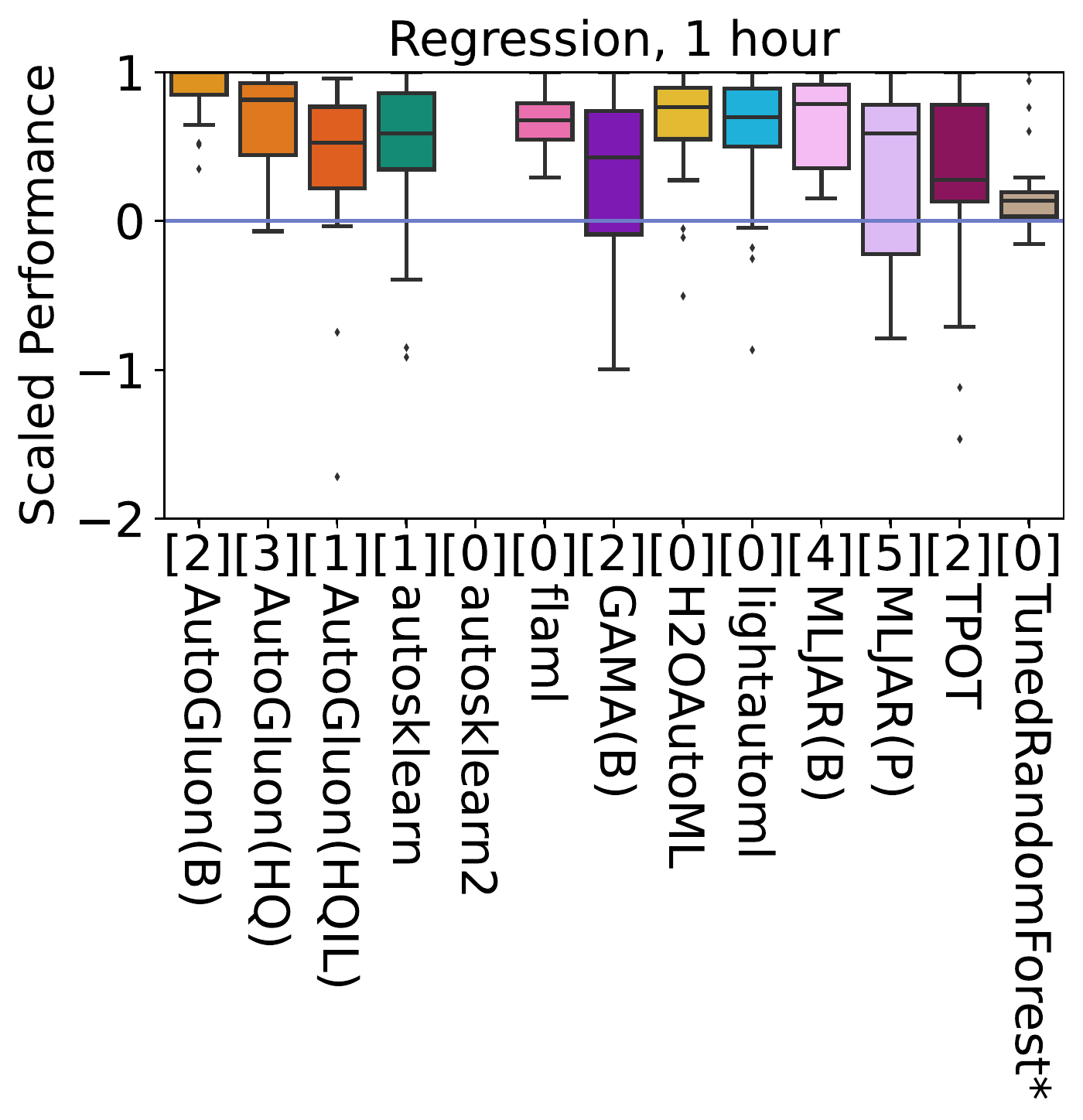}
         %\caption{Regression, 1 hour}
         %\label{fig:y equals x}
     \end{subfigure}
     \hfill
     \begin{subfigure}[b]{0.48\textwidth}
         \centering
         \includegraphics[width=\textwidth]{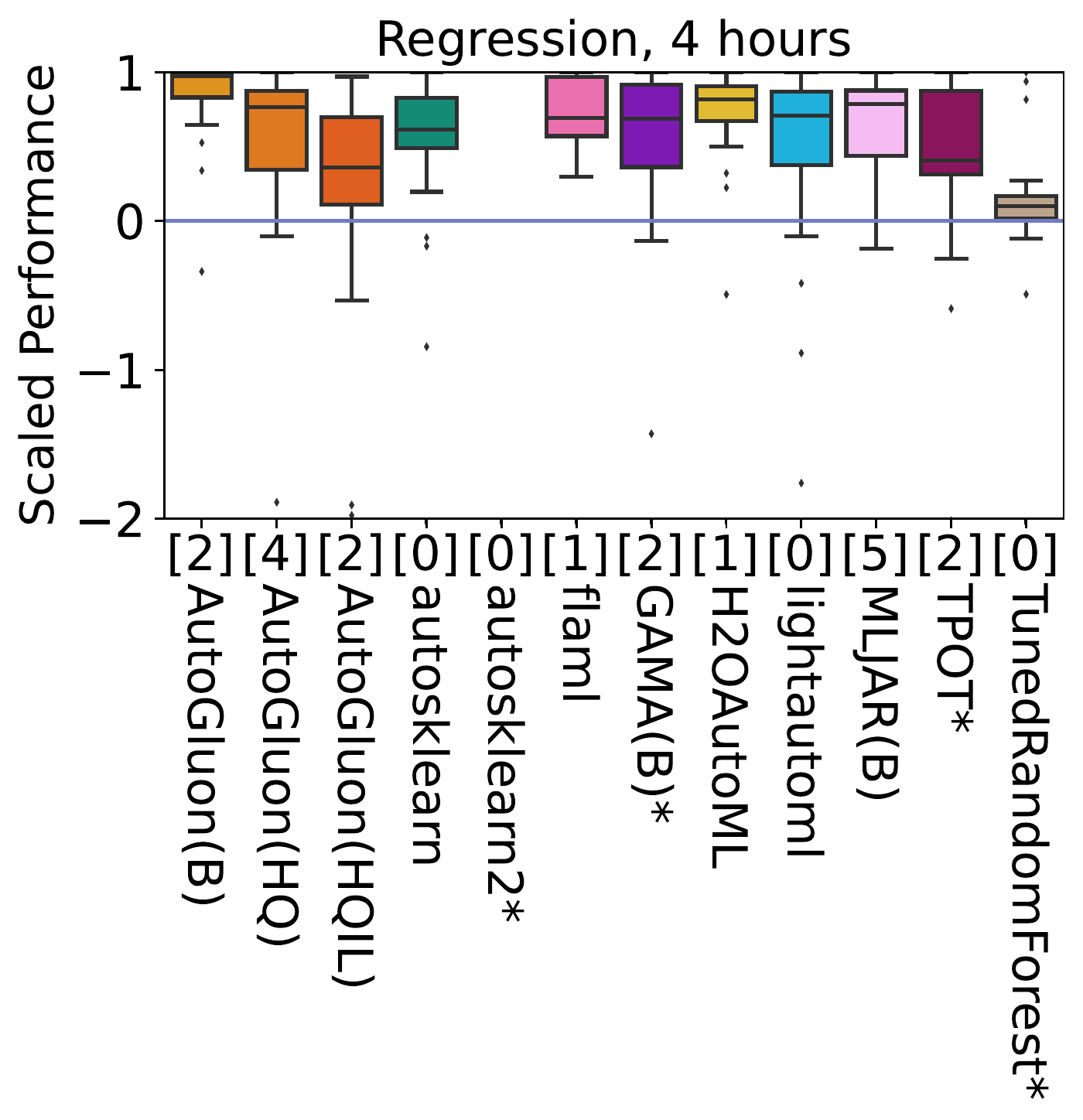}
     \end{subfigure}
     \caption{Boxplots of framework performance across tasks after scaling the performance values from random forest (0) to best observed (1). The number of outliers for each framework that are not shown in the plot are denoted on the x-axis. An asterisk (*) next to a framework name means the results were obtained in 2021.}
     \label{fig:scaled-boxplots}
\end{figure}

To complement the CD diagrams---which obfuscate the relative performance differences---we show box plots of obtained results (after imputation) across all tasks in Figure~\ref{fig:scaled-boxplots}.
Because the performances are not commensurable across tasks, we first scale all results per task between the random forest performance (0) and the best observed performance (1) (that is, higher scores correspond to better performance).
This also makes the scaled value interpretable. Furthermore, the value scales based on the improvement over the baseline that is observed to be achievable, and negative values are worse than the random forest baseline.
While the boxplots are calculated over performance data on all tasks, the plots are cut off to allow a better visualization of the most relevant area.
The number of outliers for each framework that are not shown in the plot are denoted on the x-axis.

Even if ranks are similar, the performance distribution might be noticeably different.
For example, \autogluon{}\systemcase{(HQIL)} and \mljar{}\systemcase{(P)} achieve very similar average ranks on the one hour regression tasks. However, we observe from the boxplots that while \autogluon{}\systemcase{(HQIL)} achieves lower median normalized performance in this segment, it outperforms random forest with much greater consistency than \mljar{}\systemcase{(P)}. 
Similarly, while \tpot{}'s average rank is generally close to that of the Tuned Random Forest baseline, \tpot{} exhibits much higher variance in its prediction quality.

\begin{figure}[h]
    \centering
    \includegraphics[width=\textwidth]{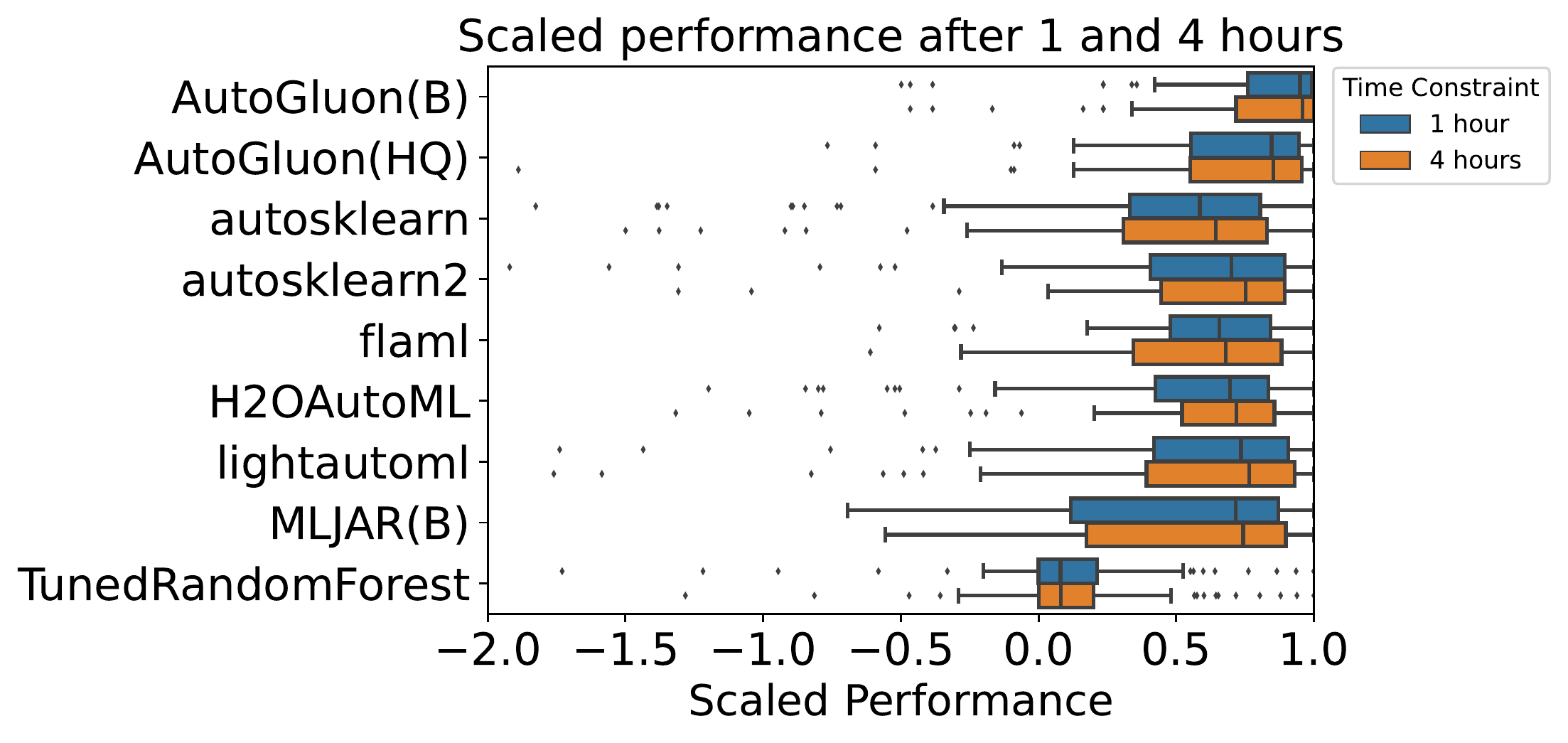}
    \caption{Scaled performance for each framework under different time constraints. Only frameworks which have evaluations on all tasks for both time constraints are shown. Performance generally does not improve much with more time.}
    \label{fig:performance-by-time}
\end{figure}

Finally, we take a closer look at the difference in performance for different time constraints in Figure~\ref{fig:performance-by-time}. Performance is scaled between 1 hour random forest performance (0) and best observed performance (1). We only include frameworks for which we evaluated all tasks on both constraints. We see that the performance is very similar overall, though generally results improve slightly with a larger time budget. The performance of \autogluon{} confirms that better performance is still possible for other frameworks, but we suspect that those frameworks are limited by their search space. For example, if a stacking ensemble is required to reach the best performance, then the other methods will not achieve the best performance regardless of time constraint.

\begin{figure}[b!]
    \centering
    \includegraphics[width=\textwidth]{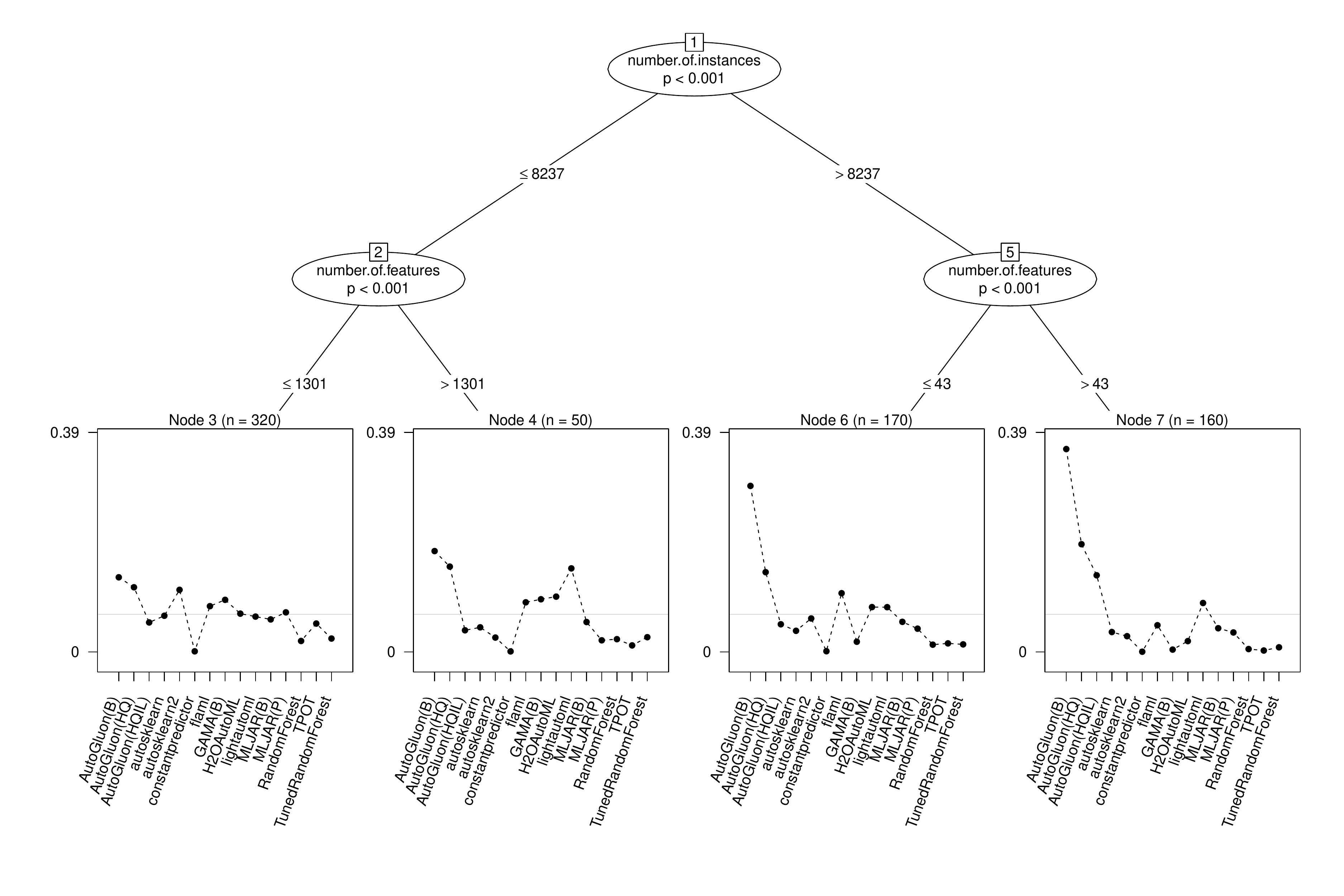}
    \caption{Bradley-Terry tree of depth three for classification tasks. Results from the one hour classification benchmark were used, and missing values were imputed by constant predictor performance. One observation within the BT tree equals the preference ranking of one fold on one data set. The $n$ value denotes the number of observations in the leaf node. }
    \label{fig:main-bt-tree-plot}
\end{figure}

\subsection{Bradley-Terry Trees}\label{sec:bt}
Bradley-Terry (BT) trees \citep{Strobl2011} can be used to statistically analyse benchmark experiments based on data set characteristics \citep{eugster2014}.
These trees use data set characteristics---such as the number of instances, the number of features, the ratio of missing values, and others---to split paired performance comparisons of the framework to find statistically significant differences in performance.
Bradley-Terry models originate in psychology to analyze paired comparison experiments of subjects preferring one stimulus over another.
For our benchmark, such a preference ranking can be easily derived by pairwise performance comparisons of all frameworks with regard to the data sets and cross-validation folds.

The underlying algorithm of model-based recursive partitioning of Bradley–Terry models works as follows:
In each split of the BT tree, a BT model is fitted for the paired comparisons based on the underlying data set characteristics.
Following \citet{zeileis2007} and \citet{eugster2014}, the BT model performs a statistical test of parameter instability for the chosen data characteristics.
If this test reveals a significant instability in the model parameters, the corresponding tree node splits the data according the the characteristic yielding the highest instability (lowest test p-value).
The splitting cut-point is then determined such that it has the highest improvement of the model fit.
This procedure is repeated until either no significant instability is left, a set tree depth is reached, or further splits would exceed a set minimum number of observations in the leaves.

Numeric values in the tree leaves are \emph{worth parameters} that can be interpreted as preferences for the different frameworks~\citep{eugster2014}.
Since these values are in $[0, 1]$ and sum up to $1$ within a leaf, they can be understood as the probability of a framework performing best, given the data characteristics in the corresponding leaf.  

Figure~\ref{fig:main-bt-tree-plot} shows a Bradley-Terry tree for classification tasks for a runtime of one hour.
For simplicity, in order to obtain an easily understood tree, only the number of instances (rows) and the number of features were chosen as data characteristics.
The first split distinguishes between data sets with more than $8237$ instances and those equal to or below that cut-point. The second split then further split into leaves based on the number of features in the datasets. In the nodes, from left to right, we see that:
\begin{itemize}
 \item For small datasets with a relatively low dimensionality (Node 3, 32 tasks) the performance differences among the best frameworks are not very pronounced.
Even though \autogluon{}\systemcase{(B)} is preferred for those kinds of data sets, \autogluon{}\systemcase{(HQ)} and \autosklearnii{} show similar performances.
\item For very wide datasets, those with a similarly small number of instances but much higher dimensionality, more than 1301 features (Node 4, 5 tasks), performance differences increase overall with \autogluon{}\systemcase{(B)} still being preferred over all others, but now followed by \autogluon{}\systemcase{(HQ)} and \lama{} with similar performances, and larger distance to \water{}, \gama{}\systemcase{(B)} and \flaml{}.
\item For datasets with a larger number of instances, more than 8237 (Node 6 and 7), we see that \autogluon{}\systemcase{(B)} is preferred by a large margin over \autogluon{}\systemcase{(HQ)} as second. There are minor differences depending on the number of features, such as \flaml{} being a preferred third option when there are few features, at most 43, but \autogluon{}\systemcase{(HQIL)} being preferred if the dataset has a relatively higher dimensionality.
\end{itemize}

More Bradley-Terry trees for the subsets of binary and multiclass classification as well as regression can be found in the appendix.
The findings from the BT trees are essentially the same as those presented here, as \autogluon{}\systemcase{(B)} is overall the preferred framework in most tree leaves. One exception is for small balanced binary classification tasks, shown in Figure~\ref{fig:bt-auc}, where various frameworks are preferred over \autogluon{}.

In conclusion, it can be observed that the performance gap of \autogluon{}\systemcase{(B)} increases particularly as data sets become more complex. 
Moreover, the reader is strongly invited to explore the aforementioned interactive visualization tool, with which deeper BT trees based on several more data set characteristics can be constructed on various task types.

\subsection{Model Accuracy vs. Inference Time Trade-offs}
\label{sec:inference}

Model accuracy (here measured by AUC, log loss, or RMSE) plays a central role in evaluating performance of machine learning models.  However, maximizing accuracy can come at the cost of added model complexity.  One practical way to consider the complexity of the model is to measure the inference speed of the resulting model.

Some of the integrated frameworks offer a ``compete" mode (\autogluon{}, \mljar{}, \lama{} and \gama{}) that maximizes accuracy, typically at the cost of increased model complexity, similar to competing in a Kaggle competition. %\footref{foot:kaggle}
This can lead to models being built that are highly accurate but are extremely slow at inference time and are therefore not practical in many real-life use-cases.  

However, some frameworks provide multiple presets that allow the user to balance the trade-off between accuracy and inference time differently. Unless the use of a specific preset is mentioned (for \autogluon{} and \mljar{}), results in this section used presets which prioritize accuracy over inference time, or in the case of \water{}, a balance of the two. Thus, performing additional experiments with other presets is advised when inference time is important. We also recognize that there are applications for which inference time is not important.

In order to evaluate the limitations of the models produced by each framework we also measured their inference time.
The inference time reported on in this section is measured by having the frameworks infer on samples of test data, either on a single row which is in memory, or on ten thousand rows loaded from disk. Each measurement is done ten times and the median is reported. Measurements for \tpot{} are omitted from this section, as \tpot{} requires data which is already encoded and the encoding step may consume a considerable part of the total inference time, making direct comparisons invalid.
This metric provides important insight into the trade-offs that tool authors make in their algorithm designs. 

\begin{figure}[!htb]
     \centering
     \begin{subfigure}[b]{0.45\textwidth}
         \centering
         \includegraphics[width=\textwidth]{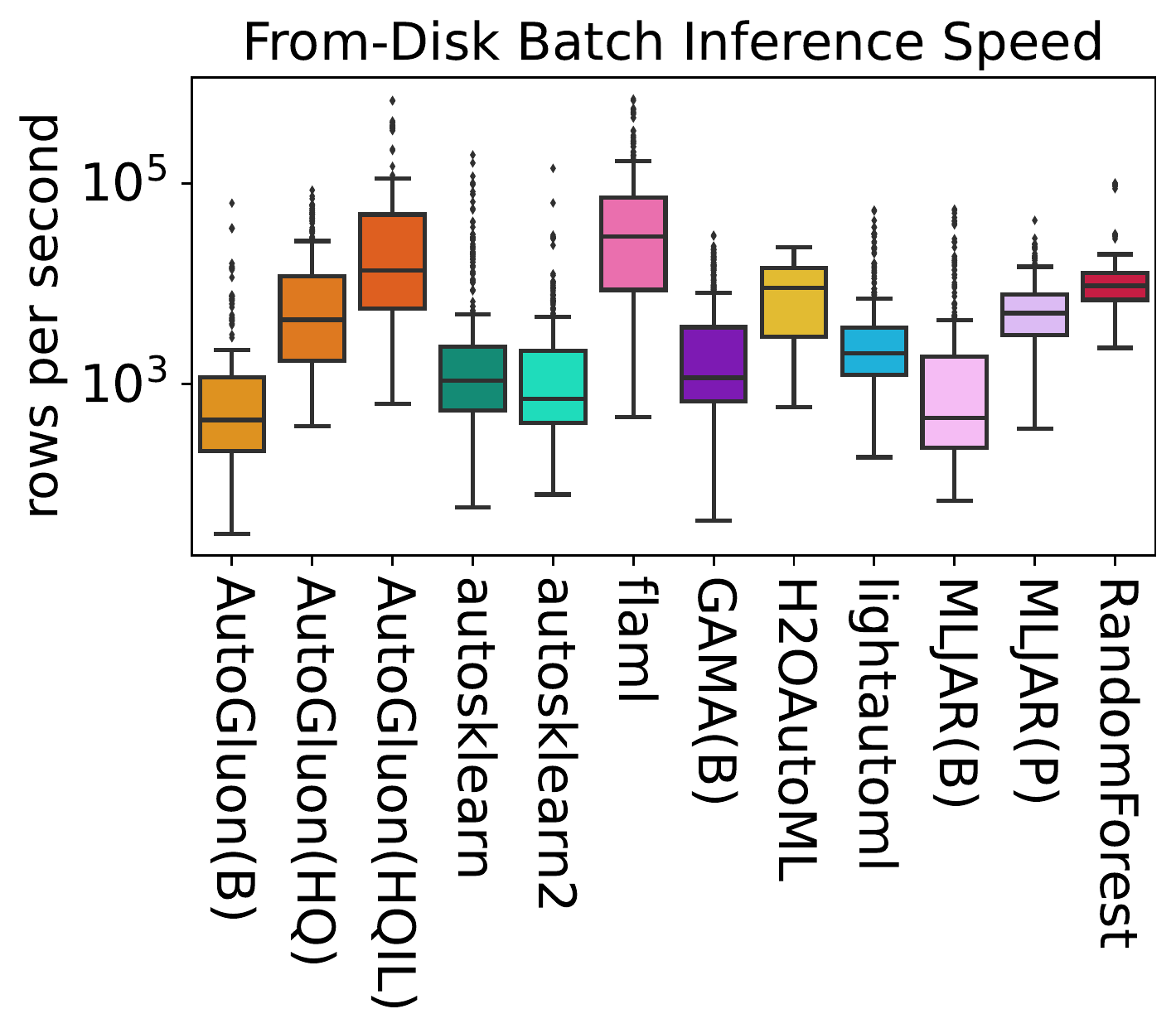}
     \end{subfigure}
     \hfill
     \begin{subfigure}[b]{0.45\textwidth}
         \centering
         \includegraphics[width=\textwidth]{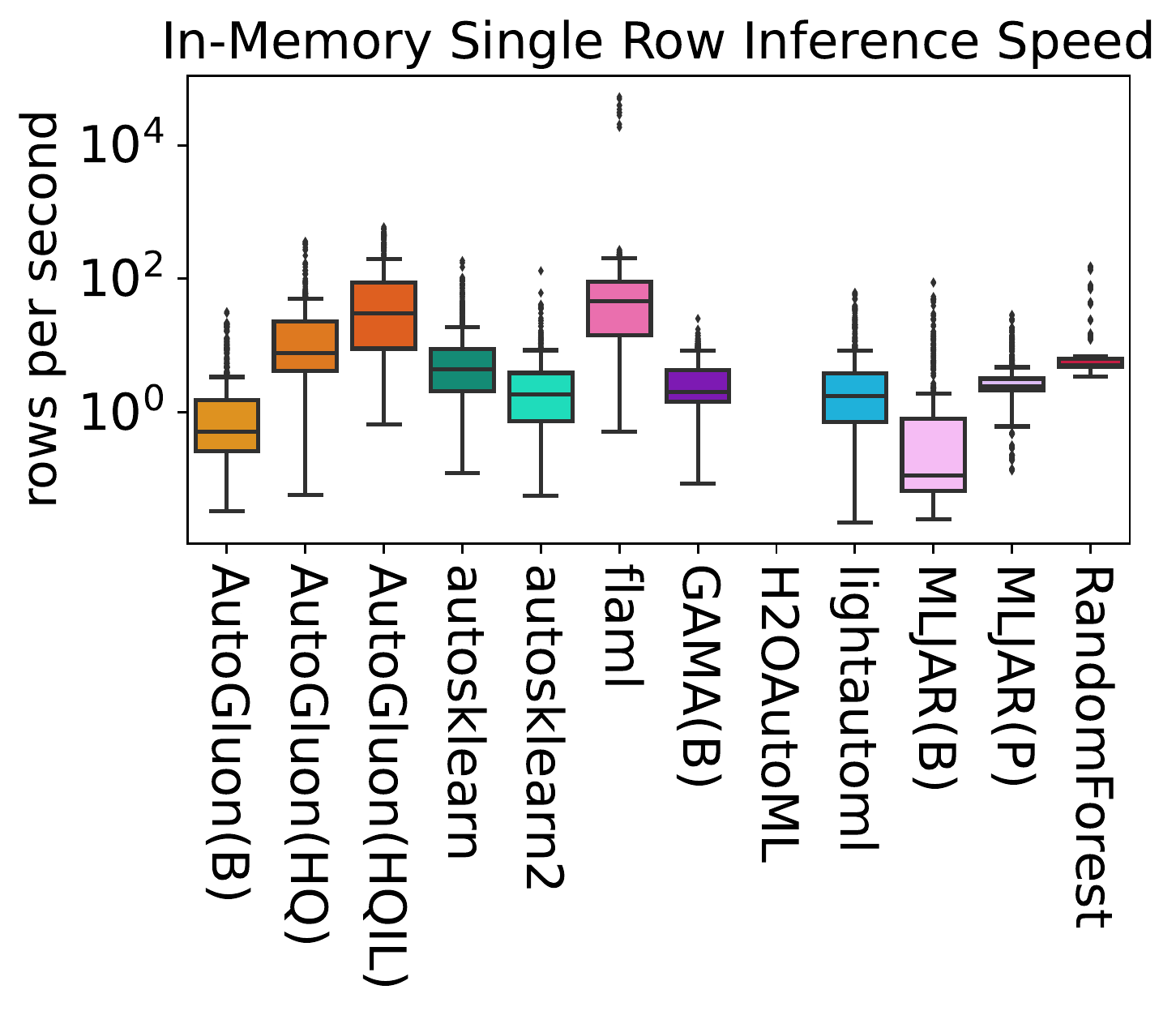}
     \end{subfigure}
     \caption{Inference speed on unseen data in rows per second, normalized from inference on 10 thousand rows from solid-state drive (left) or single rows in memory (right). Measured for models created with the one hour time constraint and all scenarios.}
     \label{fig:prediction-time}
\end{figure}

Figure~\ref{fig:prediction-time} shows aggregated inference times across all models standardized to rows per second (fewer is slower).
For \water{} we do not report in-memory inference speed as technical limitations of our benchmarking tool result in pessimistic inference time measurements for non-Python frameworks.\footnote{
Transferring data between our Python-based benchmark tool and \water{}'s Java back-end requires serialization to disk. Serialization to disk occurs significant overhead, especially when comparing to in-memory inference. Faster inference speeds will be obtained by using H2O Frames directly in \water{}, or using H2O in it's production mode, which performs inference on exported models directly from disk instead of in memory.
}
Here we see that the high accuracy of \autogluon{}\systemcase{(B)} comes at the cost of extremely slow inference times. This is explained by the large models produced by combining both stacking and ensembling to form ensembles of multiple layers. In general, we see that ensemble models are slower (\autosklearn{}, \gama{}, and \mljar{}\systemcase{(B)} also do ensembling), and \gama{} and \autosklearn{} with the same search space and ensemble methods have comparable inference speeds. \water{} and \autogluon{}(HQIL) maintain fast inference speed despite making use of ensembles, by building on more optimized models or explicitly selecting for fast inference when building the ensemble.
\flaml{} stand out as having very fast inference time, a potential explanation is that \flaml{}'s cost-frugal optimization also indirectly takes inference time into account, as it is part of the time estimate which is used to consider which model to tune.

\begin{figure}[!htb]
     \centering
     \begin{subfigure}[b]{0.45\textwidth}
         \centering
         \includegraphics[width=\textwidth]{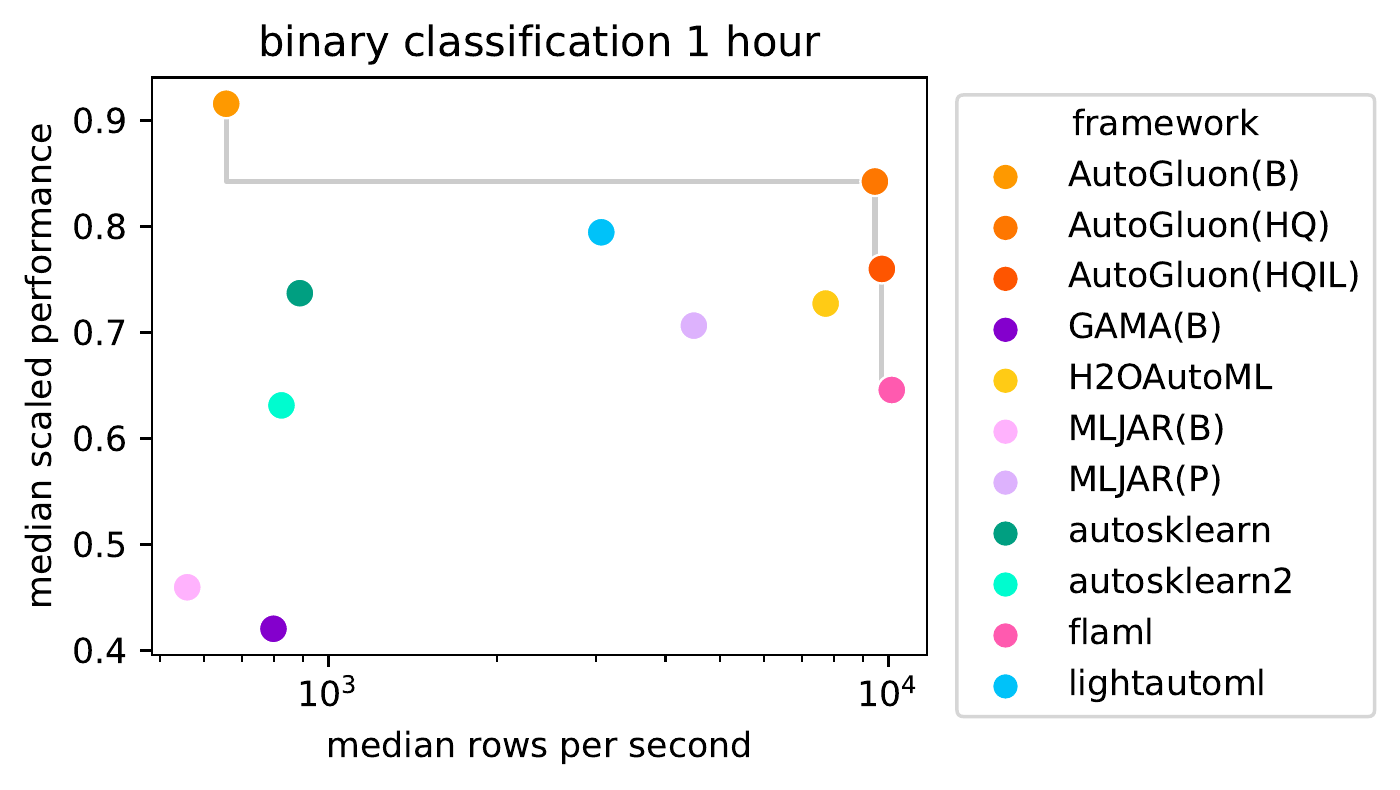}
     \end{subfigure}
     \hfill
     \begin{subfigure}[b]{0.45\textwidth}
         \centering
         \includegraphics[width=\textwidth]{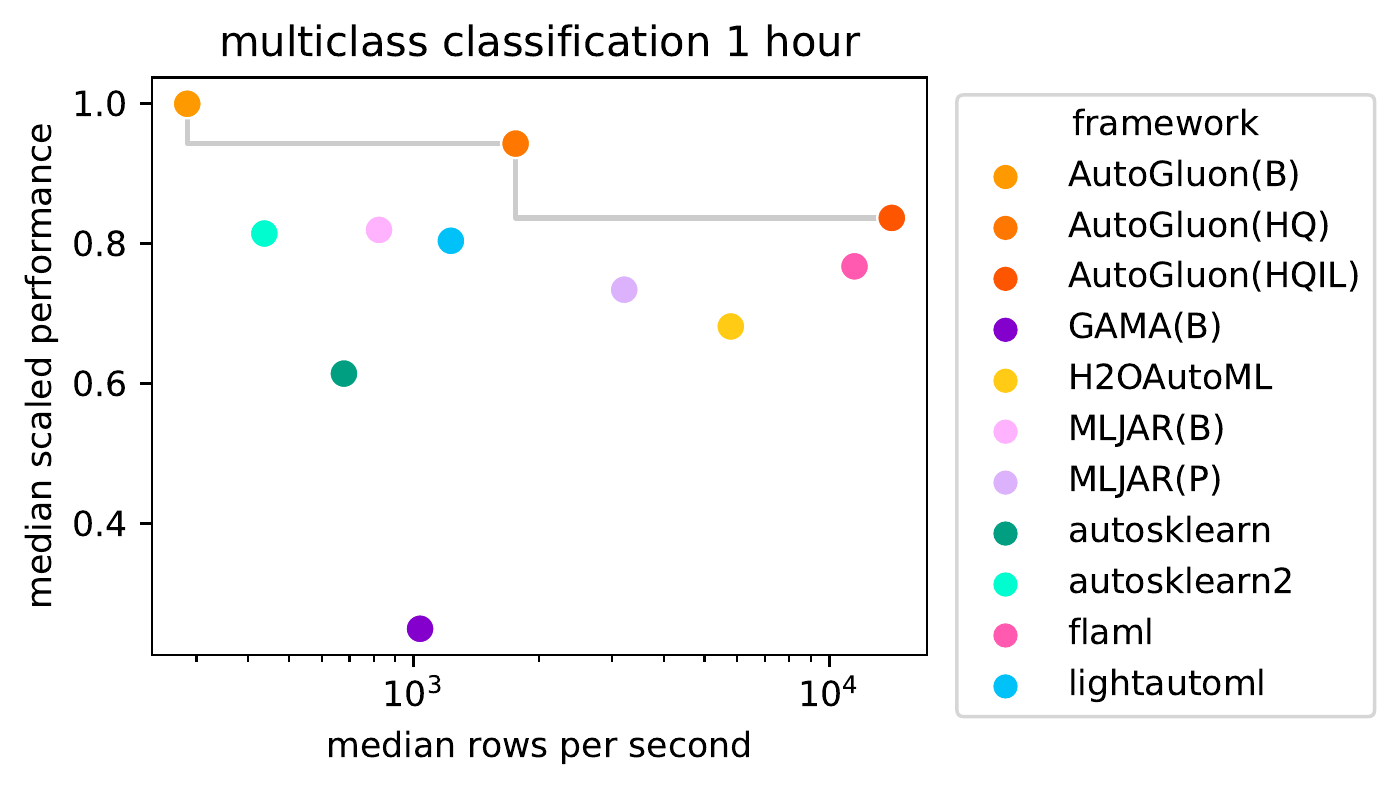}
     \end{subfigure}
     \hfill
     \begin{subfigure}[b]{0.45\textwidth}
         \centering
         \includegraphics[width=\textwidth]{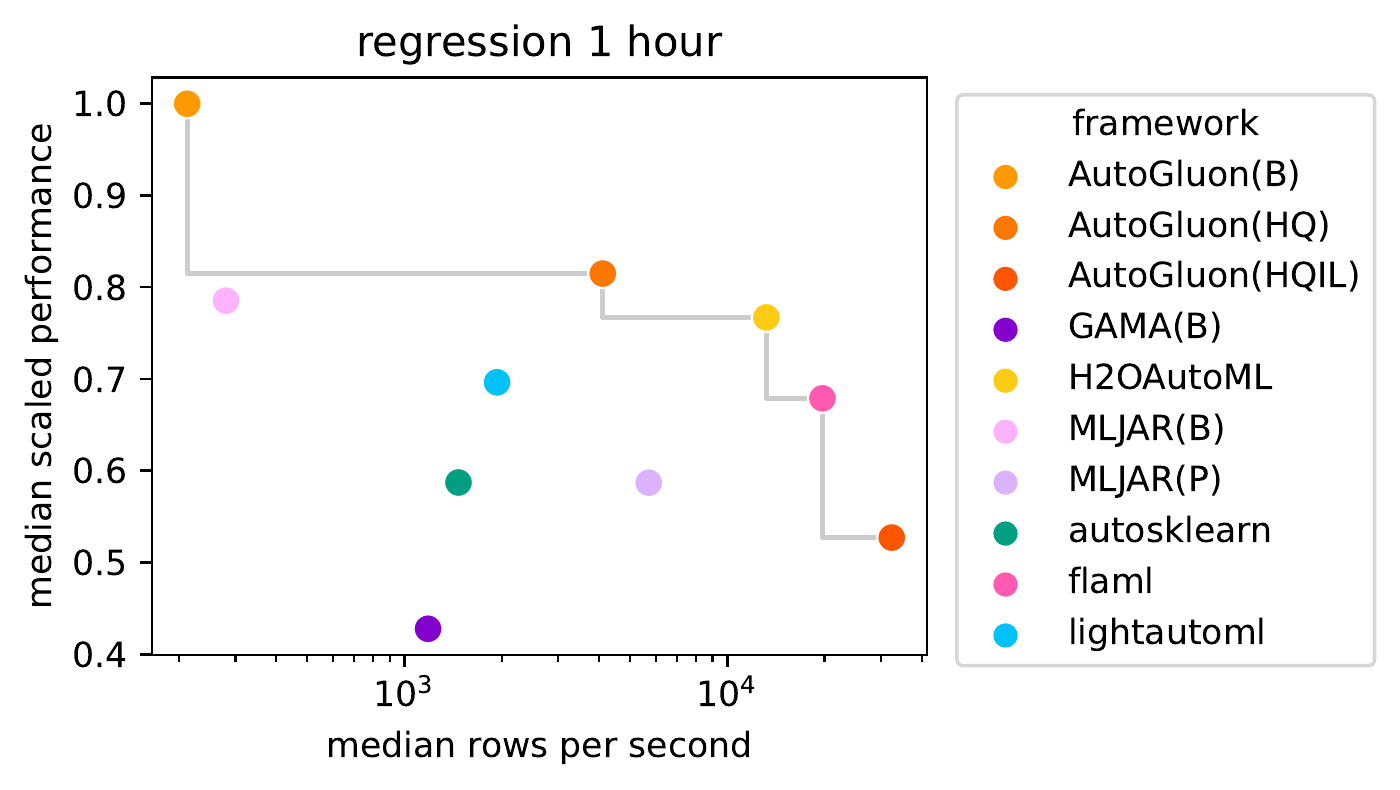}
     \end{subfigure}
     \caption{Pareto Frontiers of framework performance across tasks after scaling the performance values from the random forest (0) to best observed (1) for each task type on a one hour time budget.}
     \label{fig:pareto}
\end{figure}

In Figure~\ref{fig:pareto}, we show the Pareto front for all three scenarios with a one-hour time constraint, demonstrating the average normalized model accuracy against corresponding median per-row prediction speeds. Here, it is more apparent that the frameworks that achieve the highest accuracy do so at the cost of inference time performance. This demonstrates that when contextualizing any type of model accuracy results, it is important to consider any trade-offs that may have been made to achieve the extra performance and how that will affect the framework's usability in practice. In this case, measuring accuracy in isolation does not give the complete picture of the overall utility of a particular framework.

It should also be noted that with sufficient computing infrastructure and effort, scoring across rows or chunks of data could be parallelized in a production system, which would reduce the overall prediction time as compared to predicting a single test set on a single machine. However, our goal in this section was to compare the inference time of the high-accuracy models derived from different AutoML frameworks to each each other rather than evaluate different techniques for speeding up the inference of any individual system. 

\subsection{Observed AutoML Failures}\label{sec:errors}
While most jobs completed successfully, we observed multiple framework errors during our experiments.
In this section, we will discuss where AutoML frameworks fail, although we want to stress that development for these packages is ongoing.
For that reason, it is likely that the same frameworks will not experience the same failures in the future (especially after gaining access to all experiment logs).
We categorize the errors into the following categories:

\begin{itemize}
    \setlength\itemsep{0.1em}
    \item[] \textbf{Memory}: The framework crashed due to exceeding available memory or encountering other memory-related errors, such as segmentation faults.
    \item[] \textbf{Time}: The framework exceeded the time limit past the leniency period.
    \item[] \textbf{Data}: Errors due to specific data characteristics (such as imbalanced data) occurred.
    \item[] \textbf{Implementation}: Any errors caused by bugs in the AutoML framework code occurred.
\end{itemize}

\begin{figure}[tb]
  \centering
  \begin{subfigure}[tb]{0.48\textwidth}
      \centering
         \includegraphics[width=\textwidth]{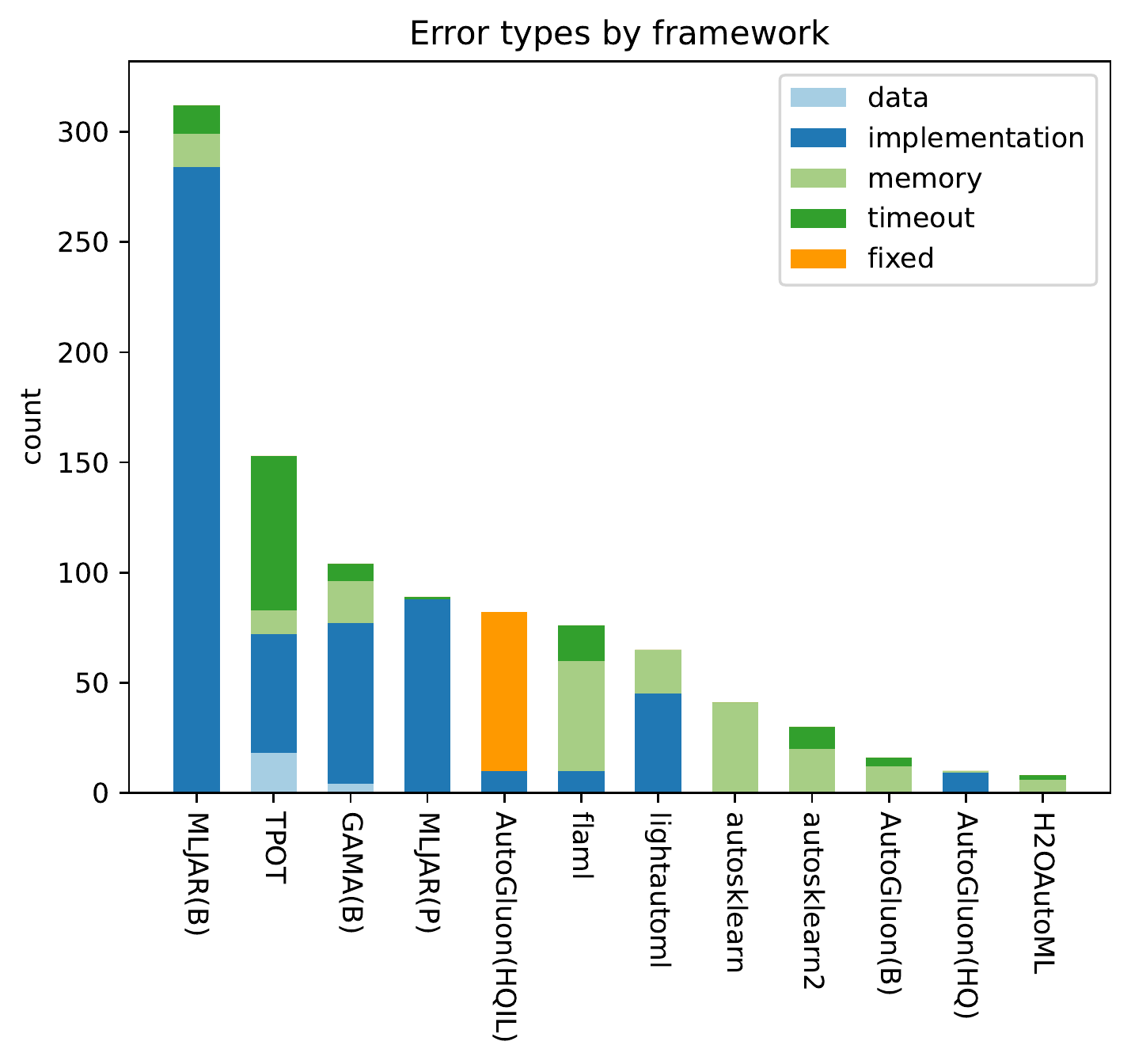}
         \caption{Errors by type for each framework.}
         \label{fig:errors-type}
  \end{subfigure}
  \hfill
  \begin{subfigure}[tb]{0.48\textwidth}
    \centering
    \includegraphics[width=\textwidth]{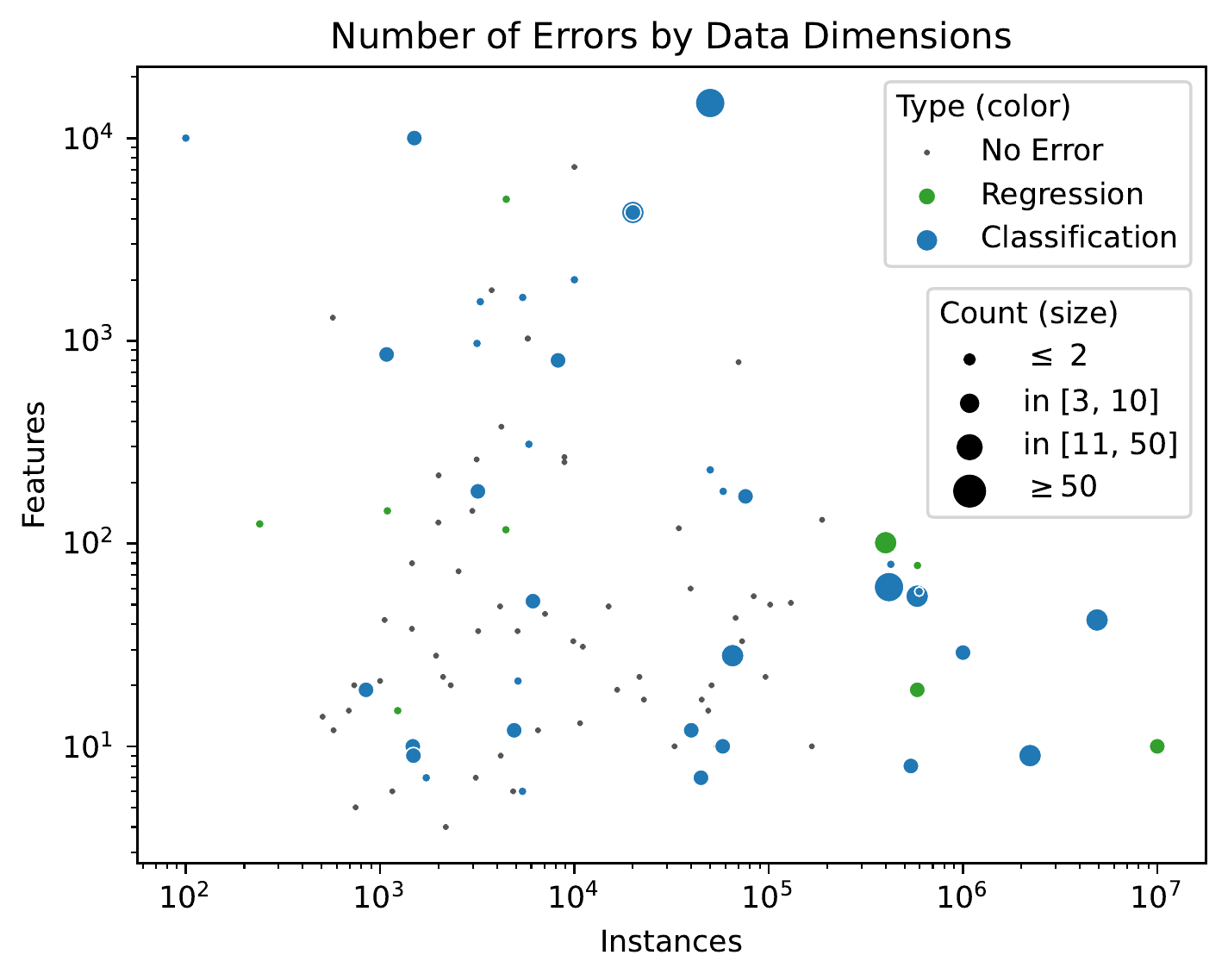}
    \caption{Number of errors by size of the dataset.}
    \label{fig:errors-size}
  \end{subfigure}
\end{figure}

\noindent These categories are a bit crude and ultimately subjective, since from a reductive viewpoint, all errors are implementation errors. However, they serve for a quick overview.
We also introduce a `fixed' category in Figure~\ref{fig:errors-type} to denote errors from a specific bug in \autogluon{}\systemcase{(HQIL)} which is already fixed in newer releases. Additional details on this, and other errors encountered, can be found in Appendix~\ref{sec:app-errors}.

Figure~\ref{fig:errors-type} shows the errors by type, and Figure~\ref{fig:errors-size} shows errors by dataset size and dimensionality on the right.
Overall, memory and time constraints are the main cause for errors, with one major exception.\footnote{
\mljar{}\systemcase{(B)} has 284 `implementation errors' which are almost exclusively caused by only 3 distinct errors.
}
We observe that errors are far more common in the classification benchmark suite than the regression suite.
This is largely accounted for by the difference in benchmarking suite size (33 and 71 tasks) and the fact that the largest data sets are mostly classification tasks, both in number of instances and features.
Unique to classification, we do observe several frameworks failing to produce models or predictions on highly imbalanced data sets.
This is also the case for the failures on the two small classification data sets (`yeast' and `wine-quality-white'), where careless use of internal validation splits yields splits that no longer contain all classes.
Interestingly, the distribution of the type of errors observed is different under different time constraints (as shown in Figure~\ref{app:fig:errors-by-type} of Appendix~\ref{sec:app-errors}).
Both memory and time constraint violations happen more frequently, which may potentially be explained by frameworks saving increasingly more models or building increasingly larger pipelines.  It is worth noting the stability of the different systems is varied, with some systems being far more stable than others.  Some types of errors, such as memory errors, could potentially be easily resolved by running the AutoML system on a machine with more RAM.  However, the implementation errors are more problematic because they represent an error from the code of the AutoML system.  Those are not currently resolvable without changes to the AutoML system itself, which may or may not be fixed by code authors in the future.

\begin{figure}[tb]
     \centering
     \begin{subfigure}[b]{0.45\textwidth}
         \centering
         \includegraphics[width=\textwidth]{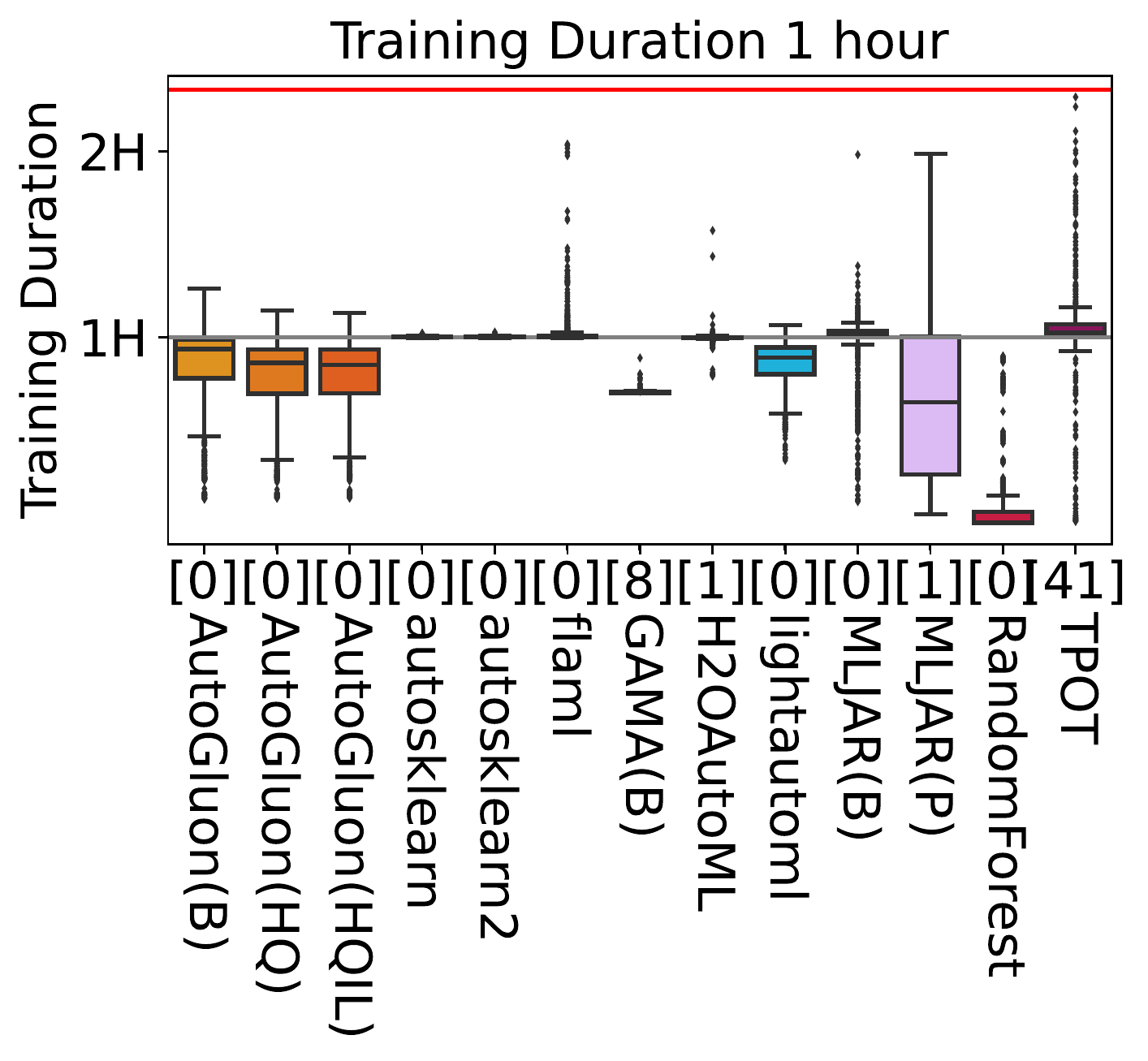}
     \end{subfigure}
     \hfill
     \begin{subfigure}[b]{0.45\textwidth}
         \centering
         \includegraphics[width=\textwidth]{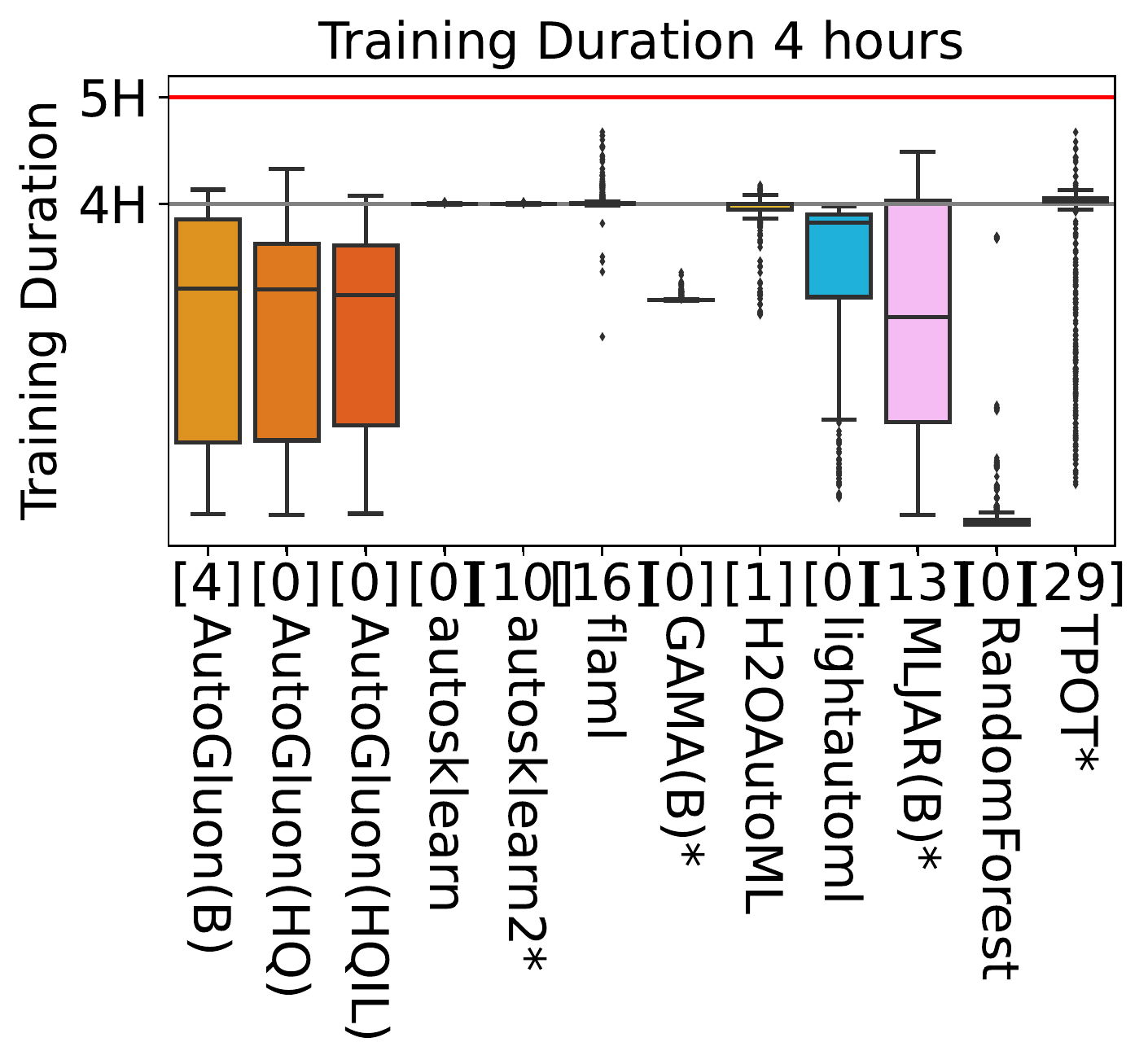}
     \end{subfigure}
     \caption{Time spent during search with a one hour budget (left) and four hour budget (right). The grey line indicates the specified time limit, and the red line denotes the end of the leniency period. The number of timeout errors for each framework are shown beside it.}
     \label{fig:search-time}
\end{figure}

Only when the framework exceeds the time budget by more than one hour do we record a time error. 
However, as we can see in Figure~\ref{fig:search-time}, not all AutoML frameworks adhere to the runtime constraints equally well, even if they finish within the leniency period.
In the figure, the training duration for each individual job (task and fold combination) are aggregated, and timeout errors are shown above each framework, where missing values due to non-time errors are excluded.
These plots reveal different design decisions around the specified runtime, with some frameworks never exceeding the limit by more than a few minutes, while others violate it by a larger margin with some regularity.
Interestingly, we see that a number of frameworks consistently tend to stop far before the specified runtime limit.

\section{Conclusion}\label{sec:conclusion}
We presented a novel benchmark for measuring and comparing AutoML frameworks.
To ensure reproducibility, fair comparison, and detailed analysis of the results, our open-source benchmarking tool automates the empirical evaluation of any integrated framework on any supported task, including installation of the AutoML framework, provisioning the data for training and inference, resource allocation, and processing of the results.
This greatly simplifies evaluating AutoML frameworks while enhancing reproducibility and reducing errors.
We worked jointly with the authors of \numframeworks{} AutoML frameworks to evaluate their systems in a large-scale study on \numclassification{} classification and \numregression{} regression tasks.

When analyzing the predictive performance of these AutoML systems, we find that the \textit{average} ranks of the AutoML frameworks are generally very competitive with each other. Still, by using Bradley-Terry trees~\citep{Strobl2011}, %as proposed in \cite{eugster2014}, 
we find that their relative performance is affected by data characteristics---such as the data set size, dimensionality, and class balance---highlighting specific strengths and weaknesses.
Overall, in terms of model performance, \autogluon{} consistently has the highest average rank in our benchmark. Additionally, in most scenarios, the AutoML frameworks outperform even our strongest baseline.

Because inference time is an important factor in real-world applications, we also reviewed the inference time and accuracy trade-off and found large differences in inference time of the produced models, at times spanning multiple orders of magnitude. The most accurate frameworks achieve higher model accuracy at a large cost to performance in terms of inference speed.
Broadly speaking, while the models with higher accuracy also have slower inference time, not all frameworks produce models that are Pareto optimal.
Furthermore, specific AutoML frameworks allow users to choose presets for customizing the accuracy-inference time trade-off.

Finally, we analyzed scenarios in which AutoML frameworks fail to produce a model and found that the main cause for failure was data set size. In other words, not all methods scale well.
To allow further analysis of our results, we provide an open-source interactive visualization tool, which includes graphical representations and statistical tests.

\subsection{Limitations}
These quantitative results are obtained by using AutoML frameworks with preset configurations only. The performance of frameworks under non-default settings can be very relevant when, for example, there is budget available to also optimize AutoML hyper-hyperparameters, or when a custom search space is used. Additionally, many frameworks are under active development, so the results presented here may not be representative of their performance in the future.

Next, all frameworks differ along multiple design axes, which prohibits attributing performance differences to any specific component of the AutoML framework (such as the search algorithm), without additional analysis.

Moreover, since we provide a purely quantitative comparison, it ignores qualitative aspects of AutoML frameworks that are very relevant in real-world settings, such as the produced model's interpretability or the level of support.

Lastly, the benchmarks were executed on 8-CPU machines, which is very modest by today's standards. Therefore, frameworks with better parallelism may show even greater advantages on higher powered machines (with, for example, 50 or 100 CPUs). On the other hand, even shorter time constraints may be useful for a human-in-the-loop or green AutoML setting.

\subsection{Future Work}
The benchmarking suites, or sets of tasks, proposed in this paper are meant as a starting point to be improved upon in collaboration with the AutoML community.
We will seek to update these suites based on community discussion so that they continue to reflect modern challenges while also decreasing the risk of AutoML frameworks overfitting to the benchmark.
In particular, we are interested in extending the benchmarking suites with problems that feature free-form text data.
Real-world tabular data often contain instances of text data with different semantic meaning, such as addresses or URLs, from which meaningful features can be extracted.
This kind of feature engineering is typically very important for model performance on such data sets, but the current selection of data sets does not reflect this, in part because not all AutoML frameworks support text features.

We would also like to extend the benchmarking tool to support new problems, such as multi-objective optimization tasks.
While model accuracy is often an important metric, `secondary' metrics---such as a model's inference time or fairness---are often crucial for real-world applications.
Multi-objective optimization can be used to convey the importance of these metrics to AutoML frameworks, for example, by using a fairness metric as secondary objective,~\citep{schmucker2021multiobjective}), which can subsequently provide Pareto fronts of models that optimize this trade-off.  
In particular for fairness related tasks, additional support to convey sensitive attributes and protected groups to the AutoML frameworks must be added.
Other interesting problem types include non-i.i.d. data, such as when temporal relationships are present in the data, or semi-supervised learning, where not all instances have an associated ground truth.

Finally, the evaluations in this paper were performed on the equivalent of commodity-level hardware using only CPU and a limited time budget.
In some cases, it is more desirable to devote a large budget to building a single model, perhaps even days of compute, potentially with GPU access.
In other cases, much smaller budgets may be desired, for example to reduce carbon footprints or improve the experience for workflows with a human in the loop.
As different behavior (both in robustness and model performance) on one-hour and four-hour budgets are already observed, future work may reveal different behavior when evaluating AutoML frameworks at different scales.

\subsection{Parting Words}
The benchmark tool presented in this work makes producing rigorous reproducible research both easier and faster.
We hope that the open and extensible nature of this benchmark motivates researchers to not only use the tool, but also to contribute their own data sets, framework integrations, or feedback and code contributions to the open source AutoML benchmark.
We strongly encourage this participation so that the benchmark may remain useful to the community for a long time to come. 

\acks{}
We would all like to give special thanks to everyone that contributed to the benchmark, both directly with pull requests and indirectly through opening issues. We also thank Rinchin Damdinov, Nick Erickson, Matthias Feurer, and Piotr P\l{}o\'{n}ski for feedback and corrections to this manuscript.

This work made use of the resources and expertise offered by the SURF Public Cloud Call which is financed by the Dutch Research Council (NWO). We also made use of research credits provided by the AWS Cloud Credit for Research program.

Pieter Gijsbers and Joaquin Vanschoren would like to acknowledge funding by AFRL and DARPA under contract FA8750-17-C-0141, and EU’s Horizon Europe research and innovation program under grant agreement No. 952215 (TAILOR).

Stefan Coors, Janek Thomas, and Bernd Bischl would like to acknowledge funding by the German Federal Ministry of Education and Research (BMBF) under Grant No. 01IS18036A. 

\newpage

\appendix

\section{OpenML Benchmark Suites}\label{app:suites}
Table~\ref{tab:269} and Table~\ref{tab:271} contain an overview of data sets used in the regression and classification benchmarking suites, respectively.
We hope to continuously update the benchmarking suites with new data sets that represent current challenges.

% Generated with https://github.com/openml/openml-python/issues/1126
\begin{longtable}{rlrr}
\caption{Tasks in the AutoML regression suite.}
\label{tab:269}
\\
\toprule
Task ID & name & n & p \\
\midrule
\midrule
\endfirsthead
\caption{Tasks in the AutoML regression suite (continued).}\\
\toprule
Task ID & name & n & p \\
\midrule
\midrule
\endhead
\addlinespace
359944 & abalone & 4177 & 9 \\
359929 & Airlines\_DepDelay\_10M & 10000000 & 10 \\
233212 & Allstate\_Claims\_Severity & 188318 & 131 \\
359937 & black\_friday & 166821 & 10 \\
359950 & boston & 506 & 14 \\
\addlinespace
359938 & Brazilian\_houses & 10692 & 13 \\
233213 & Buzzinsocialmedia\_Twitter & 583250 & 78 \\
359942 & colleges & 7063 & 45 \\
233211 & diamonds & 53940 & 10 \\
359936 & elevators & 16599 & 19 \\
\addlinespace
359952 & house\_16H & 22784 & 17 \\
359951 & house\_prices\_nominal & 1460 & 80 \\
359949 & house\_sales & 21613 & 22 \\
233215 & Mercedes\_Benz\_Greener\_Manufacturing & 4209 & 377 \\
360945 & MIP-2016-regression & 1090 & 145 \\
\addlinespace
167210 & Moneyball & 1232 & 15 \\
359943 & nyc-taxi-green-dec-2016 & 581835 & 19 \\
359941 & OnlineNewsPopularity & 39644 & 60 \\
359946 & pol & 15000 & 49 \\
360933 & QSAR-TID-10980 & 5766 & 1026 \\
\addlinespace
360932 & QSAR-TID-11 & 5742 & 1026 \\
359930 & quake & 2178 & 4 \\
233214 & Santander\_transaction\_value & 4459 & 4992 \\
359948 & SAT11-HAND-runtime-regression & 4440 & 117 \\
359931 & sensory & 576 & 12 \\
\addlinespace
359932 & socmob & 1156 & 6 \\
359933 & space\_ga & 3107 & 7 \\
359934 & tecator & 240 & 125 \\
359939 & topo\_2\_1 & 8885 & 267 \\
359945 & us\_crime & 1994 & 127 \\
\addlinespace
359935 & wine\_quality & 6497 & 12 \\
317614 & Yolanda & 400000 & 101 \\
359940 & yprop\_4\_1 & 8885 & 252 \\
\bottomrule
\end{longtable}

\begin{longtable}{rlrrrr}
\caption{Tasks in the AutoML classification suite.}
\label{tab:271}
\\
\toprule
Task ID & name & n & p & C & class ratio \\
\midrule
\midrule
\endfirsthead
\caption{Tasks in the AutoML classification suite (continued).}\\
\toprule
Task ID & name & n & p & C & class ratio \\
\midrule
\midrule
\endhead
\addlinespace
190411 & ada & 4147 & 49 & 2 & 0.33 \\
359983 & adult & 48842 & 15 & 2 & 0.31 \\
189354 & airlines & 539383 & 8 & 2 & 0.80 \\
189356 & albert & 425240 & 79 & 2 & 1.00 \\
10090 & amazon-commerce-reviews & 1500 & 10001 & 50 & 1.00 \\
\addlinespace
359979 & Amazon\_employee\_access & 32769 & 10 & 2 & 0.06 \\
168868 & APSFailure & 76000 & 171 & 2 & 0.02 \\
190412 & arcene & 100 & 10001 & 2 & 0.79 \\
146818 & Australian & 690 & 15 & 2 & 0.80 \\
359982 & bank-marketing & 45211 & 17 & 2 & 0.13 \\
\addlinespace
359967 & Bioresponse & 3751 & 1777 & 2 & 0.84 \\
359955 & blood-transfusion-service-center & 748 & 5 & 2 & 0.31 \\
359960 & car & 1728 & 7 & 4 & 0.05 \\
359973 & christine & 5418 & 1637 & 2 & 1.00 \\
359968 & churn & 5000 & 21 & 2 & 0.16 \\
\addlinespace
359992 & Click\_prediction\_small & 39948 & 12 & 2 & 0.20 \\
359959 & cmc & 1473 & 10 & 3 & 0.53 \\
359957 & cnae-9 & 1080 & 857 & 9 & 1.00 \\
359977 & connect-4 & 67557 & 43 & 3 & 0.15 \\
7593 & covertype & 581012 & 55 & 7 & 0.01 \\
\addlinespace
168757 & credit-g & 1000 & 21 & 2 & 0.43 \\
211986 & Diabetes130US & 101766 & 50 & 3 & 0.21 \\
168909 & dilbert & 10000 & 2001 & 5 & 0.93 \\
189355 & dionis & 416188 & 61 & 355 & 0.36 \\
359964 & dna & 3186 & 181 & 3 & 0.46 \\
\addlinespace
359954 & eucalyptus & 736 & 20 & 5 & 0.49 \\
168910 & fabert & 8237 & 801 & 7 & 0.26 \\
359976 & Fashion-MNIST & 70000 & 785 & 10 & 1.00 \\
359969 & first-order-theorem-proving & 6118 & 52 & 6 & 0.19 \\
359970 & GesturePhaseSegmentationProcessed & 9873 & 33 & 5 & 0.34 \\
\addlinespace
189922 & gina & 3153 & 971 & 2 & 0.97 \\
359988 & guillermo & 20000 & 4297 & 2 & 0.67 \\
359984 & helena & 65196 & 28 & 100 & 0.03 \\
360114 & Higgs & 1000000 & 29 & 2 & 0.89 \\
359966 & Internet-Advertisements & 3279 & 1559 & 2 & 0.16 \\
\addlinespace
211979 & jannis & 83733 & 55 & 4 & 0.04 \\
168911 & jasmine & 2984 & 145 & 2 & 1.00 \\
359981 & jungle\_chess\_2pcs\_raw\_endgame\_complete & 44819 & 7 & 3 & 0.19 \\
359962 & kc1 & 2109 & 22 & 2 & 0.18 \\
360975 & KDDCup09-Upselling & 50000 & 14892 & 2 & 0.08 \\
\addlinespace
3945 & KDDCup09\_appetency & 50000 & 231 & 2 & 0.02 \\
360112 & KDDCup99 & 4898431 & 42 & 23 & 0.00 \\
359991 & kick & 72983 & 33 & 2 & 0.14 \\
359965 & kr-vs-kp & 3196 & 37 & 2 & 0.91 \\
190392 & madeline & 3140 & 260 & 2 & 0.99 \\
\addlinespace
359961 & mfeat-factors & 2000 & 217 & 10 & 1.00 \\
359953 & micro-mass & 571 & 1301 & 20 & 0.18 \\
359990 & MiniBooNE & 130064 & 51 & 2 & 0.39 \\
359980 & nomao & 34465 & 119 & 2 & 0.40 \\
167120 & numerai28.6 & 96320 & 22 & 2 & 0.98 \\
\addlinespace
359993 & okcupid-stem & 50789 & 20 & 3 & 0.13 \\
190137 & ozone-level-8hr & 2534 & 73 & 2 & 0.07 \\
359958 & pc4 & 1458 & 38 & 2 & 0.14 \\
190410 & philippine & 5832 & 309 & 2 & 1.00 \\
359971 & PhishingWebsites & 11055 & 31 & 2 & 0.80 \\
\addlinespace
168350 & phoneme & 5404 & 6 & 2 & 0.42 \\
360113 & porto-seguro & 595212 & 58 & 2 & 0.04 \\
359956 & qsar-biodeg & 1055 & 42 & 2 & 0.51 \\
359989 & riccardo & 20000 & 4297 & 2 & 0.33 \\
359986 & robert & 10000 & 7201 & 10 & 0.92 \\
\addlinespace
359975 & Satellite & 5100 & 37 & 2 & 0.01 \\
359963 & segment & 2310 & 20 & 7 & 1.00 \\
359994 & sf-police-incidents & 2215023 & 9 & 2 & 0.14 \\
359987 & shuttle & 58000 & 10 & 7 & 0.00 \\
168784 & steel-plates-fault & 1941 & 28 & 7 & 0.08 \\
\addlinespace
359972 & sylvine & 5124 & 21 & 2 & 1.00 \\
190146 & vehicle & 846 & 19 & 4 & 0.91 \\
359985 & volkert & 58310 & 181 & 10 & 0.11 \\
146820 & wilt & 4839 & 6 & 2 & 0.06 \\
359974 & wine-quality-white & 4898 & 12 & 7 & 0.00 \\
\addlinespace
2073 & yeast & 1484 & 9 & 10 & 0.01 \\
\bottomrule
\end{longtable}

\clearpage
\section{Results}\label{app:results}
This appendix contains additional figures and tables with experimental results.
Tables~\ref{tab:auc-1h8c_gp3-A-H}-\ref{tab:rmse-4h8c_gp3-I-Z} report results per framework per task for different task types and time budgets.
Each value denotes the mean score of completed folds, the standard deviation in those completed folds, and the number of folds for which the AutoML framework did not return a result, if applicable.
A `-' denotes cases where the AutoML framework was unable to complete any fold of a task on a specific time budget.
A `\text{*}' next to a framework name denotes experiments that were conducted on an older version of the framework in 2021.

\footnotesize
\begin{landscape}
\begin{table}
\tiny
\begin{tabular}{rlrrrrrrrr}
\toprule
 & framework& AutoGluon(B)\ \ & AutoGluon(HQ)\ \ & AutoGluon(HQIL)\ \ & autosklearn\ \ & autosklearn2\ \ & flaml\ \ & GAMA(B)\ \ & H2OAutoML\ \ \\
 task id & task name & & & & & & & & \\
\midrule
146818 & Australi... & 0.941(0.018)$^{\hspace{0.4em}}$ & 0.942(0.017)$^{\hspace{0.4em}}$ & 0.942(0.017)$^{\hspace{0.4em}}$ & 0.931(0.022)$^{\hspace{0.4em}}$ & 0.936(0.019)$^{\hspace{0.4em}}$ & 0.938(0.023)$^{\hspace{0.4em}}$ & 0.941(0.022)$^{1}$ & 0.935(0.025)$^{\hspace{0.4em}}$ \\
146820 & wilt & 0.995(0.008)$^{\hspace{0.4em}}$ & 0.994(0.010)$^{\hspace{0.4em}}$ & 0.994(0.010)$^{\hspace{0.4em}}$ & 0.994(0.009)$^{\hspace{0.4em}}$ & 0.995(0.007)$^{\hspace{0.4em}}$ & 0.991(0.011)$^{\hspace{0.4em}}$ & 0.996(0.004)$^{\hspace{0.4em}}$ & 0.992(0.010)$^{\hspace{0.4em}}$ \\
167120 & numerai2... & 0.531(0.004)$^{\hspace{0.4em}}$ & 0.528(0.004)$^{\hspace{0.4em}}$ & 0.529(0.004)$^{\hspace{0.4em}}$ & 0.530(0.004)$^{\hspace{0.4em}}$ & 0.531(0.004)$^{\hspace{0.4em}}$ & 0.528(0.004)$^{\hspace{0.4em}}$ & 0.530(0.006)$^{\hspace{0.4em}}$ & 0.531(0.004)$^{\hspace{0.4em}}$ \\
168350 & phoneme & 0.968(0.009)$^{\hspace{0.4em}}$ & 0.968(0.009)$^{\hspace{0.4em}}$ & 0.968(0.009)$^{\hspace{0.4em}}$ & 0.963(0.009)$^{\hspace{0.4em}}$ & 0.970(0.009)$^{\hspace{0.4em}}$ & 0.972(0.008)$^{\hspace{0.4em}}$ & 0.971(0.010)$^{2}$ & 0.968(0.009)$^{\hspace{0.4em}}$ \\
168757 & credit-g & 0.796(0.041)$^{\hspace{0.4em}}$ & 0.788(0.044)$^{\hspace{0.4em}}$ & 0.763(0.053)$^{\hspace{0.4em}}$ & 0.778(0.030)$^{\hspace{0.4em}}$ & 0.796(0.041)$^{\hspace{0.4em}}$ & 0.788(0.039)$^{\hspace{0.4em}}$ & 0.794(0.033)$^{\hspace{0.4em}}$ & 0.779(0.048)$^{\hspace{0.4em}}$ \\
168868 & APSFailu... & 0.993(0.002)$^{\hspace{0.4em}}$ & 0.993(0.002)$^{\hspace{0.4em}}$ & 0.993(0.002)$^{\hspace{0.4em}}$ & 0.993(0.002)$^{\hspace{0.4em}}$ & 0.992(0.001)$^{\hspace{0.4em}}$ & 0.992(0.002)$^{\hspace{0.4em}}$ & 0.990(0.003)$^{\hspace{0.4em}}$ & 0.993(0.002)$^{\hspace{0.4em}}$ \\
\bottomrule
\end{tabular}
\caption{Results for binary classification (in AUC) on a one hour budget,denoted as \texttt{mean}(\texttt{std})$^{\mbox{\texttt{fails}}}$. }
\label{tab:auc-1h8c_gp3-A-H}
\end{table}
\end{landscape}
\footnotesize
\begin{landscape}
\begin{table}
\tiny
\begin{tabular}{rlrrrrrrr}
\toprule
 & framework& lightautoml\ \ & MLJAR(B)\ \ & MLJAR(P)\ \ & TPOT\ \ & constantpredictor\ \ & RandomForest\ \ & TunedRandomForest\text{*}\ \ \\
 task id & task name & & & & & & & \\
\midrule
146818 & Australi... & 0.946(0.020)$^{\hspace{0.4em}}$ & 0.943(0.020)$^{\hspace{0.4em}}$ & 0.944(0.017)$^{\hspace{0.4em}}$ & 0.939(0.022)$^{\hspace{0.4em}}$ & 0.500(0.000)$^{\hspace{0.4em}}$ & 0.940(0.021)$^{\hspace{0.4em}}$ & 0.938(0.022)$^{\hspace{0.4em}}$ \\
146820 & wilt & 0.994(0.007)$^{\hspace{0.4em}}$ & 0.999(0.000)$^{8}$ & 0.995(0.009)$^{\hspace{0.4em}}$ & 0.996(0.004)$^{\hspace{0.4em}}$ & 0.500(0.000)$^{\hspace{0.4em}}$ & 0.989(0.012)$^{\hspace{0.4em}}$ & 0.991(0.010)$^{\hspace{0.4em}}$ \\
167120 & numerai2... & 0.531(0.004)$^{\hspace{0.4em}}$ & 0.530(0.004)$^{\hspace{0.4em}}$ & 0.531(0.005)$^{\hspace{0.4em}}$ & 0.528(0.007)$^{\hspace{0.4em}}$ & 0.500(0.000)$^{\hspace{0.4em}}$ & 0.519(0.006)$^{\hspace{0.4em}}$ & 0.520(0.006)$^{\hspace{0.4em}}$ \\
168350 & phoneme & 0.966(0.008)$^{\hspace{0.4em}}$ & -$\hspace{0.4em}$ & 0.967(0.008)$^{\hspace{0.4em}}$ & 0.969(0.008)$^{\hspace{0.4em}}$ & 0.500(0.000)$^{\hspace{0.4em}}$ & 0.965(0.009)$^{\hspace{0.4em}}$ & 0.966(0.009)$^{\hspace{0.4em}}$ \\
168757 & credit-g & 0.796(0.037)$^{\hspace{0.4em}}$ & -$\hspace{0.4em}$ & 0.785(0.046)$^{\hspace{0.4em}}$ & 0.791(0.052)$^{\hspace{0.4em}}$ & 0.500(0.000)$^{\hspace{0.4em}}$ & 0.801(0.038)$^{\hspace{0.4em}}$ & 0.799(0.035)$^{\hspace{0.4em}}$ \\
168868 & APSFailu... & 0.993(0.002)$^{\hspace{0.4em}}$ & 0.995(nan)$^{9}$ & 0.992(0.002)$^{\hspace{0.4em}}$ & 0.989(0.004)$^{1}$ & 0.500(0.000)$^{\hspace{0.4em}}$ & 0.991(0.002)$^{\hspace{0.4em}}$ & 0.992(0.003)$^{\hspace{0.4em}}$ \\
\bottomrule
\end{tabular}
\caption{Results for binary classification (in AUC) on a one hour budget,denoted as \texttt{mean}(\texttt{std})$^{\mbox{\texttt{fails}}}$. }
\label{tab:auc-1h8c_gp3-I-Z}
\end{table}
\end{landscape}
\footnotesize
\begin{landscape}
\begin{table}
\tiny
\begin{tabular}{rlrrrrrrr}
\toprule
 & framework& AutoGluon(B)\ \ & AutoGluon(HQ)\ \ & autosklearn\ \ & autosklearn2\text{*}\ \ & flaml\ \ & GAMA(B)\text{*}\ \ & H2OAutoML\ \ \\
 task id & task name & & & & & & & \\
\midrule
146818 & Australi... & 0.941(0.018)$^{\hspace{0.4em}}$ & 0.942(0.017)$^{\hspace{0.4em}}$ & 0.931(0.023)$^{\hspace{0.4em}}$ & 0.940(0.020)$^{\hspace{0.4em}}$ & 0.941(0.025)$^{\hspace{0.4em}}$ & 0.940(0.019)$^{\hspace{0.4em}}$ & 0.935(0.021)$^{\hspace{0.4em}}$ \\
146820 & wilt & 0.995(0.008)$^{\hspace{0.4em}}$ & 0.994(0.010)$^{\hspace{0.4em}}$ & 0.994(0.009)$^{\hspace{0.4em}}$ & 0.995(0.008)$^{\hspace{0.4em}}$ & 0.991(0.010)$^{\hspace{0.4em}}$ & 0.996(0.004)$^{\hspace{0.4em}}$ & 0.994(0.008)$^{\hspace{0.4em}}$ \\
167120 & numerai2... & 0.531(0.004)$^{\hspace{0.4em}}$ & 0.529(0.004)$^{\hspace{0.4em}}$ & 0.530(0.005)$^{\hspace{0.4em}}$ & 0.531(0.004)$^{\hspace{0.4em}}$ & 0.529(0.004)$^{\hspace{0.4em}}$ & 0.532(0.004)$^{1}$ & 0.531(0.004)$^{\hspace{0.4em}}$ \\
168350 & phoneme & 0.968(0.009)$^{\hspace{0.4em}}$ & 0.968(0.009)$^{\hspace{0.4em}}$ & 0.964(0.009)$^{\hspace{0.4em}}$ & 0.970(0.009)$^{\hspace{0.4em}}$ & 0.972(0.009)$^{\hspace{0.4em}}$ & 0.971(0.009)$^{\hspace{0.4em}}$ & 0.967(0.010)$^{\hspace{0.4em}}$ \\
168757 & credit-g & 0.796(0.041)$^{\hspace{0.4em}}$ & 0.788(0.044)$^{\hspace{0.4em}}$ & 0.785(0.042)$^{\hspace{0.4em}}$ & 0.795(0.038)$^{\hspace{0.4em}}$ & 0.783(0.045)$^{\hspace{0.4em}}$ & 0.791(0.030)$^{\hspace{0.4em}}$ & 0.783(0.043)$^{\hspace{0.4em}}$ \\
168868 & APSFailu... & 0.993(0.002)$^{\hspace{0.4em}}$ & 0.992(0.002)$^{\hspace{0.4em}}$ & 0.993(0.002)$^{\hspace{0.4em}}$ & 0.992(0.003)$^{\hspace{0.4em}}$ & 0.992(0.002)$^{1}$ & 0.992(0.002)$^{\hspace{0.4em}}$ & 0.992(0.002)$^{\hspace{0.4em}}$ \\
\bottomrule
\end{tabular}
\caption{Results for binary classification (in AUC) on a four hour budget,denoted as \texttt{mean}(\texttt{std})$^{\mbox{\texttt{fails}}}$. Results obtained in 2021 are denoted with a `\text{*}' next to the framework name.}
\label{tab:auc-4h8c_gp3-A-H}
\end{table}
\end{landscape}
\footnotesize
\begin{landscape}
\begin{table}
\tiny
\begin{tabular}{rlrrrrrr}
\toprule
 & framework& lightautoml\ \ & MLJAR(B)\text{*}\ \ & TPOT\text{*}\ \ & constantpredictor\ \ & RandomForest\ \ & TunedRandomForest\text{*}\ \ \\
 task id & task name & & & & & & \\
\midrule
146818 & Australi... & 0.946(0.020)$^{\hspace{0.4em}}$ & 0.940(0.024)$^{\hspace{0.4em}}$ & 0.936(0.024)$^{\hspace{0.4em}}$ & 0.500(0.000)$^{\hspace{0.4em}}$ & 0.940(0.021)$^{\hspace{0.4em}}$ & 0.938(0.022)$^{\hspace{0.4em}}$ \\
146820 & wilt & 0.994(0.007)$^{\hspace{0.4em}}$ & 0.994(0.003)$^{5}$ & 0.985(0.025)$^{\hspace{0.4em}}$ & 0.500(0.000)$^{\hspace{0.4em}}$ & 0.989(0.012)$^{\hspace{0.4em}}$ & 0.990(0.011)$^{\hspace{0.4em}}$ \\
167120 & numerai2... & 0.531(0.004)$^{\hspace{0.4em}}$ & 0.530(0.004)$^{\hspace{0.4em}}$ & 0.527(0.006)$^{\hspace{0.4em}}$ & 0.500(0.000)$^{\hspace{0.4em}}$ & 0.519(0.006)$^{\hspace{0.4em}}$ & 0.520(0.006)$^{\hspace{0.4em}}$ \\
168350 & phoneme & 0.965(0.008)$^{\hspace{0.4em}}$ & -$\hspace{0.4em}$ & 0.971(0.009)$^{\hspace{0.4em}}$ & 0.500(0.000)$^{\hspace{0.4em}}$ & 0.965(0.009)$^{\hspace{0.4em}}$ & 0.966(0.009)$^{\hspace{0.4em}}$ \\
168757 & credit-g & 0.797(0.043)$^{\hspace{0.4em}}$ & -$\hspace{0.4em}$ & 0.787(0.034)$^{\hspace{0.4em}}$ & 0.500(0.000)$^{\hspace{0.4em}}$ & 0.801(0.038)$^{\hspace{0.4em}}$ & 0.802(0.034)$^{\hspace{0.4em}}$ \\
168868 & APSFailu... & 0.993(0.002)$^{\hspace{0.4em}}$ & 0.993(0.002)$^{6}$ & 0.989(0.003)$^{1}$ & 0.500(0.000)$^{\hspace{0.4em}}$ & 0.991(0.002)$^{\hspace{0.4em}}$ & 0.990(0.003)$^{\hspace{0.4em}}$ \\
\bottomrule
\end{tabular}
\caption{Results for binary classification (in AUC) on a four hour budget,denoted as \texttt{mean}(\texttt{std})$^{\mbox{\texttt{fails}}}$. Results obtained in 2021 are denoted with a `\text{*}' next to the framework name.}
\label{tab:auc-4h8c_gp3-I-Z}
\end{table}
\end{landscape}

\footnotesize
\begin{landscape}
\begin{table}
\tiny
\begin{tabular}{rlrrrrrrrr}
\toprule
 & framework& AutoGluon(B)\ \ & AutoGluon(HQ)\ \ & AutoGluon(HQIL)\ \ & autosklearn\ \ & autosklearn2\ \ & flaml\ \ & GAMA(B)\ \ & H2OAutoML\ \ \\
 task id & task name & & & & & & & & \\
\midrule
10090 & amazon-c... & 0.695(0.086)$^{\hspace{0.4em}}$ & 0.747(0.104)$^{\hspace{0.4em}}$ & 1.187(0.119)$^{\hspace{0.4em}}$ & 1.139(0.132)$^{\hspace{0.4em}}$ & 1.125(0.146)$^{\hspace{0.4em}}$ & 1.115(0.155)$^{\hspace{0.4em}}$ & 0.910(0.066)$^{\hspace{0.4em}}$ & 1.077(0.092)$^{\hspace{0.4em}}$ \\
168784 & steel-pl... & 0.464(0.042)$^{\hspace{0.4em}}$ & 0.465(0.050)$^{\hspace{0.4em}}$ & 0.507(0.037)$^{\hspace{0.4em}}$ & 0.516(0.032)$^{\hspace{0.4em}}$ & 0.457(0.027)$^{\hspace{0.4em}}$ & 0.482(0.037)$^{\hspace{0.4em}}$ & 0.491(0.029)$^{\hspace{0.4em}}$ & 0.490(0.036)$^{\hspace{0.4em}}$ \\
168909 & dilbert & 0.014(0.006)$^{\hspace{0.4em}}$ & 0.013(0.007)$^{\hspace{0.4em}}$ & 0.014(0.007)$^{\hspace{0.4em}}$ & 0.033(0.012)$^{1}$ & 0.037(0.007)$^{\hspace{0.4em}}$ & 0.024(0.008)$^{\hspace{0.4em}}$ & 0.176(0.033)$^{\hspace{0.4em}}$ & 0.065(0.007)$^{\hspace{0.4em}}$ \\
168910 & fabert & 0.683(0.030)$^{\hspace{0.4em}}$ & 0.684(0.030)$^{\hspace{0.4em}}$ & -$\hspace{0.4em}$ & 0.745(0.034)$^{\hspace{0.4em}}$ & 0.725(0.029)$^{\hspace{0.4em}}$ & 0.766(0.025)$^{\hspace{0.4em}}$ & 0.763(0.032)$^{\hspace{0.4em}}$ & 0.746(0.029)$^{\hspace{0.4em}}$ \\
189355 & dionis & 0.266(0.004)$^{\hspace{0.4em}}$ & 0.302(0.009)$^{\hspace{0.4em}}$ & 0.299(0.010)$^{\hspace{0.4em}}$ & 0.731(0.169)$^{\hspace{0.4em}}$ & 1.843(1.453)$^{\hspace{0.4em}}$ & 0.495(0.287)$^{\hspace{0.4em}}$ & 1.446(0.145)$^{\hspace{0.4em}}$ & 3.331(0.098)$^{5}$ \\
190146 & vehicle & 0.312(0.050)$^{\hspace{0.4em}}$ & 0.341(0.060)$^{\hspace{0.4em}}$ & -$\hspace{0.4em}$ & 0.364(0.036)$^{\hspace{0.4em}}$ & 0.322(0.040)$^{\hspace{0.4em}}$ & 0.439(0.041)$^{\hspace{0.4em}}$ & 0.378(0.056)$^{\hspace{0.4em}}$ & 0.351(0.077)$^{\hspace{0.4em}}$ \\
\bottomrule
\end{tabular}
\caption{Results for multiclass classification (in logloss) on a one hour budget,denoted as \texttt{mean}(\texttt{std})$^{\mbox{\texttt{fails}}}$. }
\label{tab:logloss-1h8c_gp3-A-H}
\end{table}
\end{landscape}
\footnotesize
\begin{landscape}
\begin{table}
\tiny
\begin{tabular}{rlrrrrrrr}
\toprule
 & framework& lightautoml\ \ & MLJAR(B)\ \ & MLJAR(P)\ \ & TPOT\ \ & constantpredictor\ \ & RandomForest\ \ & TunedRandomForest\text{*}\ \ \\
 task id & task name & & & & & & & \\
\midrule
10090 & amazon-c... & 0.843(0.088)$^{\hspace{0.4em}}$ & 1.169(0.133)$^{\hspace{0.4em}}$ & 1.252(0.178)$^{\hspace{0.4em}}$ & 1.111(0.268)$^{5}$ & 3.912(0.000)$^{\hspace{0.4em}}$ & 2.057(0.068)$^{\hspace{0.4em}}$ & 1.462(0.087)$^{\hspace{0.4em}}$ \\
168784 & steel-pl... & 0.478(0.032)$^{\hspace{0.4em}}$ & 0.468(0.026)$^{\hspace{0.4em}}$ & 0.484(0.030)$^{\hspace{0.4em}}$ & 0.509(0.044)$^{\hspace{0.4em}}$ & 1.671(0.007)$^{\hspace{0.4em}}$ & 0.592(0.061)$^{\hspace{0.4em}}$ & 0.540(0.047)$^{\hspace{0.4em}}$ \\
168909 & dilbert & 0.033(0.006)$^{\hspace{0.4em}}$ & 0.028(0.012)$^{\hspace{0.4em}}$ & 0.030(0.009)$^{\hspace{0.4em}}$ & 0.150(0.072)$^{\hspace{0.4em}}$ & 1.609(0.000)$^{\hspace{0.4em}}$ & 0.329(0.010)$^{\hspace{0.4em}}$ & 0.302(0.009)$^{\hspace{0.4em}}$ \\
168910 & fabert & 0.768(0.035)$^{\hspace{0.4em}}$ & 0.755(0.033)$^{\hspace{0.4em}}$ & 0.771(0.024)$^{\hspace{0.4em}}$ & 0.886(0.069)$^{3}$ & 1.875(0.001)$^{\hspace{0.4em}}$ & 0.808(0.030)$^{\hspace{0.4em}}$ & 0.804(0.024)$^{\hspace{0.4em}}$ \\
189355 & dionis & -$\hspace{0.4em}$ & 1.234(0.103)$^{5}$ & 1.135(0.045)$^{\hspace{0.4em}}$ & 4.241(0.403)$^{8}$ & 5.857(0.000)$^{\hspace{0.4em}}$ & 0.882(0.019)$^{\hspace{0.4em}}$ & 0.872(0.020)$^{\hspace{0.4em}}$ \\
190146 & vehicle & 0.389(0.074)$^{\hspace{0.4em}}$ & 0.321(0.037)$^{\hspace{0.4em}}$ & 0.349(0.034)$^{\hspace{0.4em}}$ & 0.417(0.062)$^{\hspace{0.4em}}$ & 1.386(0.000)$^{\hspace{0.4em}}$ & 0.498(0.035)$^{\hspace{0.4em}}$ & 0.484(0.040)$^{\hspace{0.4em}}$ \\
\bottomrule
\end{tabular}
\caption{Results for multiclass classification (in logloss) on a one hour budget,denoted as \texttt{mean}(\texttt{std})$^{\mbox{\texttt{fails}}}$. }
\label{tab:logloss-1h8c_gp3-I-Z}
\end{table}
\end{landscape}
\footnotesize
\begin{landscape}
\begin{table}
\tiny
\begin{tabular}{rlrrrrrrr}
\toprule
 & framework& AutoGluon(B)\ \ & AutoGluon(HQ)\ \ & autosklearn\ \ & autosklearn2\text{*}\ \ & flaml\ \ & GAMA(B)\text{*}\ \ & H2OAutoML\ \ \\
 task id & task name & & & & & & & \\
\midrule
10090 & amazon-c... & 0.673(0.072)$^{\hspace{0.4em}}$ & 0.707(0.079)$^{\hspace{0.4em}}$ & 1.138(0.130)$^{\hspace{0.4em}}$ & 0.837(0.122)$^{\hspace{0.4em}}$ & 1.101(0.150)$^{\hspace{0.4em}}$ & 0.907(0.094)$^{\hspace{0.4em}}$ & 1.077(0.092)$^{\hspace{0.4em}}$ \\
168784 & steel-pl... & 0.462(0.043)$^{\hspace{0.4em}}$ & 0.463(0.048)$^{\hspace{0.4em}}$ & 0.515(0.035)$^{\hspace{0.4em}}$ & 0.472(0.029)$^{\hspace{0.4em}}$ & 0.500(0.044)$^{\hspace{0.4em}}$ & 0.491(0.039)$^{\hspace{0.4em}}$ & 0.495(0.033)$^{\hspace{0.4em}}$ \\
168909 & dilbert & 0.013(0.005)$^{\hspace{0.4em}}$ & 0.013(0.006)$^{\hspace{0.4em}}$ & 0.029(0.010)$^{\hspace{0.4em}}$ & 0.029(0.008)$^{\hspace{0.4em}}$ & 0.024(0.011)$^{\hspace{0.4em}}$ & 0.115(0.044)$^{\hspace{0.4em}}$ & 0.032(0.010)$^{\hspace{0.4em}}$ \\
168910 & fabert & 0.682(0.030)$^{\hspace{0.4em}}$ & 0.683(0.030)$^{\hspace{0.4em}}$ & 0.748(0.029)$^{\hspace{0.4em}}$ & 0.733(0.026)$^{\hspace{0.4em}}$ & 0.774(0.031)$^{\hspace{0.4em}}$ & 0.737(0.029)$^{\hspace{0.4em}}$ & 0.726(0.027)$^{\hspace{0.4em}}$ \\
189355 & dionis & 0.285(0.004)$^{\hspace{0.4em}}$ & 0.323(0.004)$^{\hspace{0.4em}}$ & 0.544(0.064)$^{\hspace{0.4em}}$ & 0.523(0.144)$^{\hspace{0.4em}}$ & 0.335(0.004)$^{8}$ & 1.587(0.244)$^{\hspace{0.4em}}$ & 1.568(0.019)$^{2}$ \\
190146 & vehicle & 0.312(0.050)$^{\hspace{0.4em}}$ & 0.342(0.061)$^{\hspace{0.4em}}$ & 0.366(0.041)$^{\hspace{0.4em}}$ & 0.329(0.030)$^{\hspace{0.4em}}$ & 0.438(0.042)$^{\hspace{0.4em}}$ & 0.369(0.032)$^{\hspace{0.4em}}$ & 0.336(0.062)$^{\hspace{0.4em}}$ \\
\bottomrule
\end{tabular}
\caption{Results for multiclass classification (in logloss) on a four hour budget,denoted as \texttt{mean}(\texttt{std})$^{\mbox{\texttt{fails}}}$. Results obtained in 2021 are denoted with a `\text{*}' next to the framework name.}
\label{tab:logloss-4h8c_gp3-A-H}
\end{table}
\end{landscape}
\footnotesize
\begin{landscape}
\begin{table}
\tiny
\begin{tabular}{rlrrrrrr}
\toprule
 & framework& lightautoml\ \ & MLJAR(B)\text{*}\ \ & TPOT\text{*}\ \ & constantpredictor\ \ & RandomForest\ \ & TunedRandomForest\text{*}\ \ \\
 task id & task name & & & & & & \\
\midrule
10090 & amazon-c... & 0.817(0.077)$^{\hspace{0.4em}}$ & 1.181(0.132)$^{\hspace{0.4em}}$ & 0.852(0.159)$^{2}$ & 3.912(0.000)$^{\hspace{0.4em}}$ & 2.057(0.068)$^{\hspace{0.4em}}$ & 1.462(0.115)$^{\hspace{0.4em}}$ \\
168784 & steel-pl... & 0.475(0.030)$^{\hspace{0.4em}}$ & 0.467(0.032)$^{\hspace{0.4em}}$ & 0.486(0.021)$^{\hspace{0.4em}}$ & 1.671(0.007)$^{\hspace{0.4em}}$ & 0.592(0.061)$^{\hspace{0.4em}}$ & 0.538(0.047)$^{\hspace{0.4em}}$ \\
168909 & dilbert & 0.025(0.009)$^{\hspace{0.4em}}$ & 0.024(0.009)$^{\hspace{0.4em}}$ & 0.060(0.022)$^{\hspace{0.4em}}$ & 1.609(0.000)$^{\hspace{0.4em}}$ & 0.329(0.010)$^{\hspace{0.4em}}$ & 0.300(0.009)$^{\hspace{0.4em}}$ \\
168910 & fabert & 0.778(0.033)$^{\hspace{0.4em}}$ & 0.752(0.025)$^{\hspace{0.4em}}$ & 0.795(0.049)$^{\hspace{0.4em}}$ & 1.875(0.001)$^{\hspace{0.4em}}$ & 0.808(0.030)$^{\hspace{0.4em}}$ & 0.809(0.030)$^{\hspace{0.4em}}$ \\
189355 & dionis & -$\hspace{0.4em}$ & -$\hspace{0.4em}$ & -$\hspace{0.4em}$ & 5.857(0.000)$^{\hspace{0.4em}}$ & 0.891(0.031)$^{\hspace{0.4em}}$ & 0.870(0.017)$^{\hspace{0.4em}}$ \\
190146 & vehicle & 0.401(0.069)$^{\hspace{0.4em}}$ & 0.321(0.043)$^{\hspace{0.4em}}$ & 0.339(0.065)$^{\hspace{0.4em}}$ & 1.386(0.000)$^{\hspace{0.4em}}$ & 0.498(0.035)$^{\hspace{0.4em}}$ & 0.483(0.039)$^{\hspace{0.4em}}$ \\
\bottomrule
\end{tabular}
\caption{Results for multiclass classification (in logloss) on a four hour budget,denoted as \texttt{mean}(\texttt{std})$^{\mbox{\texttt{fails}}}$. Results obtained in 2021 are denoted with a `\text{*}' next to the framework name.}
\label{tab:logloss-4h8c_gp3-I-Z}
\end{table}
\end{landscape}

\footnotesize
\begin{landscape}
\begin{table}
\tiny
\begin{tabular}{rlrrrrrrr}
\toprule
 & framework& AutoGluon(B)\ \ & AutoGluon(HQ)\ \ & AutoGluon(HQIL)\ \ & autosklearn\ \ & flaml\ \ & GAMA(B)\ \ & H2OAutoML\ \ \\
 task id & task name & & & & & & & \\
\midrule
167210 & Moneyball & 21(0.87)$^{\hspace{0.4em}}$ & 21(0.87)$^{\hspace{0.4em}}$ & 21(0.87)$^{\hspace{0.4em}}$ & 21(0.75)$^{\hspace{0.4em}}$ & 22(0.62)$^{\hspace{0.4em}}$ & 2e+10(6.3e+10)$^{\hspace{0.4em}}$ & 21(0.93)$^{\hspace{0.4em}}$ \\
233211 & diamonds & 5e+02(19)$^{\hspace{0.4em}}$ & 5.3e+02(36)$^{\hspace{0.4em}}$ & 5.2e+02(22)$^{\hspace{0.4em}}$ & 5.2e+02(21)$^{\hspace{0.4em}}$ & 5.2e+02(19)$^{\hspace{0.4em}}$ & 5.3e+02(21)$^{1}$ & 5.2e+02(20)$^{\hspace{0.4em}}$ \\
233212 & Allstate... & 1.9e+03(44)$^{\hspace{0.4em}}$ & 1.9e+03(36)$^{\hspace{0.4em}}$ & 1.9e+03(39)$^{\hspace{0.4em}}$ & 1.9e+03(58)$^{\hspace{0.4em}}$ & 1.9e+03(54)$^{\hspace{0.4em}}$ & 2e+03(54)$^{\hspace{0.4em}}$ & 1.9e+03(48)$^{\hspace{0.4em}}$ \\
233213 & Buzzinso... & 1.5e+02(49)$^{\hspace{0.4em}}$ & 1.5e+02(49)$^{\hspace{0.4em}}$ & 1.5e+02(49)$^{\hspace{0.4em}}$ & 1.6e+02(52)$^{\hspace{0.4em}}$ & 1.5e+02(47)$^{\hspace{0.4em}}$ & 1.6e+02(45)$^{\hspace{0.4em}}$ & 1.5e+02(46)$^{\hspace{0.4em}}$ \\
233214 & Santande... & 6.8e+06(4.4e+05)$^{\hspace{0.4em}}$ & 6.8e+06(4.4e+05)$^{\hspace{0.4em}}$ & 7.1e+06(4.5e+05)$^{\hspace{0.4em}}$ & 6.8e+06(5.1e+05)$^{\hspace{0.4em}}$ & 7e+06(4.5e+05)$^{\hspace{0.4em}}$ & 7e+06(4.8e+05)$^{\hspace{0.4em}}$ & 7e+06(4.9e+05)$^{\hspace{0.4em}}$ \\
233215 & Mercedes... & 8.6(1)$^{\hspace{0.4em}}$ & 8.8(1)$^{\hspace{0.4em}}$ & 8.8(1)$^{\hspace{0.4em}}$ & 8.3(1.1)$^{\hspace{0.4em}}$ & 8.3(1.1)$^{\hspace{0.4em}}$ & 8.3(1.1)$^{\hspace{0.4em}}$ & 8.3(1.1)$^{\hspace{0.4em}}$ \\
\bottomrule
\end{tabular}
\caption{Results for regression (in RMSE) on a one hour budget,denoted as \texttt{mean}(\texttt{std})$^{\mbox{\texttt{fails}}}$. }
\label{tab:rmse-1h8c_gp3-A-H}
\end{table}
\end{landscape}
\footnotesize
\begin{landscape}
\begin{table}
\tiny
\begin{tabular}{rlrrrrrrr}
\toprule
 & framework& lightautoml\ \ & MLJAR(B)\ \ & MLJAR(P)\ \ & TPOT\ \ & constantpredictor\ \ & RandomForest\ \ & TunedRandomForest\text{*}\ \ \\
 task id & task name & & & & & & & \\
\midrule
167210 & Moneyball & 21(0.72)$^{\hspace{0.4em}}$ & 21(0.83)$^{\hspace{0.4em}}$ & 21(0.66)$^{\hspace{0.4em}}$ & 21(1.1)$^{1}$ & 92(1.9)$^{\hspace{0.4em}}$ & 24(1.2)$^{\hspace{0.4em}}$ & 24(1.2)$^{\hspace{0.4em}}$ \\
233211 & diamonds & 5.2e+02(21)$^{\hspace{0.4em}}$ & 5.1e+02(21)$^{\hspace{0.4em}}$ & 5.1e+02(21)$^{\hspace{0.4em}}$ & 5.3e+02(22)$^{\hspace{0.4em}}$ & 4.3e+03(67)$^{\hspace{0.4em}}$ & 5.4e+02(20)$^{\hspace{0.4em}}$ & 5.4e+02(19)$^{\hspace{0.4em}}$ \\
233212 & Allstate... & 1.9e+03(49)$^{\hspace{0.4em}}$ & 1.9e+03(58)$^{\hspace{0.4em}}$ & 1.9e+03(58)$^{\hspace{0.4em}}$ & 2.1e+03(1.3e+02)$^{6}$ & 3e+03(63)$^{\hspace{0.4em}}$ & 2e+03(57)$^{\hspace{0.4em}}$ & 1.9e+03(53)$^{\hspace{0.4em}}$ \\
233213 & Buzzinso... & 1.6e+02(45)$^{\hspace{0.4em}}$ & 1.5e+02(53)$^{\hspace{0.4em}}$ & 1.7e+02(52)$^{\hspace{0.4em}}$ & 1.7e+02(39)$^{2}$ & 6.3e+02(43)$^{\hspace{0.4em}}$ & 1.5e+02(50)$^{\hspace{0.4em}}$ & 1.5e+02(49)$^{\hspace{0.4em}}$ \\
233214 & Santande... & 6.9e+06(4.4e+05)$^{\hspace{0.4em}}$ & 6.9e+06(4.5e+05)$^{\hspace{0.4em}}$ & 6.9e+06(4.2e+05)$^{\hspace{0.4em}}$ & 7e+06(4.7e+05)$^{1}$ & 9e+06(6.5e+05)$^{\hspace{0.4em}}$ & 7.2e+06(3.8e+05)$^{\hspace{0.4em}}$ & 6.8e+06(4.5e+05)$^{\hspace{0.4em}}$ \\
233215 & Mercedes... & 8.3(1.1)$^{\hspace{0.4em}}$ & 8.3(1.1)$^{\hspace{0.4em}}$ & 8.3(1.1)$^{\hspace{0.4em}}$ & 8.3(1.1)$^{\hspace{0.4em}}$ & 13(0.72)$^{\hspace{0.4em}}$ & 8.9(0.93)$^{\hspace{0.4em}}$ & 8.9(0.95)$^{\hspace{0.4em}}$ \\
\bottomrule
\end{tabular}
\caption{Results for regression (in RMSE) on a one hour budget,denoted as \texttt{mean}(\texttt{std})$^{\mbox{\texttt{fails}}}$. }
\label{tab:rmse-1h8c_gp3-I-Z}
\end{table}
\end{landscape}
\footnotesize
\begin{landscape}
\begin{table}
\tiny
\begin{tabular}{rlrrrrrrr}
\toprule
 & framework& AutoGluon(B)\ \ & AutoGluon(HQ)\ \ & AutoGluon(HQIL)\ \ & autosklearn\ \ & flaml\ \ & GAMA(B)\text{*}\ \ & H2OAutoML\ \ \\
 task id & task name & & & & & & & \\
\midrule
167210 & Moneyball & 21(0.87)$^{\hspace{0.4em}}$ & 21(0.87)$^{\hspace{0.4em}}$ & 21(0.87)$^{\hspace{0.4em}}$ & 21(0.73)$^{\hspace{0.4em}}$ & 22(0.86)$^{\hspace{0.4em}}$ & 21(0.77)$^{\hspace{0.4em}}$ & 21(0.85)$^{\hspace{0.4em}}$ \\
233211 & diamonds & 5e+02(18)$^{\hspace{0.4em}}$ & 5.3e+02(26)$^{\hspace{0.4em}}$ & 5.2e+02(22)$^{\hspace{0.4em}}$ & 5.2e+02(20)$^{\hspace{0.4em}}$ & 5.2e+02(20)$^{\hspace{0.4em}}$ & 5.2e+02(20)$^{\hspace{0.4em}}$ & 5.1e+02(18)$^{\hspace{0.4em}}$ \\
233212 & Allstate... & 1.9e+03(40)$^{\hspace{0.4em}}$ & 1.9e+03(39)$^{\hspace{0.4em}}$ & 1.9e+03(1.3e+02)$^{\hspace{0.4em}}$ & 1.9e+03(59)$^{\hspace{0.4em}}$ & 1.9e+03(55)$^{\hspace{0.4em}}$ & 8.6e+12(2.7e+13)$^{\hspace{0.4em}}$ & 1.9e+03(59)$^{\hspace{0.4em}}$ \\
233213 & Buzzinso... & 1.5e+02(49)$^{\hspace{0.4em}}$ & 2.3e+02(2.9e+02)$^{\hspace{0.4em}}$ & 2.3e+02(2.9e+02)$^{\hspace{0.4em}}$ & 1.5e+02(54)$^{\hspace{0.4em}}$ & 1.5e+02(52)$^{\hspace{0.4em}}$ & 1.6e+02(47)$^{\hspace{0.4em}}$ & 1.5e+02(47)$^{\hspace{0.4em}}$ \\
233214 & Santande... & 6.8e+06(4.6e+05)$^{\hspace{0.4em}}$ & 6.8e+06(4.7e+05)$^{\hspace{0.4em}}$ & 7e+06(4.3e+05)$^{\hspace{0.4em}}$ & 6.9e+06(5.6e+05)$^{\hspace{0.4em}}$ & 6.9e+06(4.8e+05)$^{\hspace{0.4em}}$ & 6.9e+06(4.3e+05)$^{\hspace{0.4em}}$ & 6.9e+06(4.8e+05)$^{\hspace{0.4em}}$ \\
233215 & Mercedes... & 8.6(1)$^{\hspace{0.4em}}$ & 8.8(1)$^{\hspace{0.4em}}$ & 8.8(0.97)$^{\hspace{0.4em}}$ & 8.3(1.1)$^{\hspace{0.4em}}$ & 8.3(1.1)$^{\hspace{0.4em}}$ & 8.3(1.1)$^{\hspace{0.4em}}$ & 8.3(1.1)$^{\hspace{0.4em}}$ \\
\bottomrule
\end{tabular}
\caption{Results for regression (in RMSE) on a four hour budget,denoted as \texttt{mean}(\texttt{std})$^{\mbox{\texttt{fails}}}$. Results obtained in 2021 are denoted with a `\text{*}' next to the framework name.}
\label{tab:rmse-4h8c_gp3-A-H}
\end{table}
\end{landscape}
\footnotesize
\begin{landscape}
\begin{table}
\tiny
\begin{tabular}{rlrrrrrr}
\toprule
 & framework& lightautoml\ \ & MLJAR(B)\ \ & TPOT\text{*}\ \ & constantpredictor\ \ & RandomForest\ \ & TunedRandomForest\text{*}\ \ \\
 task id & task name & & & & & & \\
\midrule
167210 & Moneyball & 21(0.73)$^{\hspace{0.4em}}$ & 21(0.84)$^{\hspace{0.4em}}$ & 21(0.87)$^{\hspace{0.4em}}$ & 92(1.9)$^{\hspace{0.4em}}$ & 24(1.2)$^{\hspace{0.4em}}$ & 24(1.3)$^{\hspace{0.4em}}$ \\
233211 & diamonds & 5.2e+02(21)$^{\hspace{0.4em}}$ & 5.1e+02(22)$^{\hspace{0.4em}}$ & 5.3e+02(23)$^{\hspace{0.4em}}$ & 4.3e+03(67)$^{\hspace{0.4em}}$ & 5.4e+02(20)$^{\hspace{0.4em}}$ & 5.4e+02(20)$^{\hspace{0.4em}}$ \\
233212 & Allstate... & 1.9e+03(49)$^{\hspace{0.4em}}$ & 1.9e+03(54)$^{\hspace{0.4em}}$ & 1.9e+03(50)$^{\hspace{0.4em}}$ & 3e+03(63)$^{\hspace{0.4em}}$ & 2e+03(57)$^{\hspace{0.4em}}$ & 2e+03(79)$^{\hspace{0.4em}}$ \\
233213 & Buzzinso... & 1.6e+02(46)$^{\hspace{0.4em}}$ & 1.5e+02(47)$^{\hspace{0.4em}}$ & 1.6e+02(46)$^{\hspace{0.4em}}$ & 6.3e+02(43)$^{\hspace{0.4em}}$ & 1.5e+02(49)$^{\hspace{0.4em}}$ & 1.5e+02(50)$^{\hspace{0.4em}}$ \\
233214 & Santande... & 7.1e+06(2.7e+05)$^{5}$ & 6.9e+06(4.3e+05)$^{\hspace{0.4em}}$ & 7e+06(4.4e+05)$^{\hspace{0.4em}}$ & 9e+06(6.5e+05)$^{\hspace{0.4em}}$ & 7.2e+06(3.8e+05)$^{\hspace{0.4em}}$ & 6.8e+06(4.6e+05)$^{\hspace{0.4em}}$ \\
233215 & Mercedes... & 8.3(1.1)$^{\hspace{0.4em}}$ & 8.3(1.1)$^{\hspace{0.4em}}$ & 8.3(1.1)$^{\hspace{0.4em}}$ & 13(0.72)$^{\hspace{0.4em}}$ & 8.9(0.93)$^{\hspace{0.4em}}$ & 8.9(0.96)$^{\hspace{0.4em}}$ \\
\bottomrule
\end{tabular}
\caption{Results for regression (in RMSE) on a four hour budget,denoted as \texttt{mean}(\texttt{std})$^{\mbox{\texttt{fails}}}$. Results obtained in 2021 are denoted with a `\text{*}' next to the framework name.}
\label{tab:rmse-4h8c_gp3-I-Z}
\end{table}
\end{landscape}

\subsection{BT-Trees}
\begin{figure}
     \centering
     \begin{subfigure}[b]{\textwidth}
         \centering
         \includegraphics[width=\textwidth]{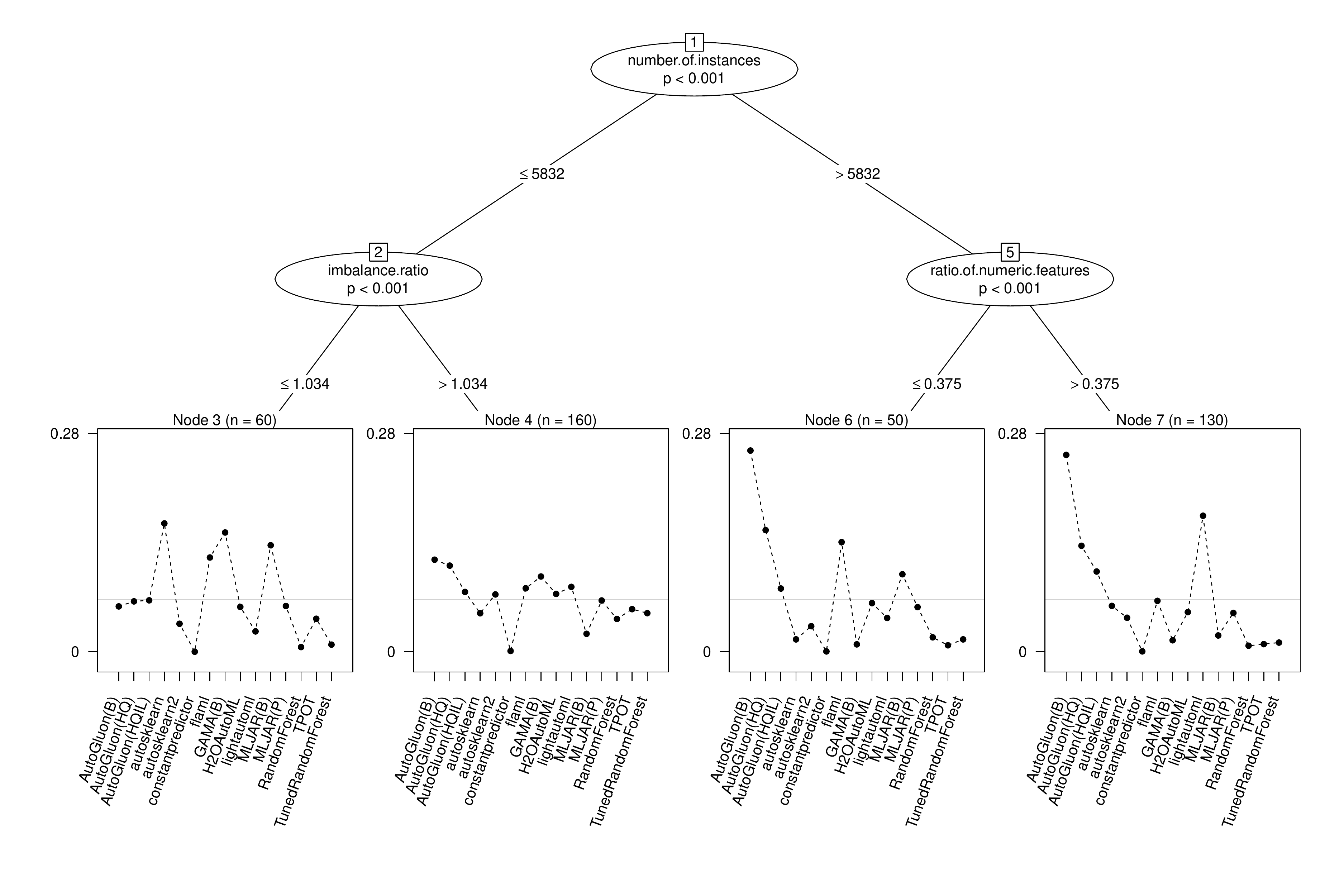}
         \caption{Binary classification datasets, 1h.}
         %\label{fig:three sin x}
     \end{subfigure}
     \begin{subfigure}[b]{\textwidth}
         \centering
         \includegraphics[width=\textwidth]{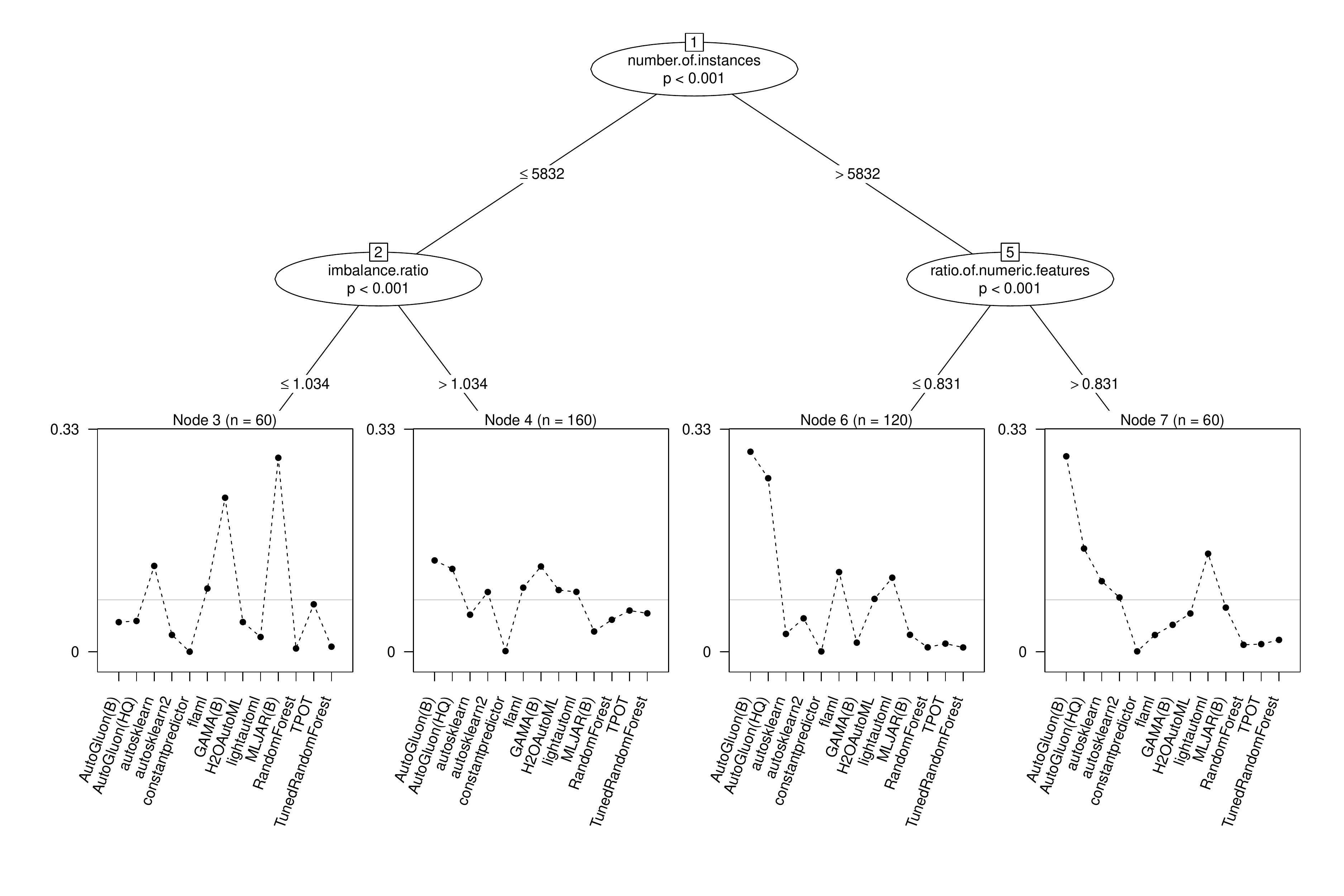}
         \caption{Binary classification datasets, 4h.}
         %\label{fig:three sin x}
     \end{subfigure}
     \caption{Binary classification datasets, 1h and 4h.}
     \label{fig:bt-auc}
\end{figure}

\begin{figure}
     \begin{subfigure}[b]{\textwidth}
         \centering
         \includegraphics[width=\textwidth]{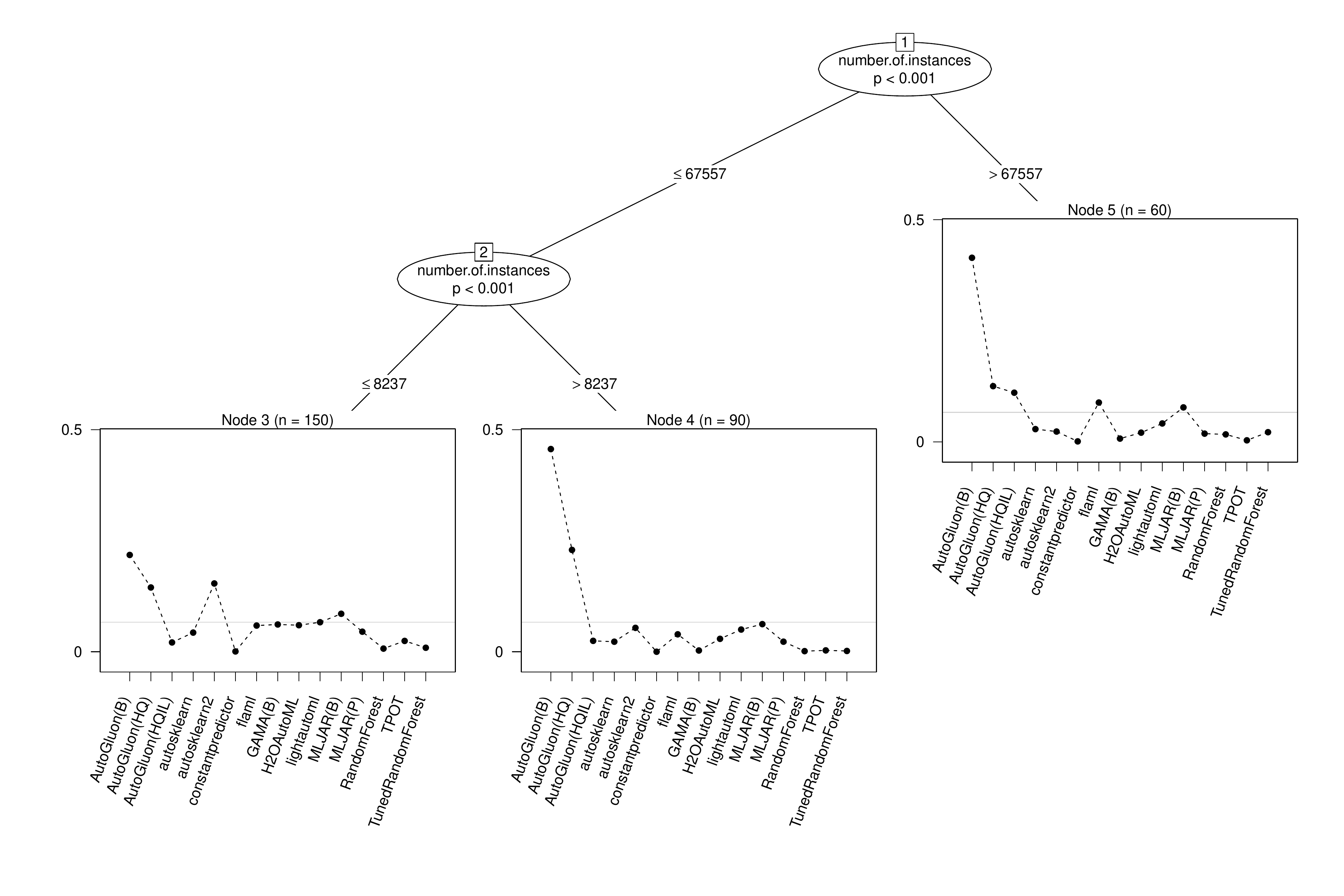}
         \caption{Multiclass classification datasets, 1h.}
     \end{subfigure}
     \begin{subfigure}[b]{\textwidth}
         \centering
         \includegraphics[width=\textwidth]{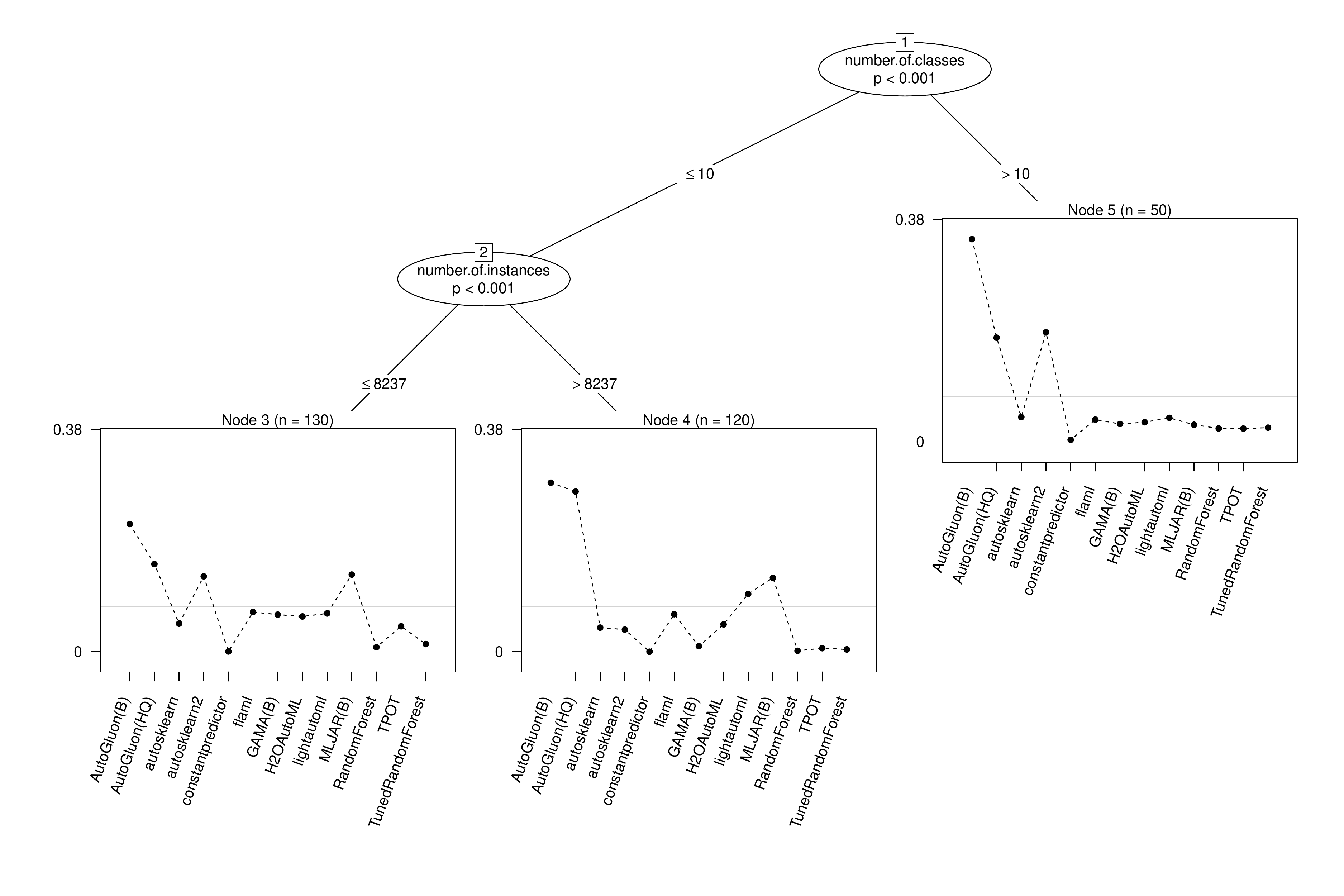}
         \caption{Multiclass classification datasets, 4h.}
     \end{subfigure}
     \caption{Multiclass classification datasets, 1h and 4h.}
     \label{fig:bt-logloss}
\end{figure}

\begin{figure}
     \begin{subfigure}[b]{\textwidth}
         \centering
         \includegraphics[width=\textwidth]{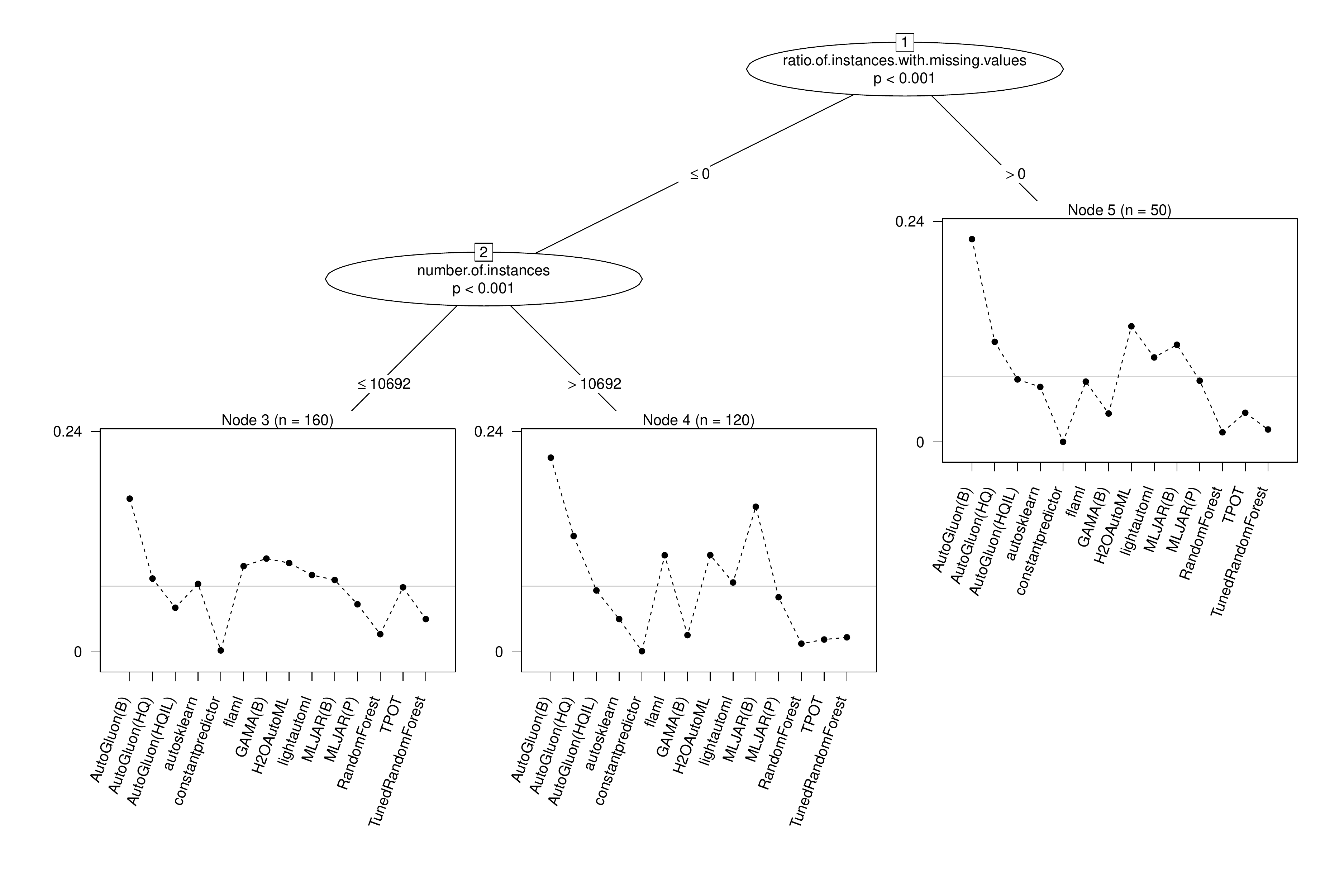}
         \caption{Regression datasets, 1h.}
     \end{subfigure}
     \begin{subfigure}[b]{\textwidth}
         \centering
         \includegraphics[width=\textwidth]{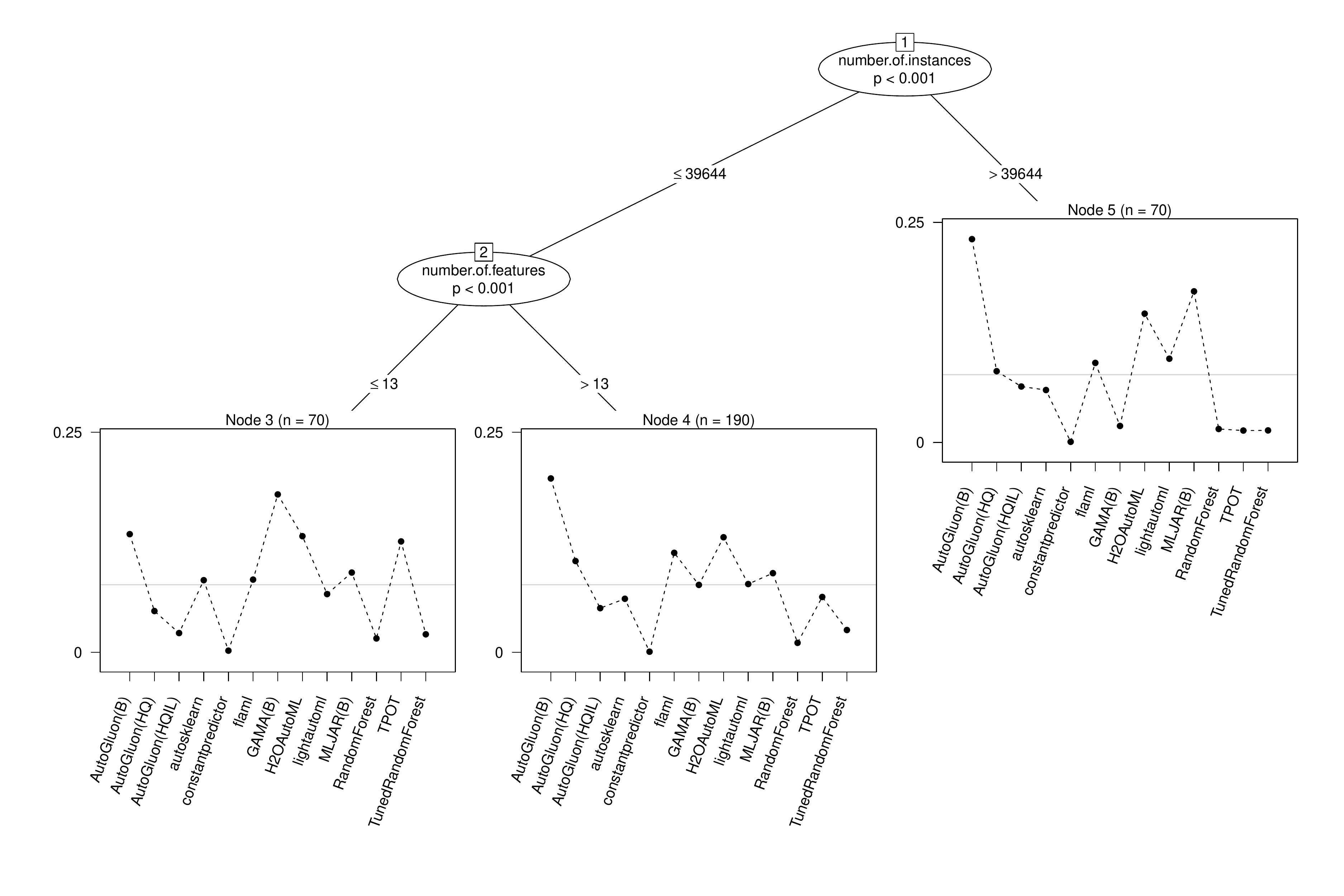}
         \caption{Regression datasets, 4h.}
     \end{subfigure}
     \caption{Regression datasets, 1h and 4h.}
     \label{fig:bt-rmse}
\end{figure}

As described in Section~\ref{sec:bt}, Bradley-Terry (BT) trees may be used to identify subsets of tasks for which the `preferred' framework is significantly different.
Figures~\ref{fig:bt-auc}-\ref{fig:bt-rmse} show BT trees for each task type and time budget, generated by splitting based on different data set characteristics.
We observe that \autogluon{}(B) is the preferred framework in many cases, especially for large number of observations, and more complex data sets in terms of class imbalance, number of classes or missing data.
In contrast, differences in small datasets are generally smaller, and distinct preferred frameworks often emerge for various task types.
The online visualization tool may be used to generate additional BT trees, from different subsets of tasks or allowing for different meta-features when calculating splits.

\clearpage
\section{Software}\label{app:software}
This appendix contains additional details about the developed and used software.

\subsection{Architecture Overview}\label{sec:architecture}

\begin{figure}[htb!]
    \includegraphics[width=\textwidth]{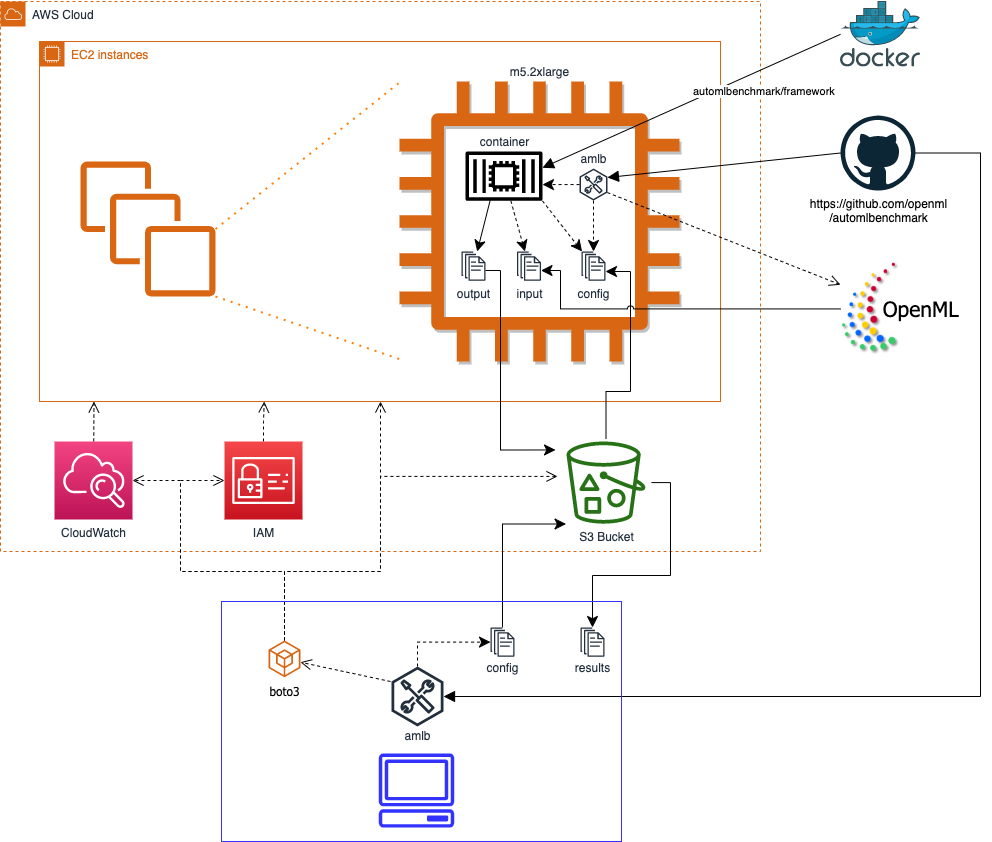}
    \caption{Architecture Overview of the AWS+docker mode, as used for this paper}
    \label{fig:amlb-aws-archi}
\end{figure}

Figure~\ref{fig:amlb-aws-archi} shows the architecture and information flow when running the benchmarking tool in the setup used in the experiments for this paper.
The dotted lines indicate calls (communication) whereas the solid lines indicate transfer of files.
In this paper we distribute our workload to AWS EC2 containers which run experiments within a docker environment.

The experiments are initiated from a `local' machine, shown enclosed in a blue rectangle in the bottom.
This local instantiation first reads information from a local configuration file and the command line, after which it uses boto3 to connect to AWS.
It uploads local files required for the experiments, such as the configuration files, to an S3 bucket accessible by the EC2 instances, deploys the jobs to EC2 instances, and tracks the instance status through CloudWatch.

The EC2 instance then installs the specified version of the benchmarking tool
(here of type m5.2xlarge), retrieves the user configuration from the S3 bucket, the specified docker image from the docker hub, and the dataset (task) from OpenML.
The experiment is then run in a docker container (which in turn uses the benchmarking software in `local' mode) which has the benefits of reproducibility of the software stack and requiring much less time than installing the framework from scratch.

After the experiment has completed, results are uploaded to an S3 bucket and the EC2 instance terminates.
The local machine will observe this shutdown and subsequently fetch the downloaded results from the S3 bucket.

As you can see, with this setup there are almost no requirements to the local machine.
Provided the docker images are built and published, the specified benchmark tool version is available online and OpenML is used, the only data transfer between the local machine and the cloud is the upload of configuration files and the download of results.

\subsection{Framework Versions}\label{app:framework-versions}

Experimental results in this paper are mostly from the latest version of each AutoML framework as of early June 2023. Our original submission contained results from frameworks as of September 2021. Because of the delay between our 2021 experiments and submission as well as the delay between our submission and the reviews, and the additional presets we wanted to evaluate, we felt it best to redo our all of our experimental evaluations to reflect a more recent state of the AutoML frameworks.
However, because of budget constraints we could not re-evaluate everything so we chose to re-use some of our old experimental results instead. In the text, these results are denoted by an asterisk (*). We only used old data for frameworks which did not have many updates between the two experimental evaluations.

Below are the exact versions of the frameworks used in both the 2021 and 2023 evaluations. Presets with a $^{'23}$ are only evaluated in 2023.
For each framework, the latest available version as of the $21^{st}$ of September 2023 is also shown.

\begin{table}[ht]
\begin{tabular}{lrrrl}
\toprule
framework & 2021 & 2023 & latest & notes \\
\midrule
\autogluon{} & 0.3.1 & 0.8.0 & 0.8.2 & `best quality' (B), \\
&&&& 'high quality'$^{'23}$ (HQ), and \\
&&&& 'HQ with limited inference time'$^{'23}$ (HQIL) \\
\autosklearn{} & 0.14.0 & 0.15.0 & 0.15.0 & \\
\autosklearnii{} & 0.14.0 & 0.15.0 & 0.15.0 & \\
\flaml{} & 0.6.2 & 1.2.4 & 2.1.0 & \\
\gama{} & 21.0.1 & 23.0.0 & 23.0.0 & `performance' preset \\
\water{} & 3.34.0.1 & 3.40.0.4 &  3.42.0.3 &  \\
\lama{} & 0.2.16 & 0.3.7.3 & 0.3.7.3 & \\
\mljar{} & 0.11.0 & 0.11.5 & 1.0.2  & `compete' and 'perform'$^{'23}$ preset \\
\naml{} & - & 0.0.27 & 0.0.27  & \\
\tpot{} & 0.11.7 & 0.12.0 & 0.12.1 & \\
\bottomrule
\end{tabular}
\caption{Used AutoML framework versions in the experiments.}
\label{table:framework-versions}
\end{table}

\subsubsection{AutoGluon}
With the 'best quality' preset, the models used in the final ensembles are the same ones created to produce out-of-bag predictions during the optimization step. This means that the same algorithm and configuration is represented multiple times, but trained on different subsets of the training data. For the `high quality' preset, instead of using these multiple models a single model for the given algorithm and configuration is trained instead. This typically results in lower predictive performance, but faster inference times. When the `limited inference' constraint is applied, first all models which have too slow inference time are discarded, but otherwise the procedure continues as normal.

\autogluon{}'s `high quality' presets execute a post-processing step, \ie{} refitting the models, after search regardless of elapsed time. Because this possible time constraint violation is by design, we agreed with the authors to we reduce the time constraint communicated to the framework for those presets by $10\%$. With this adjustment, we expected the final training time to observe our time constraints better. For \naml{} we disable early-stopping as detailed in Appendix~\ref{app:naml}.

\subsubsection{TPOT}
\tpot{} requires encoded data to function. The AutoML benchmark provides this encoded data, but the time to encode the data is not include in training time or inference time measurements. For this reason, TPOT's inference time is not reported on in this paper.

\clearpage
\section{AutoML Framework Errors}\label{sec:app-errors}

This appendix contains additional information on the AutoML framework errors encountered while running the experiments.
In the results below, the `fixed' category from Section~\ref{sec:errors} is simply included in the `implementation' error category.
More information on this particular error can be found later in section~\ref{app:sec:error-ag} of this appendix.

\begin{figure}[tb]
     \centering
    \begin{subfigure}{0.48\textwidth}
      \centering
      \includegraphics[width=\textwidth]{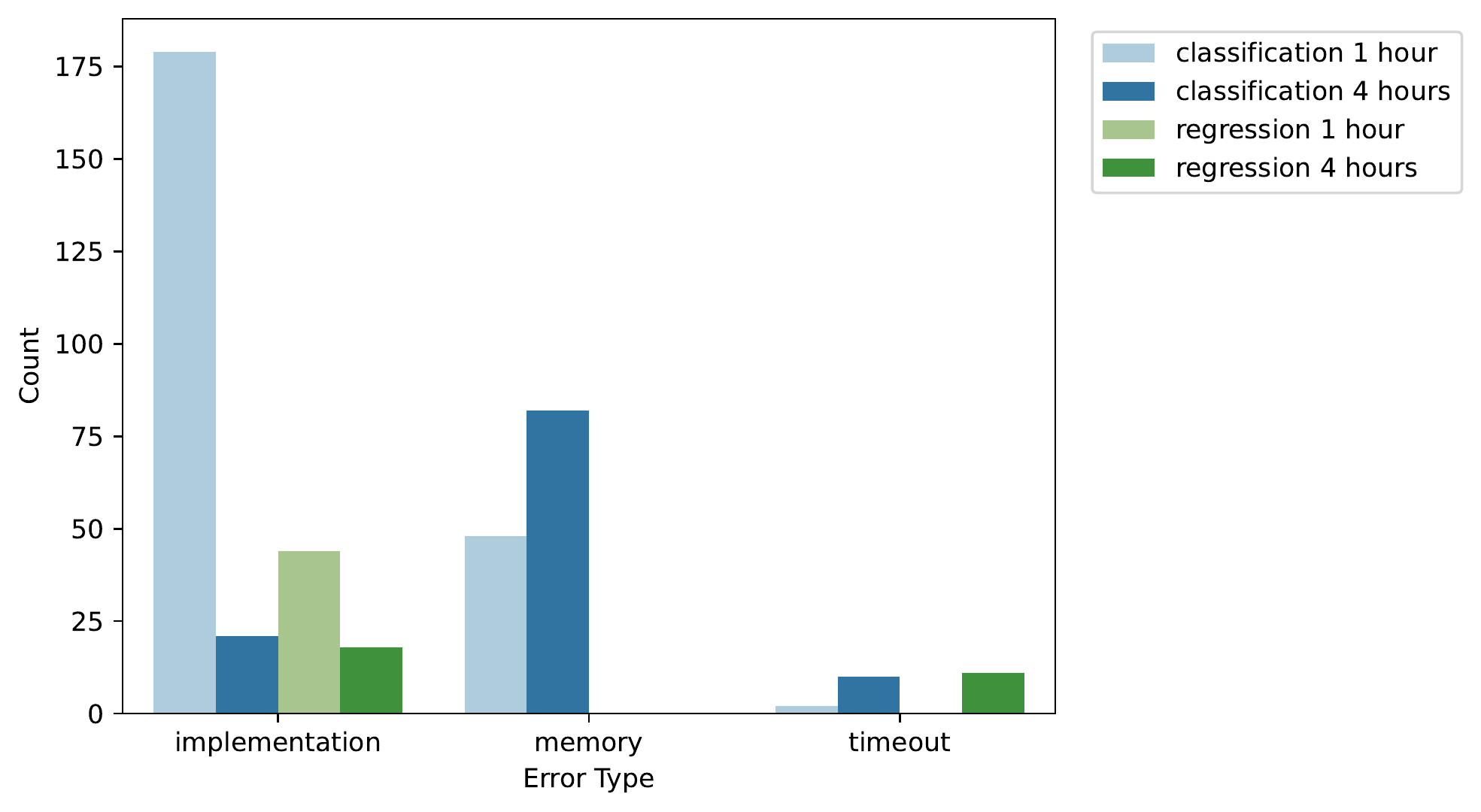}
      \caption{Error count by category and time constraint.}
      \label{app:fig:errors-by-type}
    \end{subfigure}
     \hfill
    \begin{subfigure}{0.48\textwidth}
      \centering
      \includegraphics[width=\textwidth]{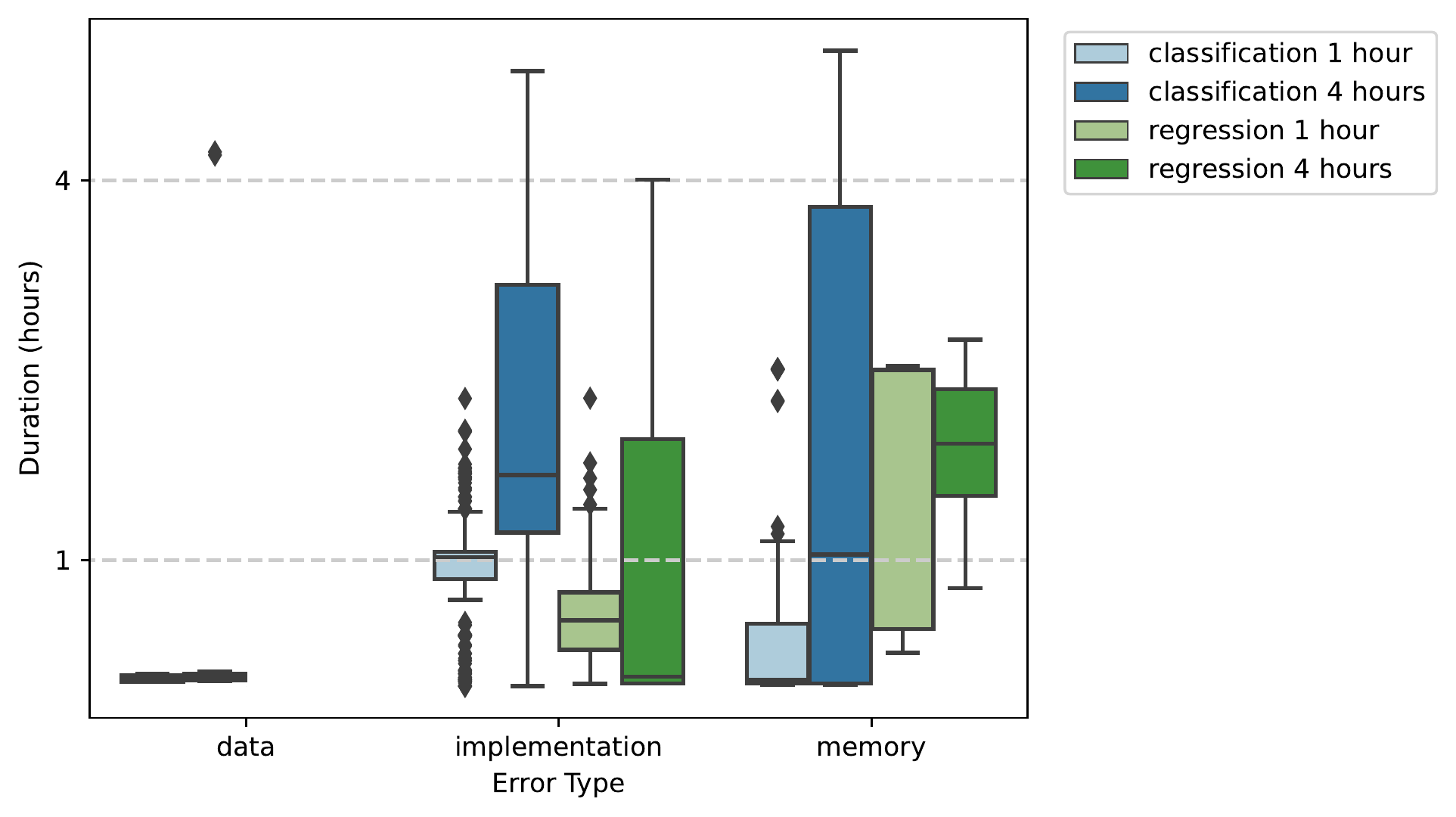}
      \caption{Distribution of when errors occurred.}
        \label{app:fig:errors-by-time}
    \end{subfigure}
    \caption{Figure denoting for each benchmarking suite and time constraint how often errors occurred (on the left) and when (on the right).}
    \label{app:fig:errors}
\end{figure}

Figure~\ref{app:fig:errors-by-type} shows the number of errors encountered across different time constraints and benchmarking suites.
In this comparison, we only report on frameworks which had both a one hour and four hour evaluation in 2023.
As mentioned in Section~\ref{sec:errors}, the amount of memory an timeout errors increase with higher time budgets.
However, the number of implementation errors decrease, lowering the total amount of errors encountered overall. 
We suspect a number of implementation errors on the one hour constraints are due to the framework aborting its pipeline optimization early to adhere to the time constraint while not being prepared to produce predictions yet.
Figure~\ref{app:fig:errors-by-time}, which shows the time at which various errors occurred, supports this hypothesis for the one hour classification tasks where the majority of implementation errors occur.
With a more generous time budget, the AutoML framework might be able to fall back on already optimized models, which could explain why we see fewer implementation errors under a larger time constraint.

\subsection{Class Imbalance}

The two classification tasks with a large amount of failures despite being small are `yeast' and `wine-quality-white', which feature a minority class with only 5 instances.
This means that within the 10-fold cross-validation we perform in our experiments, either 4 or 5 of those instances are available in the training splits.
We see that only in the case where one of those samples is in the test set failures occur.
The exact error message differs per framework, though they indicate that evaluating pipelines fails.
This is likely due to e.g., using 5-fold cross-validation out of the box.
Failure on these specific datasets (and folds) is only observed for \texttt{GAMA}, \texttt{LightAutoML}, and
\texttt{TPOT}.

\subsection{MLJarSupervised}

\texttt{MLJarSupervised}(B) the most `implementation errors', more than twice the next framework.
Of its 284 errors, 256 failures are caused by variations of the following three unique errors:
\begin{itemize}
    \item[25 times] \texttt{['Ensemble\_prediction\_0\_for\_neg\_1\_for\_pos', $\ldots$, \\ '2\_DecisionTree\_prediction\_0\_for\_neg\_1\_for\_pos'] not in index"}
    \item[149 times] \texttt{catboost/libs/data/model\_dataset\_compatibility.cpp:81:\\
    At position 6 should be feature with name 60\_NeuralNetwork\_prediction\_0\_for\_1\_1\_for\_2\\ (found 60\_NeuralNetwork\_prediction).}
    \item[82 times] \texttt{The feature names should match those that were passed during fit. Feature names unseen at fit time:- 100\_Xgboost\_prediction-$\ldots$}
\end{itemize}

This is specific to \texttt{MLJarSupervised}.
While we can only guess, we assume it is related to the extensive AutoML pipeline \texttt{MLJarSupervised} has. 
It includes 10 different steps, including three steps for feature generation and selection and three steps for ensembling and stacking.
These steps are not turned on by default\footnote{\url{https://supervised.mljar.com/features/modes/}}.
Since these three errors only occur in \texttt{MLJARSupervised} in `compete' mode and not in `performance' mode, it likely stems from the ensembling step, which is only used in `compete'.

\subsection{AutoGluon}
\label{app:sec:error-ag} 
In the version of \autogluon{}(HQIL) that we benchmarked (0.8.0), there was a bug in the newly added model calibration which made it crash on some tasks (and only under one hour constraints). However, this bug has since been fixed. We also confirmed that in the version with the fix, \autogluon{}(HQIL) performs very similar to the \autogluon{}(HQ) of version 0.8.0. Because the bug had been fixed and a very accurate imputation strategy was available, we found it more reasonable to impute missing values of \autogluon{}(HQIL) which were specifically caused by this bug with values obtained by \autogluon{}(HQ) instead.

\clearpage
\section{Naive AutoML}\label{app:naml}
\naml{} is introduced as a baseline to compare AutoML frameworks to~\citep{mohr2023naive}.
It designs the machine learning pipeline one step at a time in sequence, ignoring the fact that better performance may be obtained by performing joint algorithm selection and hyperparameter optimization over the whole machine learning pipeline at once. It was not designed to find the best possible model, but instead to provide an adequate model fast. To this end, it employs an aggressive early-stopping heuristic.

We developed the \naml{} integration together with the framework authors. We changed a few hyperparameters from their default configuration in order to disable \naml{}'s early stopping. This allows for comparisons of models produced under similar time constraints. The integration was developed with version $0.0.16$, which proved too unstable to benchmark. The experiments in this appendix ran on version $0.0.27$. We changed the following hyperparameters from their default configuration:

\begin{itemize}
    \item $\systemcase{max\_hpo\_iterations}$ defines after how many hyperparameter optimization iterations of no improvement \naml{} should stop. The default is $100$, but we effectively disable early stopping by setting it to $10^{10}$ iterations.
    \item $\systemcase{execution\_timeout}$ defines how long a single evaluation at any step in the process may take (such as evaluating a random forest model). The default was set to $10$ seconds in version $0.0.16$, which is too short to evaluate models on larger datasets, so we set it to $5\%$ of the total allowed time (that is, 3 minutes for a 1 hour timeout). A later version of \naml{} changed the default to 5 minutes, but we were not aware of this change and the integration script was not changed to reflect that. As we will see below, the $5\%$ limit was too strict.
\end{itemize}

\begin{figure}[hb]
    \centering
    \includegraphics[scale=0.42]{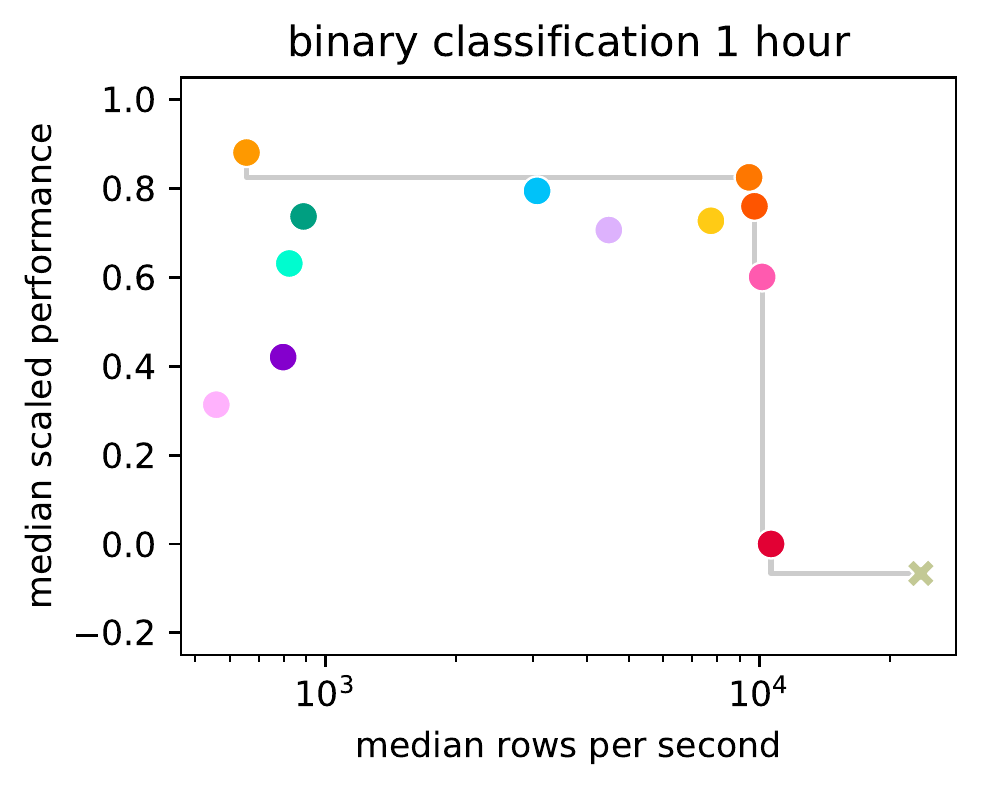}
    \includegraphics[scale=0.42]{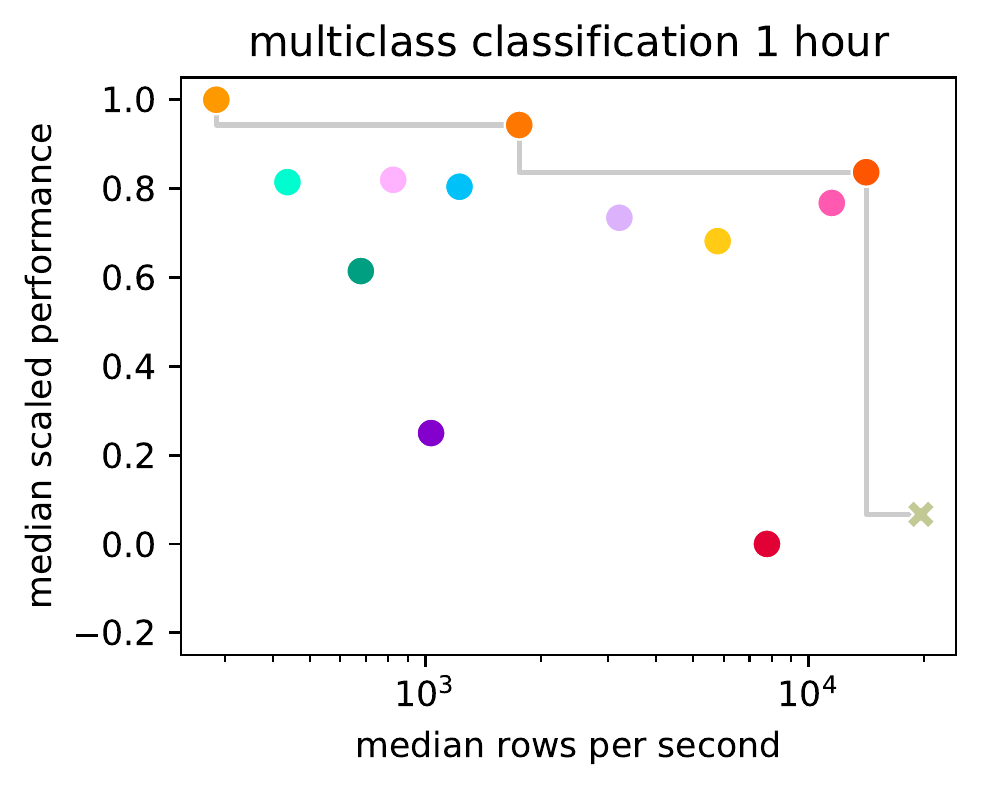}
    \includegraphics[scale=0.42]{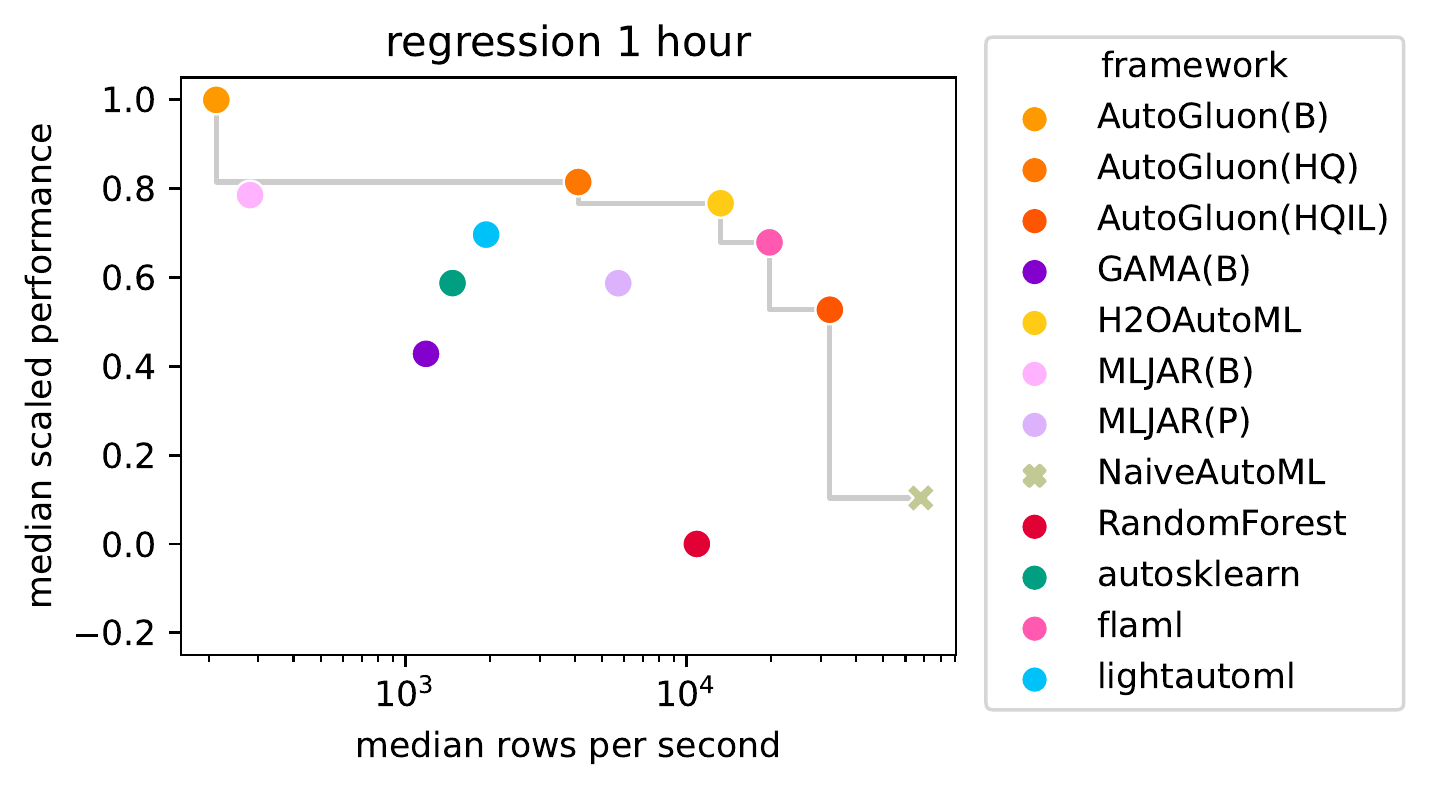}

    \caption{Performance and inference time trade-off as in Figure~\ref{fig:pareto} of Section~\ref{sec:inference}, but with \naml{} results included. \naml{} has bad performance but fast inference speeds due to a bad configuration.}
    \label{fig:naml-inference}
\end{figure}

We evaluated \naml{} on a one hour budget for both the classification and regression benchmark suites. Figure~\ref{fig:naml-inference} shows \naml{} achieving poor performance compared to other AutoML frameworks. This low performance contrasts the evaluation by~\cite{mohr2023naive}, where it achieves performance similar to \autosklearn{}.

This can be entirely explained due to the low $\systemcase{execution\_timeout}$. In many cases, \naml{} fails to evaluate good baseline models such as random forests, and instead only manages to evaluate models such as Gaussian Naive Bayes which have fast train and inference times but relatively poor performance. 
The experiment logs and framework authors confirmed our findings, and explained that their original work also used a larger $\systemcase{execution\_timeout}$. If we had set our $\systemcase{execution\_timeout}$ to a fixed $10$ minutes (instead of $3$), we might have observed a similar performance to that reported by~\cite{mohr2023naive}.

The framework authors state that they plan to integrate their work on learning curve cross-validation (LCCV,~\cite{mohr2023fast}), which would provide automatic scaling of their evaluation resources and make setting $\systemcase{execution\_timeout}$ obsolete. We hope to evaluate this future version of \naml{} and publish the results online.

\newpage

\vskip 0.2in
\bibliography{ref}

\end{document}